%% file: main.tex
\documentclass[10pt,twocolumn,letterpaper]{article}

\usepackage[pagenumbers]{cvpr} 

\input{preamble}

\definecolor{cvprblue}{rgb}{0.21,0.49,0.74}
\usepackage[pagebackref,breaklinks,colorlinks,allcolors=cvprblue]{hyperref}

\usepackage[accsupp]{axessibility}

\usepackage{multirow}
\usepackage{array}
\usepackage{float}
\usepackage{amsmath}
\usepackage[normalem]{ulem}
\useunder{\uline}{\ul}{}

\def\paperID{982} 
\def\confName{CVPR}
\def\confYear{2026}

\title{Bridging the Perception Gap in Image Super-Resolution Evaluation}

\author{Shaolin Su$^{1}$\hspace{1cm} Josep M. Rocafort$^{1,2}$\hspace{1cm} Danna Xue$^{1,2}$ \\
David Serrano-Lozano$^{1,2}$\hspace{1cm} Lei Sun$^{3}$\hspace{1cm} Javier Vazquez-Corral$^{1,2}$
\\$^1$Computer Vision Center
\\$^2$Universitat Autonoma de Barcelona
\\$^3$INSAIT, Sofia University ``St. Kliment Ohridski'' \\
\small{\url{https://color.cvc.uab.cat/rqi/}}
}

\begin{document}
\maketitle

\begin{abstract}

As super-resolution (SR) techniques advance, we observe a growing distrust of evaluation metrics in recent SR research. An inconsistency often emerges between certain evaluation criteria and human perceptual preference. Although current SR research employs varying metrics to evaluate SR performance, it remains underexplored how robust and reliable these metrics actually are. 
To bridge this gap, we conduct a comprehensive analysis of widely used image quality metrics, examining their consistency with human perception when evaluating state-of-the-art SR models. We show that some metrics exhibit only limited—or even negative—correlation with human preferences. We further identify several intrinsic challenges in SR evaluation that compromise the effectiveness of both full-reference (FR) and no-reference (NR) image quality assessment (IQA) frameworks. To address these issues, we propose a simple yet effective Relative Quality Index (RQI) framework, which assesses the relative quality discrepancy between image pairs. Our framework enables easy integration and notable improvements for existing IQA metrics in SR evaluation. Moreover, it can be utilized as a valuable training guide for SR models, enabling the generation of images with more realistic details while maintaining structural fidelity.

\vspace{-2mm}
\end{abstract}

\section{Introduction}
\label{sec1}

Super-resolution aims to reconstruct high-resolution (HR) images from their low-resolution (LR) counterparts, thereby recovering fine textures and structural details that are often lost during image degradation. In recent years, SR techniques \cite{RDN,EDSR,BSRGAN,SwinIR,SeeSR,RealESRGAN} have witnessed rapid progress, with models showing substantial enhancement in visual performance.

However, as SR models continue to evolve, a critical issue has emerged: the commonly used evaluation metrics, such as PSNR, SSIM \cite{ssim} and LPIPS \cite{LPIPS}, often fail to reflect the true perceptual quality of the reconstructed images \cite{pipal,StableSR,SeeSR}. In many cases, models that achieve higher scores on these conventional metrics may not actually produce images that appear more realistic or visually pleasing to human observers. To address this inconsistency, researchers have either conducted extensive user studies \cite{StableSR,SeeSR,Pisasr,dong2025tsd} -- which are costly and time-consuming -- or combined multiple quantitative metrics \cite{PASD, wu2025one,faithdiff,zhang2025uncertainty} in an attempt to provide a more comprehensive evaluation.

Yet, this situation reveals a deeper contradiction: \textit{while SR models have evolved rapidly, the evaluation criteria used to assess them have remained largely unchanged}. Whether these long-standing metrics are still capable of distinguishing the true perceptual performance of modern SR models remains an open and underexplored question. This gap motivates a re-evaluation of existing SR evaluation practices and calls for a deeper understanding of how well current metrics align with human visual perception.

To better understand the reliability of existing evaluation criteria, we first conduct a large-scale user study to thoroughly analyze the alignment between commonly used distortion-oriented FR, perception-oriented FR, NR quality metrics and human perceptual judgments. Specifically, we examine the performance of seven state-of-the-art (SOTA) SR models across five benchmark datasets and compare their metric-based scores with subjective human ratings. The results reveal that the consistency between current quantitative metrics and human perception is rather limited, with some metrics even exhibiting negative correlations with human preference.

We further analyze the challenging aspects that lead to this inconsistency. We identify that each existing evaluation framework suffers from inherent limitations. NR metrics \cite{NIQE,blau20182018,wang2023exploring,yang2022maniqa}, which lack reference images, struggle to capture textural or structural fidelity. Distortion-oriented FR metrics \cite{ssim, FSIM, gmsd} tend to regress toward the average solution across multiple plausible reconstructions, thereby underestimating perceptually superior results. Perception-oriented FR metrics \cite{LPIPS,DISTS}, on the other hand, may fail when the ground-truth images themselves are of suboptimal quality. In addition, as the perceptual quality of algorithm-generated images continues to improve, metrics that perform well in assessing clearly degraded images become less effective in distinguishing subtle differences among high-quality results. Together, these factors  highlight the inadequacy of existing evaluation metrics in accurately reflecting human visual perception of modern SR outputs.

Building upon these observations, we propose a remarkably simple but effective IQA training framework termed Relative Quality Index (RQI), which provides a more reliable and perceptually aligned evaluation for SR. Specifically, the proposed framework (1) accurately assesses fidelity based on reference images, (2) effectively handles cases where the ground-truth images are of suboptimal quality, and (3) enables fine-grained evaluation of subtle perceptual differences by constructing dense training pair comparisons.
Moreover, since the proposed method is designed as a general framework rather than a specific metric, it can be seamlessly integrated into any existing quality assessment model or dataset for training. Extensive experiments demonstrate that the proposed framework not only provides more accurate evaluations of SR models, but it can also be employed as a loss function to guide SR training, further enhancing the performance of advanced SR models.

Note that due to the ill-posed nature of the SR problem, we focus on the perceptual evaluation of SR models rather than pixel-wise distortion-based measurements. Nevertheless, our approach proves effective in assessing and preserving textural and structural fidelity.

The contributions of this work are threefold:

\begin{itemize}
  
  \item
  We conduct an in-depth investigation into how accurately current full-reference and no-reference quality metrics reflect human perception in SR evaluation, and provide detailed insights into the underlying causes of their limited effectiveness.
  
  \item
  We propose a general and lightweight framework that addresses the key issues limiting existing metrics, while being easily adaptable for integration with various quality assessment models and datasets.

  \item 
  We demonstrate through comprehensive experiments that the proposed framework achieves more reliable and perceptually consistent SR evaluation results. Moreover, the metric can also serve as a loss function guiding SR model optimization toward improved visual performance.
\end{itemize}

\section{Related Work}
\label{sec2}

\subsection{Image Super-Resolution}
Early SR approaches~\cite{SRCNN,EDSR,RDN,dai2019second,dong2015image,chen2021pre} operate under the assumption of a predefined degradation process, such as bicubic downsampling or blurring with a known prior. However, their effectiveness is often constrained in real-world scenarios where noise or compression artifacts may occur. To handle practical SR, BSRGAN \cite{BSRGAN} and RealESRGAN \cite{RealESRGAN} assume complex degradation in LR images and adversarially train models to remove multiple degradations. Transformer-based methods further improve HR quality by capturing long dependencies between pixels: SwinIR \cite{SwinIR} utilizes window attention in Swin Transformer \cite{liu2021swin} for global dependencies, while HAT \cite{HAT} optimizes hierarchical feature integration. With the emergence of diffusion techniques \cite{ho2020denoising}, diffusion-based SR models \cite{StableSR,SeeSR,PASD,wu2024one,wang2024sinsr,Pisasr,faithdiff} have demonstrated powerful ability in reconstructing image details and generating high perceptual quality outputs. StableSR \cite{StableSR} utilizes rich prior from the Stable Diffusion model conditioned by the LR input, SeeSR \cite{SeeSR} explores the roles of tag-style prompts in guiding semantic robust image generation, FaithDiff \cite{faithdiff} proposes to align the noisy latents with the LR features for maintaining structural consistency, and PiSA-SR \cite{Pisasr} balances pixel-level fidelity and perceptual quality by introducing dual LoRA \cite{hu2022lora} adapters.

However, as the perceptual quality of SR outputs continues to advance, recent studies have revealed a persistent inconsistency between human-perceived quality and quantitative evaluations produced by existing image quality metrics \cite{StableSR,SeeSR,Pisasr,dong2025tsd,AFINE}. Consequently, researchers often resort to either labor-intensive user studies \cite{StableSR,SeeSR,PASD,Pisasr,dong2025tsd} or the use of multiple, sometimes conflicting, image metrics \cite{SeeSR,PASD,Pisasr,faithdiff,zhang2025uncertainty} to validate SR model performance. In this paper, we investigate this inconsistency issue more closely and propose a framework that effectively addresses the underlying challenges.

\subsection{Image Quality Metrics}

\begin{table*}[t]
\centering
\setlength{\tabcolsep}{7pt}
\renewcommand\arraystretch{1.1}
\caption{Consistency evaluations of quality metrics with human perception. SRCC, PLCC and winning rate are reported.}
\vspace{-3mm}
\footnotesize
\resizebox{0.98\textwidth}{!}{
\begin{tabular}{c|c|ccccccccccc}
\bottomrule[0.12em]
\rowcolor{tableHeadGray}
Dataset                    & Criterion & SSIM   & PSNR   & DISTS   & LPIPS   & NIQE         & PI    & Clip-IQA     & MANIQA  & AFINE & DeQA-Score     & RQI            \\ \hline \hline
\multirow{3}{*}{DIV2K \cite{div2k}}       & SRCC      & -0.348 & -0.079 & 0.610  & 0.415  & 0.516         & 0.565 & 0.593          & 0.554  & 0.581 & 0.613 & \textbf{0.744} \\
                             & PLCC      & -0.360 & -0.124 & 0.627  & 0.452  & 0.492         & 0.549 & 0.593          & 0.553  & 0.653      & 0.662 & \textbf{0.785} \\
                             & Win Rate  & 0.05   & 0.19   & 0.44   & 0.41   & 0.44          & 0.57  & 0.50            & 0.47   & 0.40  & 0.51  & \textbf{0.65}  \\ \hline
\multirow{3}{*}{RealSR \cite{RealSR}}      & SRCC      & -0.220 & -0.116 & 0.048  & 0.008  & 0.263         & 0.317 & 0.377          & 0.187  & 0.449 & 0.452    & \textbf{0.504} \\
                             & PLCC      & -0.289 & -0.160 & 0.027  & -0.031 & 0.282         & 0.325 & 0.437          & 0.212  & 0.453 & 0.467  & \textbf{0.484} \\
                             & Win Rate  & 0.04   & 0.05   & 0.12   & 0.11   & 0.47          & 0.51  & \textbf{0.58}  & 0.32   & 0.36  & 0.43  & 0.49           \\ \hline
\multirow{3}{*}{DRealSR \cite{DRealSR}}     & SRCC      & -0.354 & -0.355 & -0.102 & -0.141 & 0.240         & 0.303 & 0.268          & 0.284  & 0.484 & 0.437   & \textbf{0.529} \\
                             & PLCC      & -0.409 & -0.405 & -0.129 & -0.143 & 0.222         & 0.301 & 0.268          & 0.331  & 0.542 & 0.491  & \textbf{0.603} \\
                             & Win Rate  & 0.02   & 0.01   & 0.10   & 0.04   & 0.44          & 0.46  & 0.38           & 0.35   & 0.44  & 0.47   & \textbf{0.53}  \\ \hline
\multirow{3}{*}{Set5\&Set14 \cite{bevilacqua2012low,zeyde2012single}} & SRCC      & -0.321 & -0.204 & 0.403  & 0.282  & 0.578         & 0.466 & 0.642          & 0.437  & 0.578 & \textbf{0.699}    & 0.664 \\
                             & PLCC      & -0.387 & -0.239 & 0.414  & 0.293  & 0.527         & 0.506 & 0.683 & 0.443  & 0.543 & \textbf{0.735} & 0.673          \\
                             & Win Rate  & 0.06   & 0.06   & 0.35   & 0.24   & 0.41 & 0.29  & 0.29           & 0.24   & 0.41  & \textbf{0.52}  & 0.35        \\ \bottomrule
\end{tabular}}
\vspace{-3mm}
\label{tab1}
\end{table*}

Early studies on IQA primarily focused on images with clearly noticeable degradations. Under such conditions, pixel-wise FR metrics such as PSNR and SSIM \cite{ssim} were found to correlate reasonably well with perceived quality. However, these distortion-oriented approaches inherently favor the expected average among multiple plausible reconstructions, thereby rewarding overly smooth or blurry outputs, contradicting human perceptual judgments \cite{blau2018perception, lee2025auto}. Subsequent FR metrics shifted their focus toward perceptual similarity by measuring distances in deep feature spaces, such as LPIPS \cite{LPIPS}, DISTS \cite{DISTS} and NeuralSBS \cite{khrulkov2021neural}, which better capture high-level perceptual cues. Nevertheless, recent work \cite{AFINE} has pointed out that references themselves may exhibit suboptimal quality, introducing systematic bias to the evaluation. This limitation becomes particularly evident in SR, where advanced SR model outputs are surpassing ground-truths (GT) that are collected in early years without careful control of degradations.
Meanwhile, recent SR studies also adopt NR-IQA metrics for evaluations. Widely applied metrics include Natural Scene Statistics (NSS) based metrics such as NIQE \cite{NIQE}, IL-NIQE \cite{IL-NIQE} and PI \cite{blau20182018}, and deep-learning based metrics such as MUSIQ \cite{ke2021musiq}, MANIQA \cite{yang2022maniqa} and Clip-IQA \cite{wang2023exploring}. While these methods can effectively capture low-level distortions without requiring GTs, the absence of reliable references makes it difficult for them to make fidelity evaluations.

Note that AFINE \cite{AFINE} also targets SR evaluation under the imperfect GT assumption, however, our work distinguishes from it in three aspects: (1) we conduct a comprehensive analysis of the SR evaluation problem and identify the inherent challenges from multiple aspects. (2) Rather than designing a specific model, we propose a general and adaptable framework that effectively mitigates these challenges. (3) Our approach achieves superior perceptual consistency while being trained solely on existing IQA datasets, without the need to collect SR-specific data as required by AFINE.

\section{Revisiting Image Metrics for SR Evaluation}
\label{sec3}

\subsection{User Study on SR Benchmarks}

To analyze how existing image quality metrics correlate with human perception, we conduct a comprehensive user study to collect subjective scores on outputs from different SR models. Specifically, we focus on the classic $\times4$ SR task and gather human opinions on the results of seven representative SR models, including two GAN-based models (RealESRGAN~\cite{RealESRGAN} and BSRGAN~\cite{BSRGAN}), two transformer-based models (SwinIR~\cite{SwinIR} and HAT~\cite{HAT}), and three diffusion-based models (StableSR~\cite{StableSR}, SeeSR~\cite{SeeSR}, and PASD~\cite{PASD}). As recently observed \cite{AFINE}, even GT images may exhibit suboptimal visual quality; therefore, we also collect user ratings on the corresponding GT images. All models are evaluated using their officially released weights on five widely used SR benchmarks: DIV2K-wild~\cite{div2k}, RealSR~\cite{RealSR}, DRealSR~\cite{DRealSR}, Set5~\cite{bevilacqua2012low}, and Set14~\cite{zeyde2012single}.

After obtaining all the images, we conduct the user study following a two-alternative forced choice (2AFC) paradigm. Observers are asked to select the HR image that exhibits better perceptual quality from a pair of images showing the same content. 
The experiment was carried out in a well-controlled environment, with participants provided with clear instructions. Each comparison was rated by at least 15 participants.
Finally, we employ Thurstone’s Case V model~\cite{thurstone1927psychophysical} to reconstruct a global ranking based on user preferences. In total, we collect subjective ratings for $8 \times 312 = 2,496$ images, covering all test images from the five SR benchmarks. See more details about the study in our Supplementary Material.

\subsection{Analysis of Existing Metrics}
\label{sec3.2}

With the collected user perception scores, we perform an in-depth analysis of how existing image quality metrics align with human visual judgments. We select a set of widely adopted metrics commonly used in SR research and evaluate their perceptual consistency. The selected metrics cover a wide range of IQA categories, including distortion-oriented FR metrics (SSIM~\cite{ssim} and PSNR), perception-oriented FR metrics (DISTS~\cite{DISTS} and LPIPS~\cite{LPIPS}), traditional NR IQA metrics (NIQE~\cite{NIQE} and PI~\cite{blau20182018}), deep learning-based NR metrics (Clip-IQA~\cite{wang2023exploring} and MANIQA~\cite{yang2022maniqa}), one recent FR based asymmetric metric AFINE~\cite{AFINE}, and one recent LLM based model DeQA-Score \cite{deqa}.  
All metrics are implemented using their officially released weights and applied to evaluate SR outputs. For each source image, we compute the Spearman Rank-Order Correlation Coefficient (SRCC) and Pearson Linear Correlation Coefficient (PLCC) between metric scores and user opinions on each source content, and report their averages over each dataset. Additionally, since SR evaluation often emphasizes on the best-performing model, we also calculate the prediction accuracy for the top-ranked model within each dataset (Winning Rate). The results are summarized in the first 12 columns of Table~\ref{tab1}, based on which we make several observations.

First, traditional distortion-oriented FR metrics show poor correlation with human perceptual judgments. As shown in Table~\ref{tab1}, both FR metrics even produce evaluations opposite to human preference across the datasets. The result is consistent with the well-known perception-distortion trade-off~\cite{blau2018perception}, where the metrics tend to minimize the expected reconstruction error toward the average of multiple plausible solutions, thereby favoring overly smooth or blurry patterns.  
Second, we observe a similar behavior from the perception-oriented FR metrics DISTS~\cite{DISTS} and LPIPS~\cite{LPIPS}. While they exhibit moderate consistency on DIV2K and Set5\&Set14, their performance drops significantly on RealSR and DRealSR, probably due to the relatively low quality of GT images in RealSR and the large image resolutions in DRealSR, which make perceptual comparisons more difficult.  
Third, and somewhat counterintuitively, NR-IQA metrics---including both traditional models (NIQE~\cite{NIQE}, PI~\cite{blau20182018}) and deep learning-based models (Clip-IQA~\cite{wang2023exploring}, MANIQA~\cite{yang2022maniqa})---exhibit higher consistency with human perception. However, they may also suffer from the risk of lacking reference to make moderate evaluations, as will be analyzed in the following. 
Fourth, the AFINE model \cite{AFINE}, built upon the imperfect reference assumption, achieves better consistency with human preferences. This can be attributed both to the asymmetric design and training on SR specific data.
Finally, benefiting from the strong representation capabilities of foundation models and large-scale cross-dataset training, the LLM-based model DeQA-Score \cite{deqa} demonstrates relatively good performance. However, its large model size and substantial computational cost inherently limit its broader applicability, such as being employed as a perceptual feedback loss for SR optimization. In contrast, we show that with our proposed framework, comparative or better alignment with human preferences can be achieved with conventional, non-LLM architectures.

\begin{figure}[t]
\centering
\includegraphics[width=0.95\linewidth]{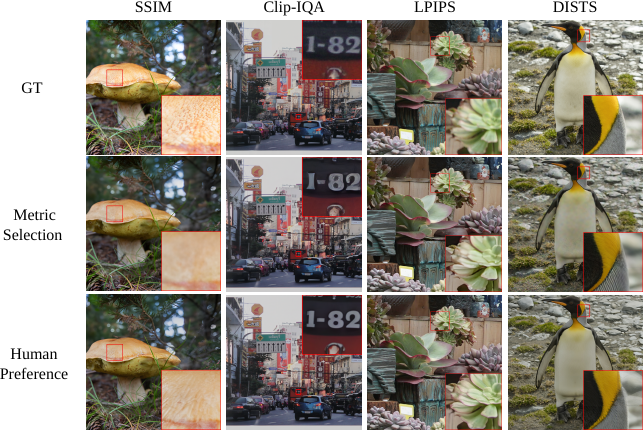}
\caption{Visual illustration of how SR evaluation challenges current metrics in different aspects. Zoom in for better comparison.}
\label{fig1}
\vspace{-6pt}
\end{figure}

In the following, we provide a more detailed analysis of the inherent challenges that SR evaluation poses to current image quality metrics. As shown in Figure~\ref{fig1}, distortion-based FR metrics (first column) tend to favor averaged and smoother regions over perceptually diverse textures, leading to evaluations that contradict human preference. For NR metrics (second column), they often fail to capture fidelity changes in fine details such as characters due to the absence of proper references, therefore assigning higher scores to more textural results rather than how the image is supposed to appear.  
For perceptual FR-IQA metrics (third column), due to the presence of suboptimal GT images and increasingly powerful SR models, they can also fail when the GT image quality is inferior and the SR output perceptually surpasses it. Consequently, these similarity metrics favor results that more closely resemble the GTs, rather than those exhibit genuinely higher perceptual quality. 
Lastly, we observe that SOTA SR model outputs often exhibit only subtle perceptual differences (fourth column). However, many existing IQA metrics are primarily designed to capture large, easily noticeable degradations. Consequently, they may fail to generalize to fine-grained SR evaluation tasks involving subtle quality variations.

In summary, our analysis reveals that current SR evaluation settings introduce multiple challenges to perceptual IQA metrics from different aspects. And a robust SR metric is expected to be able to \textbf{Goal 1:} accurately assess fidelity based on references, \textbf{Goal 2:} remain reliable even when ground truths are of suboptimal quality, and \textbf{Goal 3:} perform fine-grained discrimination among high-quality SR results. In the next section, we show that these goals can be achieved through a remarkably simple framework, without the need of collecting any additional training data, or adapting existing IQA methods.

\section{The RQI Framework}
\label{sec4}

Considering Goal 1, we design our approach as a reference-based framework, ensuring reliable measurement of structural fidelity between SR outputs and ground-truth images.

Regarding Goal 2, where reference images may contain distortions, we propose to boldly serve arbitrary images, including distorted ones as references during training. Due to this paradigm change, we then propose to calculate the relative quality discrepancy between target and reference image as training objective.
Moreover, considering cases in SR evaluation where model outputs may surpass GTs, we accept that the relative quality discrepancy can be either higher or lower when comparing target image to the reference. This formulation enables the model, during training, to naturally learn the capability required for robust evaluation under imperfect references.

An additional and elegant advantage of this design lies in that, since existing IQA datasets typically contain multiple degraded versions of the same scene, computing pairwise relative quality differences among any of them inherently provides both large and subtle quality variations. Consequently, the model learns to distinguish not only significant degradations but also fine-grained perceptual differences, thereby achieving Goal 3. In the following, we illustrate the RQI framework in training and evaluation settings.

\subsection{Training Under the RQI Framework}

In existing IQA datasets, given one reference image $I_{0}$ and a sequence of $n$ distorted versions $\{I_1,I_2, ...,I_n \}$ under the same scene but covering varying distortion types and degradation levels, traditional FR framework constructs image pairs $\{I_{0},I_i\}$ and quality score $q_i$ for training IQA models, where $q_i$ is the Mean Opinion Scores (MOS) for $I_i$. In the RQI framework,
we select any image $I_{j},j\in \{1, 2, \ldots, n\}$ from the sequence as the reference. Then, given the target image $I_{i},i\in \{1, 2, \ldots, n\},i\neq j$,
we train IQA models to learn the relative quality discrepancy between the image pair $\{I_{i},I_j\}$, denoting the relative quality of target $I_{i}$ when referenced by $I_{j}$.
Since the MOS in existing IQA datasets already represent averaged human perceptual judgments with reduced uncertainty, we directly use the difference between the two MOS values, 
$q_i-q_j$, as the training supervision signal, where $q_i$ and $q_j$ are subjective quality scores for image $I_{i}$ and $I_{j}$. This formulation can be regarded as a linear approximation of the perceptual quality discrepancy between two images. By modeling RQI as a regression problem --- in contrast to the probability-based formulations \cite{bradley1952rank, AFINE, unique}, it provides stable and dense gradients even for small score differences. Consequently, our framework performs robustly for samples with subtle perceptual distinctions, a challenging aspect in SR evaluation.

Note that $q_i-q_j$ can be either positive or negative, indicating that the quality of the target image may surpass or fall below that of the reference. Meanwhile, the constructed discrepancy labels are order-sensitive; that is, the image pair $\{I_{i},I_j\}$, representing the relative quality of $I_{i}$ referenced by $I_{j}$, and its reversed pair $\{I_{j},I_i\}$, will yield opposite evaluations, $q_i-q_j$ and $q_j-q_i$, respectively.
After constructing training pairs, we adopt Huber loss as the training objective:

\begin{equation}
L = 
\begin{cases}
\frac{1}{2}\left(\hat{y}_{ij} - (q_i - q_j)\right)^2, \text{if } \left| \hat{y}_{ij} - (q_i - q_j) \right| \le \delta, \\[6pt]
\delta \left( \left| \hat{y}_{ij} - (q_i - q_j) \right| - \frac{1}{2}\delta \right), \text{otherwise.}
\end{cases}
\label{eq:loss}
\end{equation}

\noindent
where $\hat{y}_{ij}=f_{RQI}(I_i,I_j)$, denoting the model predicted quality of $I_i$ relative to $I_j$, and $\delta$ is the smooth threshold.
The Huber loss provides smooth gradients for small prediction errors and thus enabling effective learning of fine-grained perceptual relations.

Under this framework, we can train arbitrary models on arbitrary datasets. In practice, we only need to adjust the training data preparation by constructing denser image pairs and recalculating their training labels, without modifying the architectures of concrete FR models.

\begin{figure}[t]
\hspace{-0.3cm}
\centering
\includegraphics[width=0.9\linewidth]{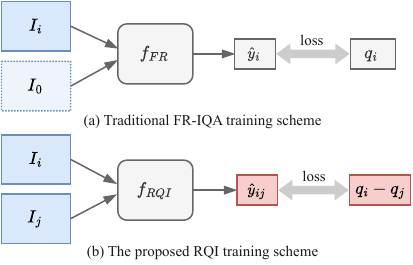}
\vspace{-2mm}
\caption{The proposed RQI training scheme differs from traditional FR-IQA training scheme in three aspects: 1. RQI is asymmetric. 2. RQI calculates relative discrepancy. 3. Denser image pairs are constructed to facilitate challenging predictions.}
\vspace{-0.4cm}
\label{fig2}
\end{figure}

We analyze the differences between the proposed RQI framework and the traditional FR-IQA training paradigm from three aspects, as illustrated in Figure~\ref{fig2}:

\noindent(1)~\textit{RQI is asymmetric.}
The output depends on the order of the input pair: swapping the two inputs produces the opposite result. In contrast, in many FR metrics~\cite{ssim,LPIPS,DISTS}, $f_{FR}(\cdot)$ produce identical outputs for $f_{FR}(I_{i},I_{j})$ and $f_{FR}(I_{j},I_{i})$, as they compute absolute differences under the assumption that the reference image is always of perfect quality.

\noindent(2)~\textit{RQI calculates relative discrepancies.}~This is achieved by comparing the target image against any other degraded image from the same sequence, enabling the model to learn how complex degradation relationships map to perceptual preferences. In contrast, the traditional FR-IQA framework relies on absolute quality labels $q_i$ assigned to $I_i$.

\noindent(3)~\textit{RQI constructs dense pairwise comparisons.} Traditional FR schemes only compare distorted images with the reference, forming pairs $\{I_{0},I_i\},i\in \{1, \ldots, n\}$. RQI instead constructs dense pairwise comparisons between arbitrary image pairs $\{I_{i},I_j\},i,j\in \{1, 2, \ldots, n\},i\neq j$, covering more complex cases where both the “reference’’ and the “target’’ may contain different degradations. This also enables fine-grained quality prediction, as many of the image pairs exhibit subtle perceptual differences.

\subsection{Evaluation Under RQI}

Evaluating under the RQI framework is straightforward. Given any SR model output $I_{HR}$, we compute its perceptual quality relative to the corresponding GT image $I_{GT}$ in a fixed input order, where $I_{HR}$ is always placed first. The model outputs a scalar score $s$, representing the perceptual quality of the SR result with respect to its reference. This score can be directly used to assess the performance of the 
SR model by comparing across all the evaluated models, with a higher score indicating better perceptual quality. The evaluation process can be formulated as:

\begin{equation}
s_i=f_{RQI}(I_{HR},I_{GT})
\label{eq:eval}
\end{equation}

\noindent
where $f_{RQI}(\cdot)$ denotes the IQA model after training under the RQI framework.

\subsection{Implementation Details}

Since the proposed RQI is a general framework, it can be applied to train various IQA models on different datasets.
In practice, we train three representative IQA architectures --- AHIQ \cite{lao2022attentions}, MANIQA \cite{yang2022maniqa}, and TOPIQ \cite{chen2024topiq} --- on three commonly used IQA datasets: Kadid-10K \cite{lin2019kadid}, PieAPP \cite{PieAPP}, and PIPAL \cite{pipal}.
We do not train on the perceptual IQA dataset BAPPS \cite{BAPPS} due to its limited patch resolution ($64\times64$), which is insufficient for the selected model architectures.
Since MANIQA \cite{yang2022maniqa} is originally designed as an NR-IQA model, we modify it to accept both the target and reference images. 
For all models, we remove the activation function in the final regression layer to allow the output of negative values, which are necessary for representing bidirectional quality discrepancies.

For each dataset, we construct relative quality discrepancy labels for all possible image pairs and normalize them to $[-1, 1]$, where larger values indicate better perceptual quality. Datasets are split into training and validation subsets (8:2) with non-overlapping scenes. The best-performing model on the validation set is then evaluated in a zero-shot manner on other datasets.
More implementation details can be found in our Supplementary Material.

\section{Experiments}\label{sec5}

In this section, we first evaluate the effectiveness of the proposed RQI framework by training different IQA models across multiple datasets.
We then assess its generalization capability on both SR and general IQA benchmarks.
Finally, we apply RQI-based metric as alternative optimization objective for training SR models, demonstrating that RQI not only enables more accurate quality assessment but also leads to visually superior reconstruction results. In all our experiments, the models are evaluated in a zero-shot manner, \textit{i.e.}, each model is trained on the source dataset and directly tested on the entire target dataset. This design further validates the practical effectiveness of the metrics.

\begin{table}[t]
\centering
\setlength{\tabcolsep}{7pt}
\renewcommand\arraystretch{1.1}
\caption{The effectiveness of the proposed RQI framework. We train different IQA models across multiple datasets following the traditional FR-IQA setting and the RQI scheme (with subscript `R'). SRCC consistency with user opinions are reported.}
\vspace{-3mm}
\smallskip
\resizebox{\columnwidth}{!}{
\begin{tabular}{c|cc|cc|cc}
\bottomrule[0.12em]
\rowcolor{tableHeadGray}
Train Set & Kadid & Kadid$_{\text{R}}$ & PieAPP & PieAPP$_{\text{R}}$ & PIPAL & PIPAL$_{\text{R}}$ \\ 
\hline \hline
Model & \multicolumn{6}{c}{AHIQ} \\ 
\hline 
DIV2K & 0.365 & \textbf{0.506} & 0.459 & \textbf{0.515} & 0.431 & \textbf{0.626} \\
RealSR & \textbf{0.196} & 0.181 & 0.095 & \textbf{0.284} & 0.452 & \textbf{0.474} \\
DRealSR & 0.267 & \textbf{0.350} & 0.244 & \textbf{0.378} & 0.413 & \textbf{0.467} \\
Set5\&14 & 0.292 & \textbf{0.426} & 0.203 & \textbf{0.378} & 0.280 & \textbf{0.472} \\ 
\hline
Model & \multicolumn{6}{c}{MANIQA} \\ 
\hline
DIV2K & 0.502 & \textbf{0.550} & \textbf{0.573} & 0.570 & 0.624 & \textbf{0.744} \\
RealSR & 0.238 & \textbf{0.258} & 0.098 & \textbf{0.208} & 0.470 & \textbf{0.504} \\
DRealSR & 0.343 & \textbf{0.343} & 0.285 & \textbf{0.417} & 0.372 & \textbf{0.529} \\
Set5\&14 & 0.472 & \textbf{0.504} & 0.386 & \textbf{0.483} & 0.544 & \textbf{0.588} \\ 
\hline
Model & \multicolumn{6}{c}{TOPIQ} \\ 
\hline
DIV2K & 0.462 & \textbf{0.490} & 0.374 & \textbf{0.414} & 0.490 & \textbf{0.561} \\
RealSR & 0.101 & \textbf{0.233} & 0.107 & \textbf{0.133} & 0.277 & \textbf{0.328} \\
DRealSR & 0.164 & \textbf{0.282} & 0.003 & \textbf{0.322} & 0.042 & \textbf{0.357} \\
Set5\&14 & 0.060 & \textbf{0.334} & \textbf{0.035} & 0.010 & 0.025 & \textbf{0.271} \\ 
\bottomrule
\end{tabular}
}
\vspace{-0.4cm}
\label{tab2}
\end{table}

\subsection{The Effectiveness of the RQI Framework}

We train AHIQ \cite{lao2022attentions}, MANIQA \cite{yang2022maniqa}, and TOPIQ \cite{chen2024topiq} across three IQA datasets Kadid-10K \cite{lin2019kadid}, PieAPP \cite{PieAPP}, and PIPAL \cite{pipal}, following the traditional FR-IQA and the RQI training settings respectively. We test them on the user perceptual data collected in Section \ref{sec3}, covering seven SOTA SR models from four different SR benchmarks (Set5 and Set14 are merged for evaluation).
We calculate SRCC between model predictions and user opinions for each source image, and report the averaged SRCC value in Table \ref{tab2}, where subscript ``R'' denotes trained under the RQI setting.

\begin{table*}[t]
\centering
\setlength{\tabcolsep}{7pt} 
\renewcommand\arraystretch{1.1}
\caption{Consistency evaluations of image metrics on four IQA benchmarks. BSD-SR \cite{ma2017learning}, QADS \cite{zhou2022quality} and SRIQA-Bench \cite{AFINE} are  SR-IQA datasets, and Kadid-10K \cite{lin2019kadid} is a general IQA dataset. The best and second best performances are in \textbf{bold} and {\ul underscore}.}
\vspace{-2mm}
\resizebox{0.98\textwidth}{!}{
\footnotesize
\begin{tabular}{c|c|ccccccccccc}
\bottomrule[0.12em]
\rowcolor{tableHeadGray}
Dataset                    & Criterion & SSIM   & PSNR   & DISTS   & LPIPS   & NIQE         & PI    & Clip-IQA     & MANIQA  & AFINE          & DeQA-Score  & RQI            \\ \hline \hline
\multirow{4}{*}{BSD-SR \cite{ma2017learning}}   & SRCC$_{\text{mean}}$ & \textbf{0.949} & {\ul 0.945} & 0.947          & 0.901          & 0.668 & 0.875          & 0.793    & 0.789  & 0.915          & 0.889          & 0.901          \\
                           & PLCC$_{\text{mean}}$ & 0.617          & 0.438       & 0.827          & 0.624          & 0.639 & \textbf{0.849} & 0.703    & 0.759  & 0.831          & 0.841          & {\ul 0.842}    \\
                           & SRCC$_{\text{all}}$  & {\ul 0.945}    & 0.940       & \textbf{0.950} & 0.910          & 0.664 & 0.893          & 0.786    & 0.815  & 0.816          & 0.856          & 0.901          \\
                           & PLCC$_{\text{all}}$  & 0.625          & 0.454       & 0.828          & 0.611          & 0.643 & \textbf{0.868} & 0.708    & 0.760  & 0.752          & 0.836          & {\ul 0.840}    \\ \hline
\multirow{4}{*}{QADS \cite{zhou2022quality}}      & SRCC$_{\text{mean}}$ & \textbf{0.927} & 0.727       & 0.887          & 0.832          & 0.420 & 0.760          & 0.704    & 0.842  & {\ul 0.925}    & {\ul 0.933}    & 0.912     \\
                           & PLCC$_{\text{mean}}$  & 0.552          & 0.193       & 0.703          & 0.619          & 0.394 & 0.708          & 0.489    & 0.759  & \textbf{0.930} & {\ul 0.911} & 0.828     \\
                           & SRCC$_{\text{all}}$ & \textbf{0.923} & 0.562       & 0.869          & 0.823          & 0.398 & 0.704          & 0.685    & 0.826  & 0.865          & 0.836          & {\ul 0.910}     \\
                           & PLCC$_{\text{all}}$  & 0.547          & 0.213       & 0.706          & 0.618          & 0.327 & 0.651          & 0.488    & 0.733  & \textbf{0.844} & 0.824          & {\ul 0.832} \\ \hline
\multirow{4}{*}{SRIQA-Bench \cite{AFINE}}  & SRCC$_{\text{mean}}$ & -0.282         & -0.346      & 0.547          & 0.466          & 0.557 & 0.616          & 0.658    & 0.605  & \textbf{0.778} & 0.681          & {\ul 0.733}    \\
                           & PLCC$_{\text{mean}}$ & 0.098          & 0.337       & 0.564          & 0.543          & 0.587 & 0.682          & 0.708    & 0.635  & 0.726          & \textbf{0.760} & {\ul 0.739}    \\
                           & SRCC$_{\text{all}}$  & 0.064          & 0.078       & 0.432          & 0.362          & 0.466 & 0.511          & 0.532    & 0.489  & 0.574          & {\ul 0.598}    & \textbf{0.609}    \\
                           & PLCC$_{\text{all}}$  & 0.109          & 0.330       & 0.471          & 0.420          & 0.457 & 0.552          & 0.559    & 0.504  & 0.555          & \textbf{0.622} & {\ul 0.564}          \\ \hline
\multirow{4}{*}{Kadid-10K \cite{lin2019kadid}} & SRCC$_{\text{mean}}$  & 0.649          & 0.261       & -              & \textbf{0.809} & 0.393 & 0.406          & 0.558    & 0.548  & -              & -              & {\ul 0.669}    \\
                           & PLCC$_{\text{mean}}$ & 0.633          & 0.247       & -              & \textbf{0.801} & 0.379 & 0.376          & 0.480    & 0.531  & -              & -              & {\ul 0.651}    \\
                           & SRCC$_{\text{all}}$  & 0.595          & 0.231       & -              & \textbf{0.741} & 0.435 & 0.474          & 0.534    & 0.574  & -              & -              & {\ul 0.666}    \\
                           & PLCC$_{\text{all}}$  & 0.585          & 0.229       & -              & \textbf{0.720} & 0.389 & 0.425          & 0.485    & 0.548  & -              & -              & {\ul 0.649}           \\ 
                           \bottomrule
\end{tabular}}
\label{tab3}
\end{table*}

\begin{figure*}[thbp]
\centering
\subfloat{
\hspace{-0.2cm}
\begin{minipage}[t]{1\textwidth}
   \centering
  \includegraphics[angle=0,width=1\textwidth]{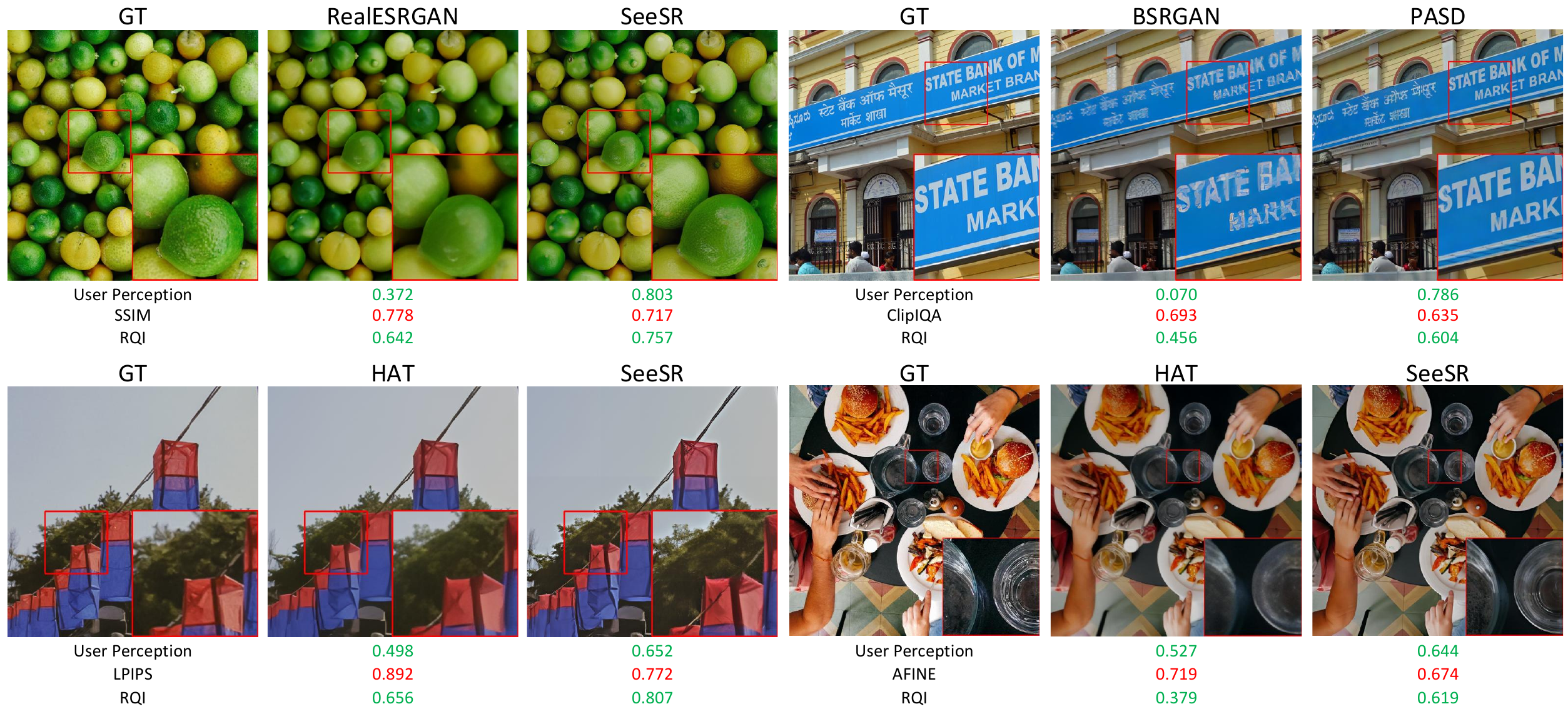}
\end{minipage}
}
\vspace{-0.2cm}
\caption{We show different cases where existing metrics fail. As a comparison, RQI handles all the cases correctly. All scores are normalized to [0,1] for easier comparisons. Please zoom in for better view.}
\vspace{-12pt}
\label{fig3}
\end{figure*}

From Table \ref{tab2} we make several observations. First, there exists a consistent SRCC improvement on all four testing sets by training under the RQI scheme, regardless of which IQA model or training set is applied. This indicates the effectiveness of the proposed scheme. Second, by calculating the average improvement for each testing dataset, the improvements on DIV2K, RealSR, DRealSR, and Set5\&Set14 are 0.077, 0.085, 0.146, and 0.138, respectively. The relatively larger improvements on DRealSR and Set5\&Set14 can be attributed to the limited quality of their GT images, which validates the design of our framework in addressing Goal~2. Third, among the three selected IQA models, we observe MANIQA \cite{yang2022maniqa} achieves better consistency, while among all training sets, models trained on PIPAL \cite{pipal} perform better. We attribute this to PIPAL’s inclusion of diverse distortion types with subtle perceptual variations. Thus, we select the best-performing model $\text{MANIQA}_{RQI} $ trained on the PIPAL dataset for analysis in the following experiments.

\subsection{Comparison with Existing Metrics}

We perform the same analysis as in Section \ref{sec3.2} to compare the proposed metric with existing image quality metrics. 
The results are shown in the last column in Table \ref{tab1}, where RQI achieves the best consistency with human perception in most cases, demonstrating its effectiveness in addressing different challenging aspects in SR evaluations.

\begin{figure*}[t]
\centering
\begin{minipage}[b]{0.12\textwidth}
    \centering
    \includegraphics[width=\linewidth]{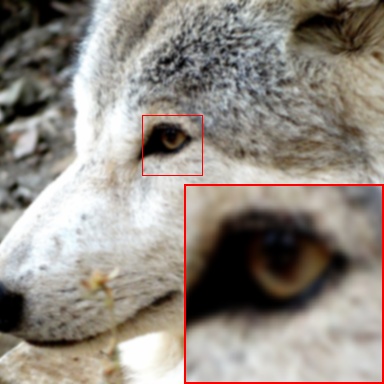}
\end{minipage}
\begin{minipage}[b]{0.12\textwidth}
    \centering
    \includegraphics[width=\linewidth]{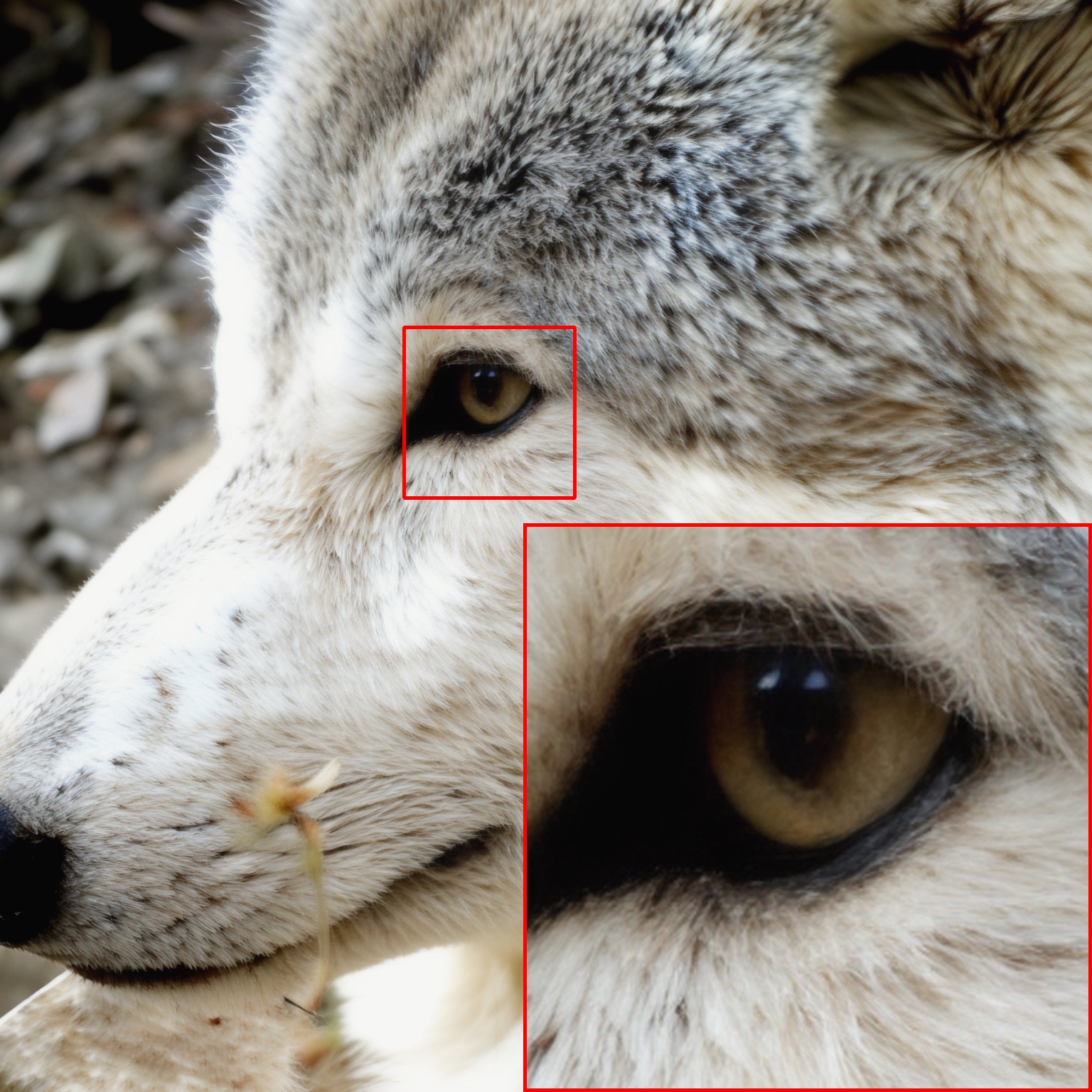}
\end{minipage}
\begin{minipage}[b]{0.12\textwidth}
    \centering
    \includegraphics[width=\linewidth]{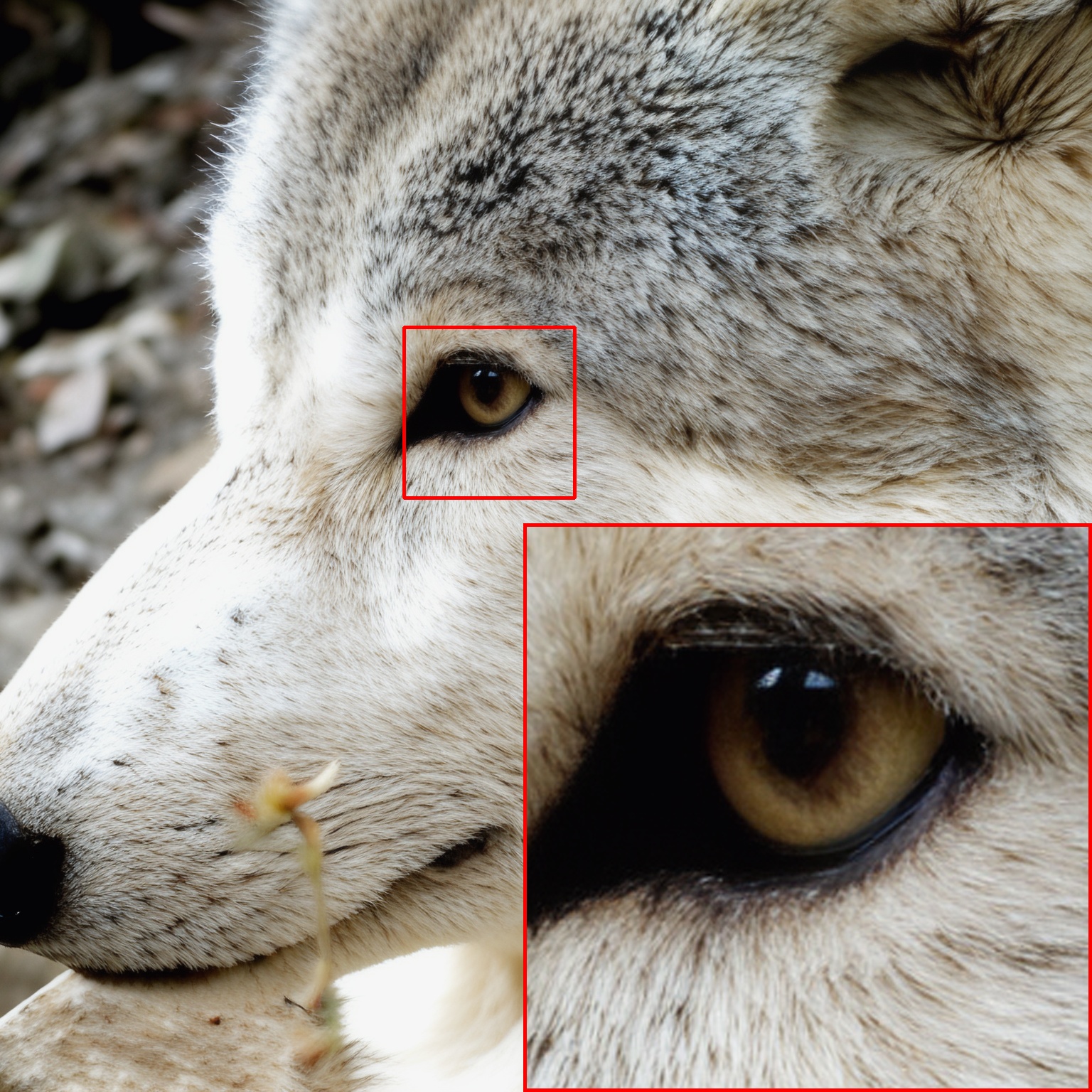}
\end{minipage}
\begin{minipage}[b]{0.12\textwidth}
    \centering
    \includegraphics[width=\linewidth]{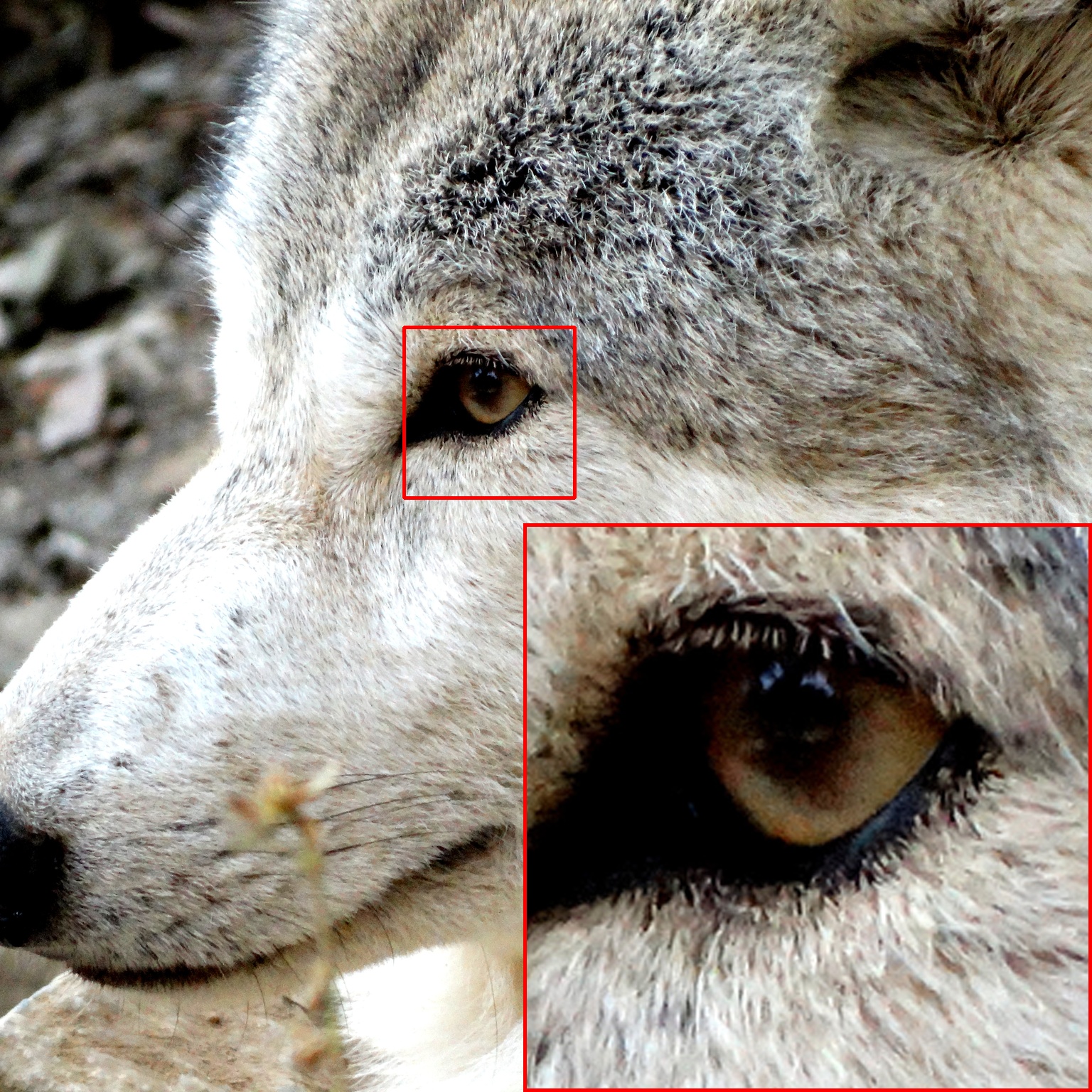}
\end{minipage}
\begin{minipage}[b]{0.12\textwidth}
    \centering
    \includegraphics[width=\linewidth]{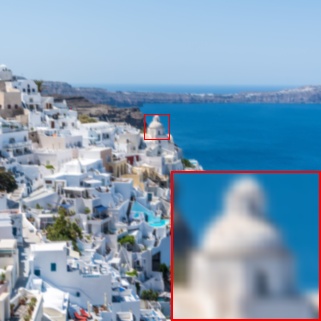}
\end{minipage}
\begin{minipage}[b]{0.12\textwidth}
    \centering
    \includegraphics[width=\linewidth]{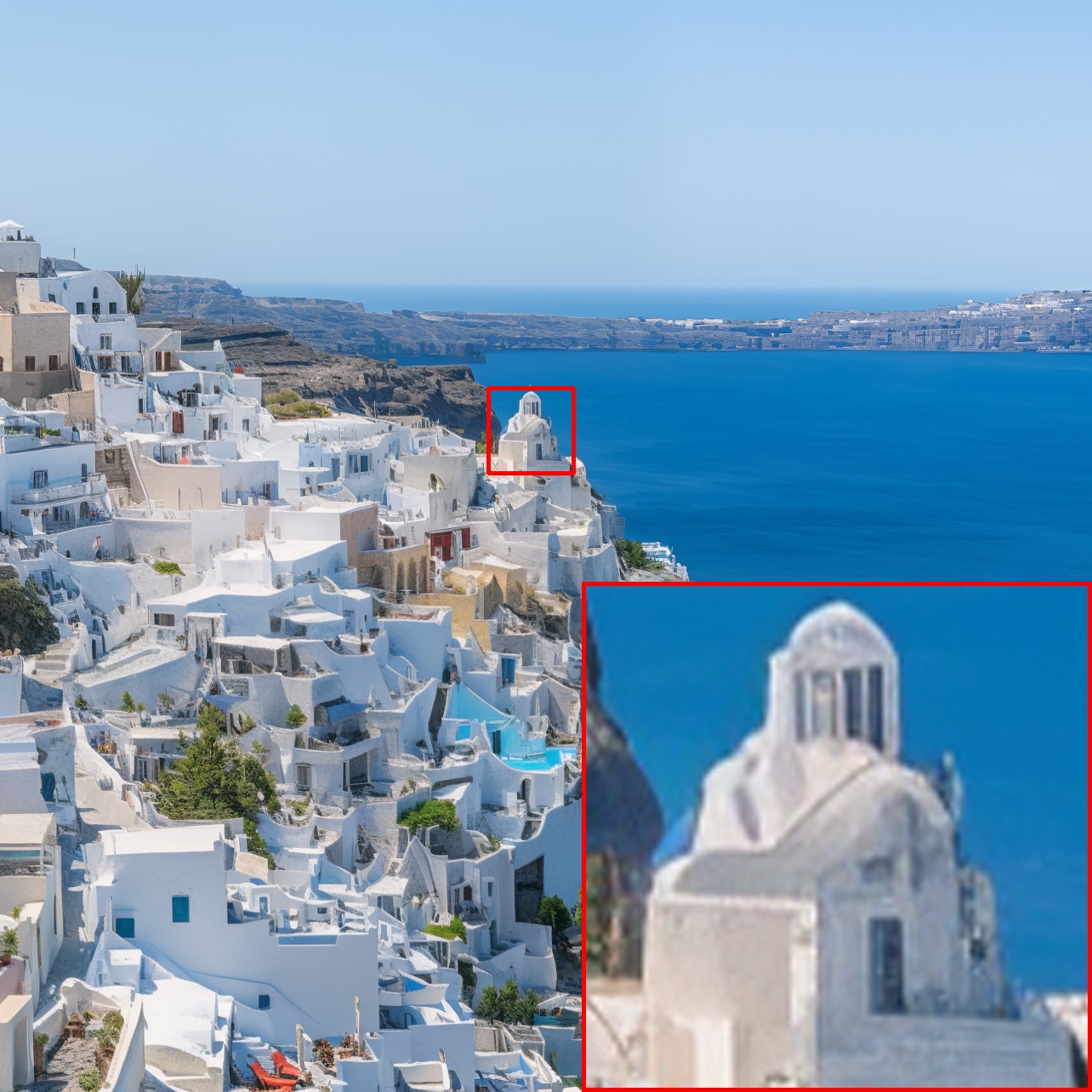}
\end{minipage}
\begin{minipage}[b]{0.12\textwidth}
    \centering
    \includegraphics[width=\linewidth]{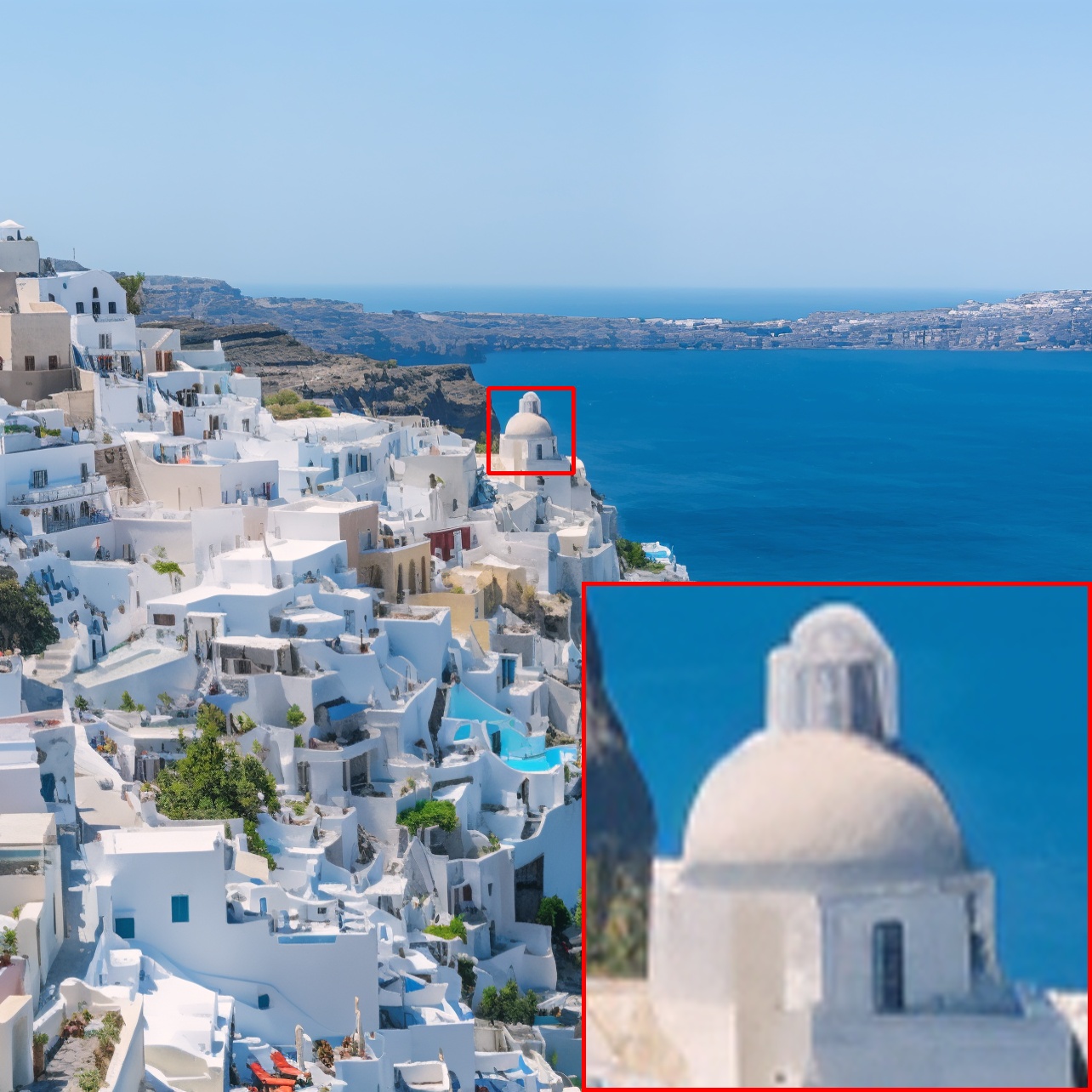}
\end{minipage}
\begin{minipage}[b]{0.12\textwidth}
    \centering
    \includegraphics[width=\linewidth]{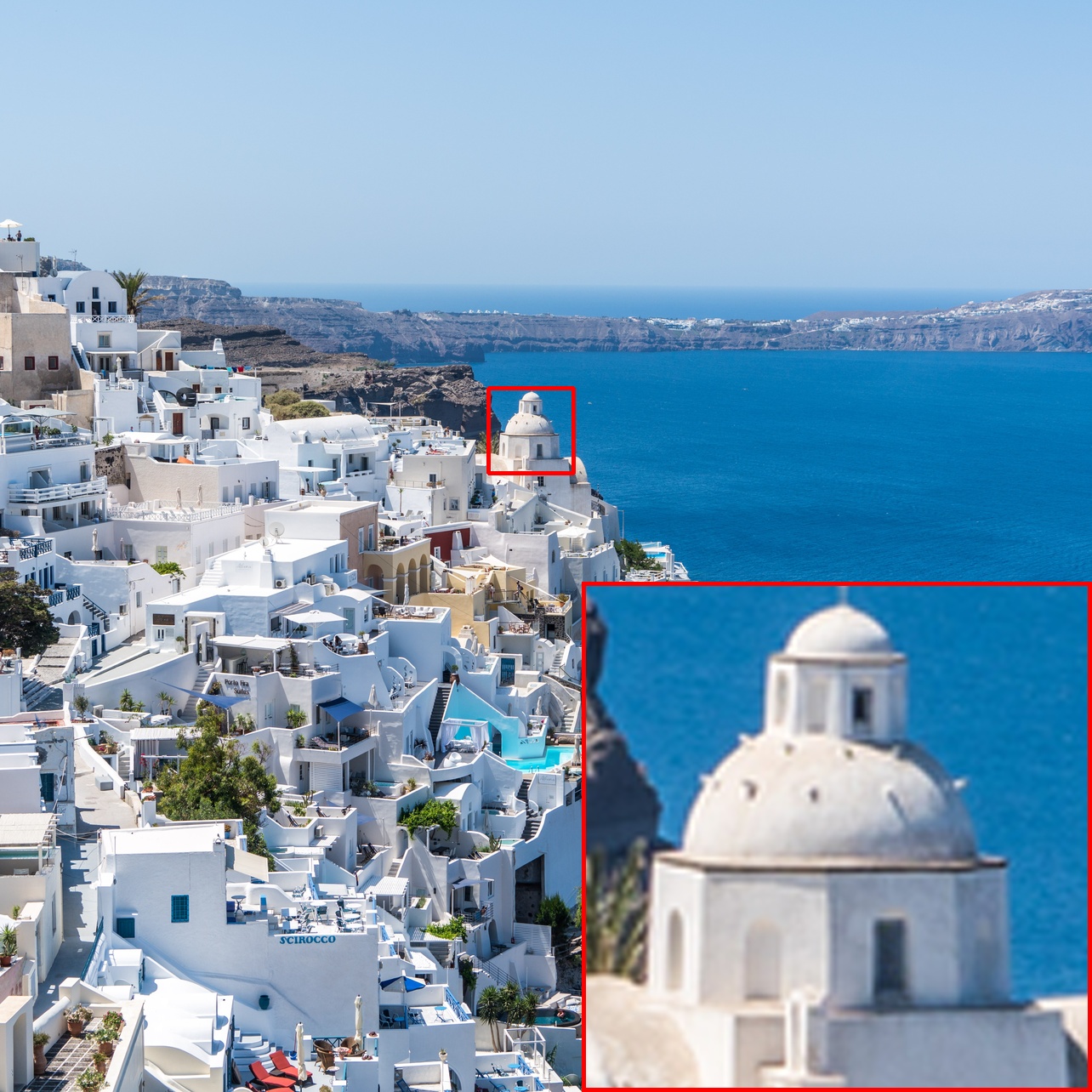}
\end{minipage}


\begin{minipage}[b]{0.12\textwidth}
    \centering
    \includegraphics[width=\linewidth]{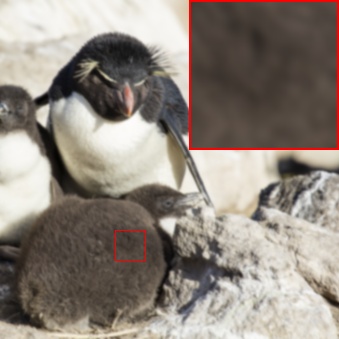}\\[-5pt]{\scriptsize LR}
\end{minipage}
\begin{minipage}[b]{0.12\textwidth}
    \centering
    \includegraphics[width=\linewidth]{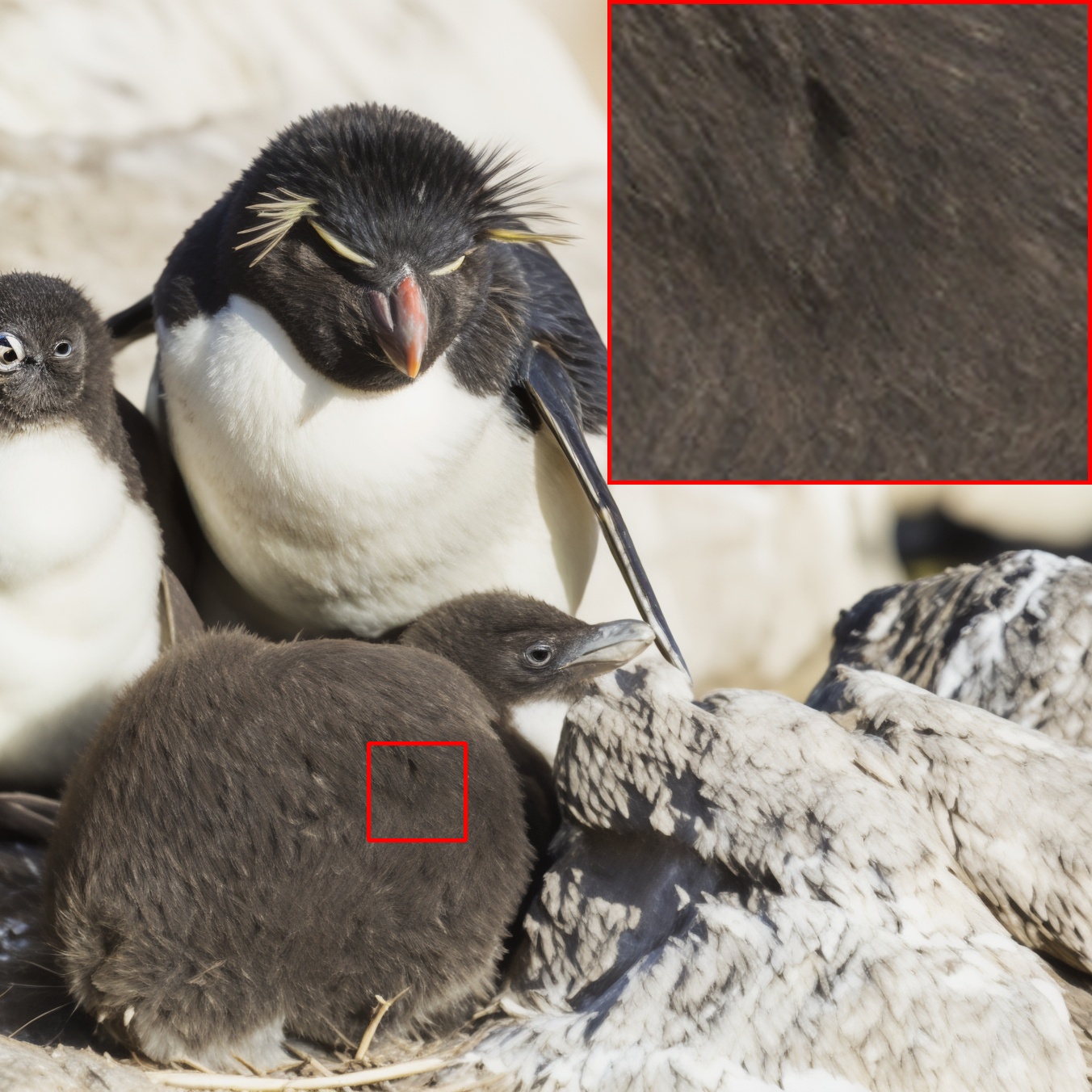}\\[-5pt]{\scriptsize PiSA-SR}
\end{minipage}
\begin{minipage}[b]{0.12\textwidth}
    \centering
    \includegraphics[width=\linewidth]{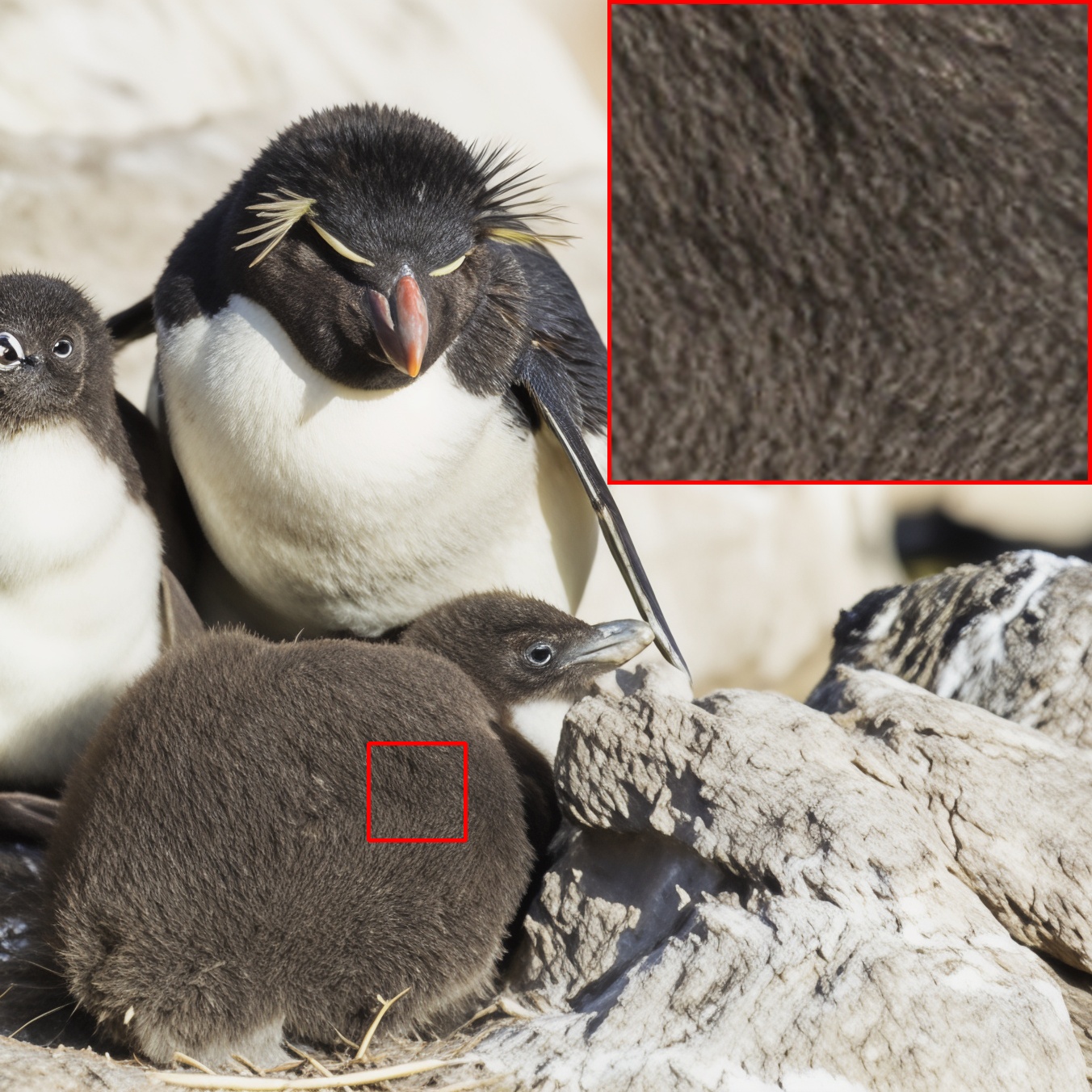}\\[-5pt]{\scriptsize PiSA-SR+RQI}
\end{minipage}
\begin{minipage}[b]{0.12\textwidth}
    \centering
    \includegraphics[width=\linewidth]{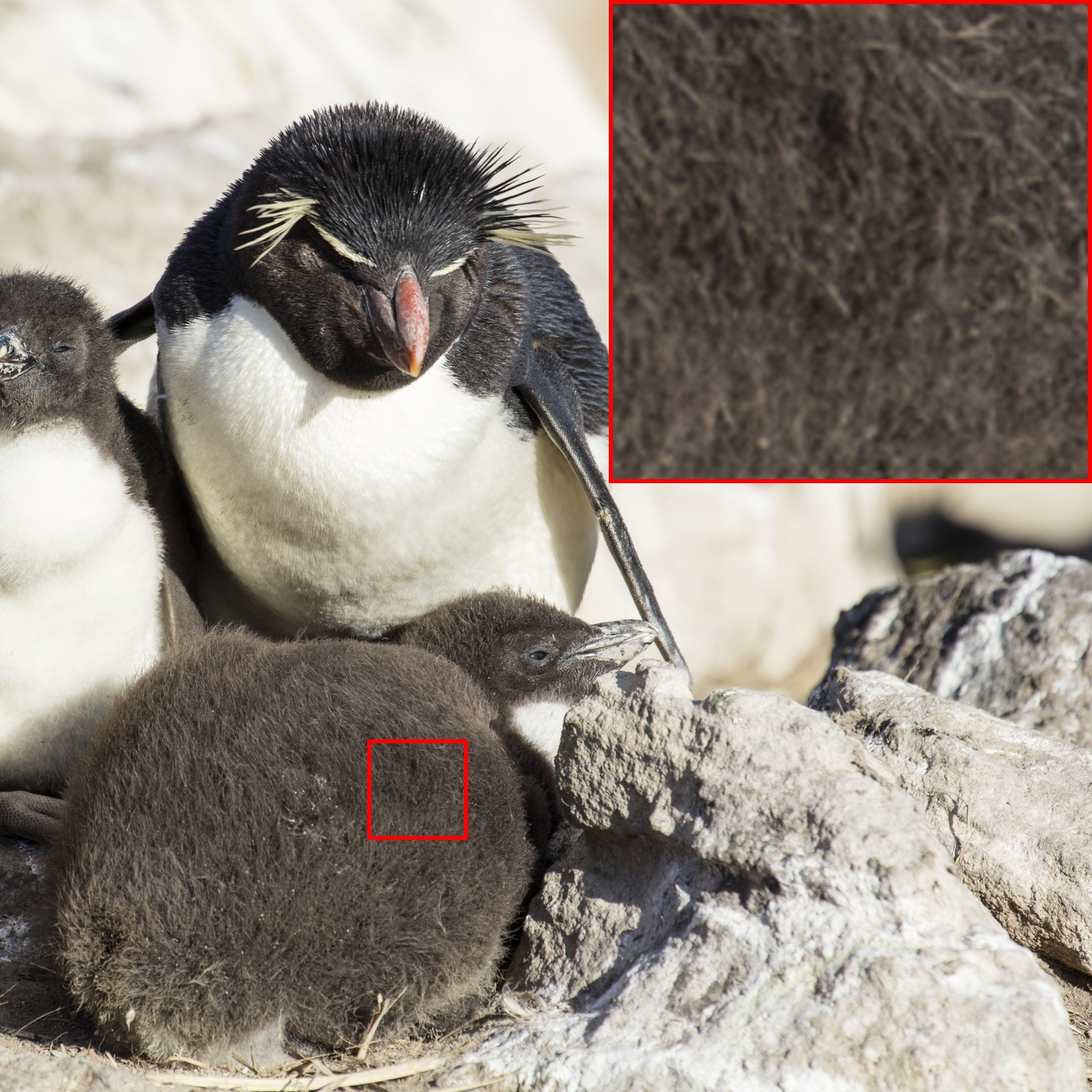}\\[-5pt]{\scriptsize GT}
\end{minipage}
\begin{minipage}[b]{0.12\textwidth}
    \centering
    \includegraphics[width=\linewidth]{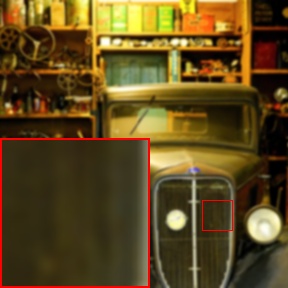}\\[-5pt]{\scriptsize LR}
\end{minipage}
\begin{minipage}[b]{0.12\textwidth}
    \centering
    \includegraphics[width=\linewidth]{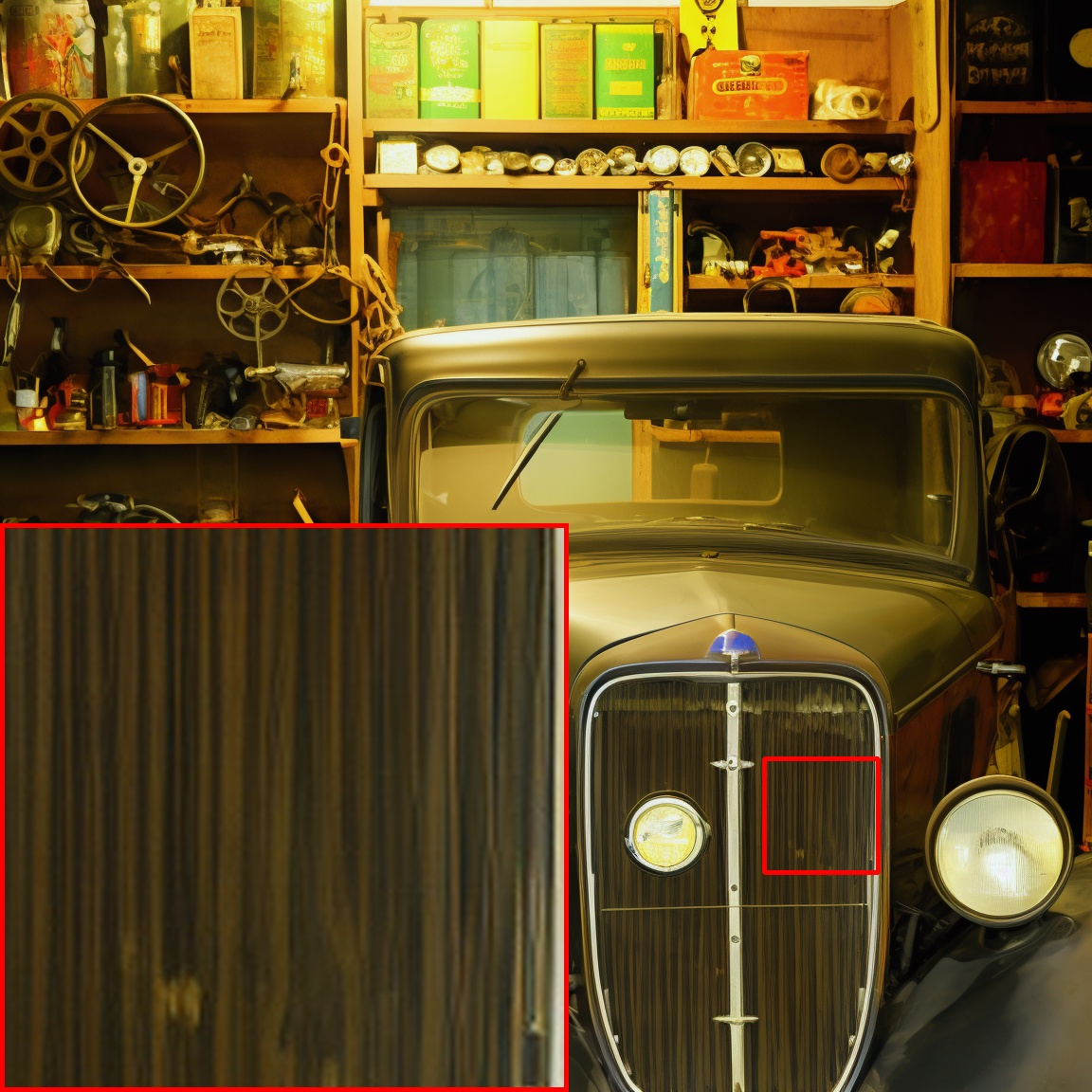}\\[-5pt]{\scriptsize SeeSR}
\end{minipage}
\begin{minipage}[b]{0.12\textwidth}
    \centering
    \includegraphics[width=\linewidth]{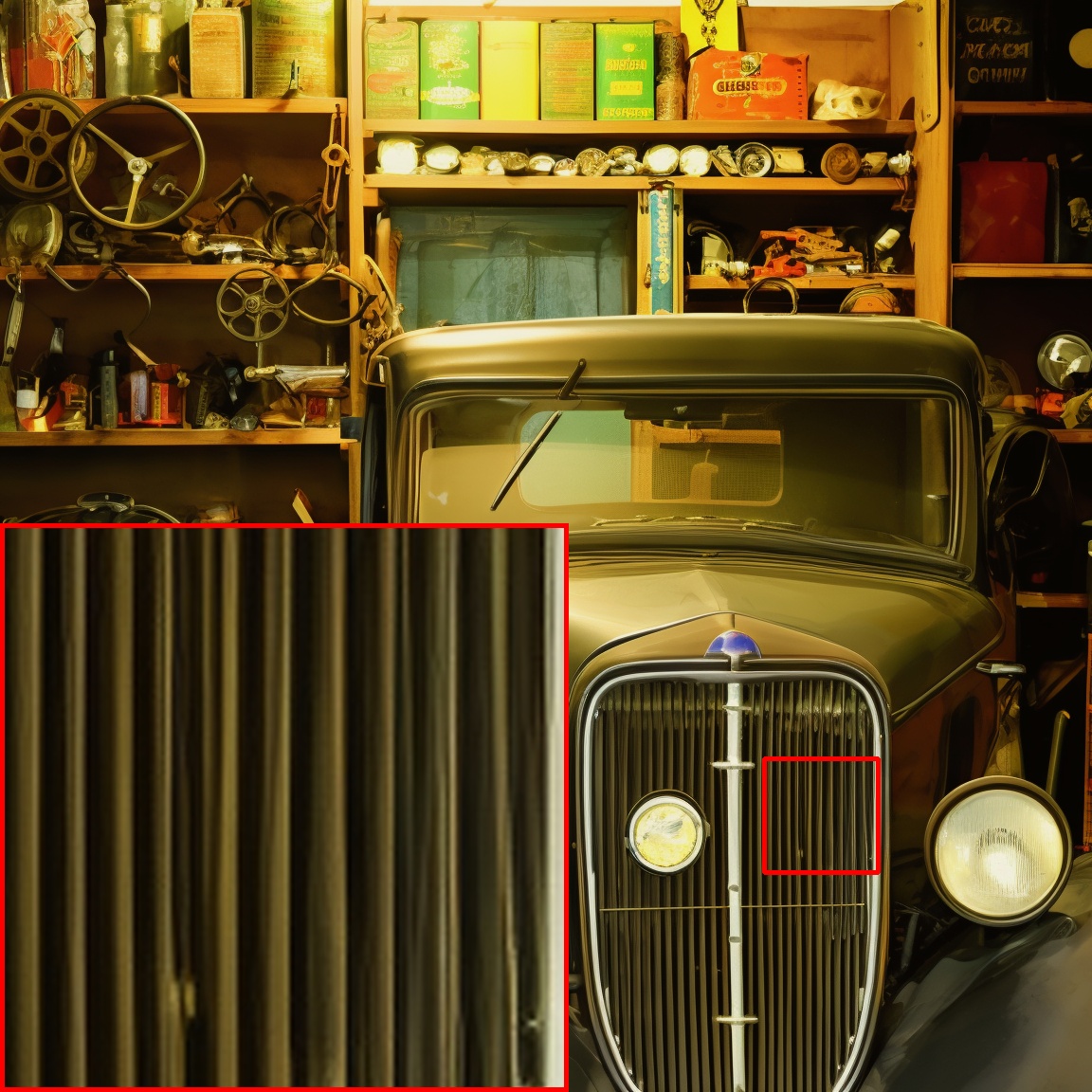}\\[-5pt]{\scriptsize SeeSR+RQI}
\end{minipage}
\begin{minipage}[b]{0.12\textwidth}
    \centering
    \includegraphics[width=\linewidth]{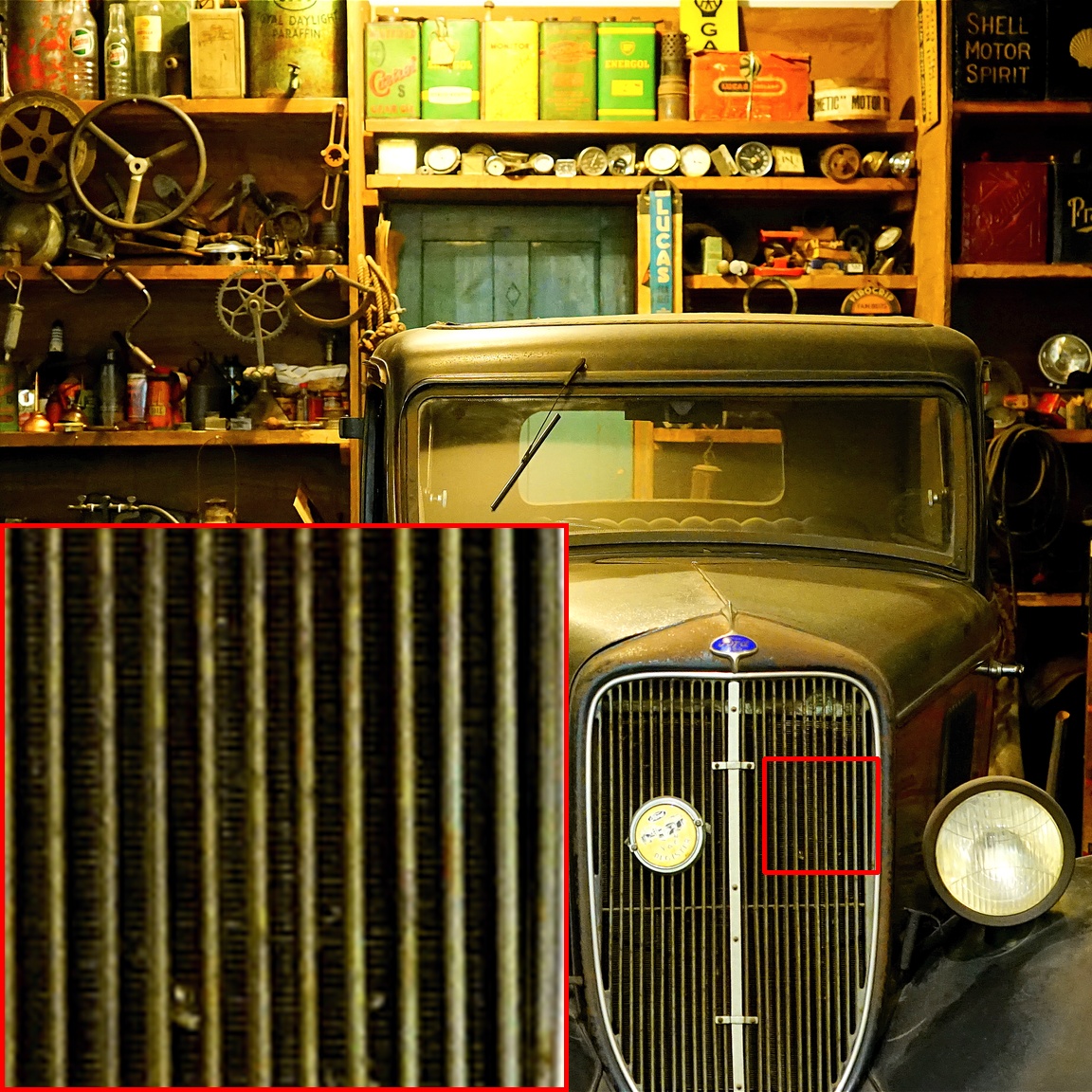}\\[-5pt]{\scriptsize GT}
\end{minipage}

\vspace{-4pt}
\caption{Visual comparison of training advancing SR models with RQI metric as an auxiliary loss. Please zoom in for a better view.}
\label{fig4}
\vspace{-10pt}
\end{figure*}

\begin{table}[t]
\centering
\setlength{\tabcolsep}{4pt}
\renewcommand\arraystretch{1.3}
\caption{Quantitative comparisons between baseline SR methods, training them using AFINE \cite{AFINE}, and using RQI as auxiliary loss.}
\vspace{-2mm}
\resizebox{0.48\textwidth}{!}{
\footnotesize
\begin{tabular}{c|l|ccccc}
\bottomrule[0.12em]
\rowcolor{tableHeadGray}
Dataset                & \multicolumn{1}{c|}{Model}          & SSIM$\uparrow$   & PSNR$\uparrow$    & LPIPS$\downarrow$  & Clip-IQA$\uparrow $ & DeQA$\uparrow $ \\ \hline \hline
\multirow{9}{*}{LSDIR}                            & SwinIR         & 0.6337          & 21.4834          & 0.2417          & 0.6254          & 4.0156      \\
                                                  & SwinIR+AFINE  & 0.6323          & \textbf{21.4842} & 0.2416          & \textbf{0.6256} & 4.0600      \\
                                                  & SwinIR+RQI    & \textbf{0.6345} & 21.4802          & \textbf{0.2414} & 0.6252          & \textbf{4.0762}      \\ \cline{2-7} 
                                                  & SeeSR          & 0.5888          & 20.9827          & 0.2497          & 0.6364          & 4.1650      \\
                                                  & SeeSR+AFINE   & 0.5774          & 21.1804          & \textbf{0.2483} & 0.6469          & 4.1675      \\
                                                  & SeeSR+RQI     & \textbf{0.6041} & \textbf{21.3457} & 0.2513          & \textbf{0.6655} & \textbf{4.1726}      \\ \cline{2-7} 
                                                  & PiSA-SR        & 0.6175          & \textbf{21.2771} & 0.2382          & 0.6452          & 4.2127      \\
                                                  & PiSA-SR+AFINE & 0.6131          & 21.0324          & 0.2363          & 0.6481          & 4.2201      \\
                                                  & PiSA-SR+RQI   & \textbf{0.6190} & 21.0777          & \textbf{0.2320} & \textbf{0.6732} & \textbf{4.2520}      \\ \hline
\multirow{9}{*}{DIV2K}                            & SwinIR         & 0.6635          & 23.6498          & 0.2842          & \textbf{0.6032} & 3.9531      \\
                                                  & SwinIR+AFINE  & 0.6620          & 23.6535          & 0.2829          & 0.6018          & 3.9725      \\
                                                  & SwinIR+RQI    & \textbf{0.6667} & \textbf{23.7056} & \textbf{0.2824} & 0.6006          & \textbf{3.9790}      \\ \cline{2-7} 
                                                  & SeeSR          & 0.6307          & 22.9572          & 0.2889          & 0.6337          & 4.1087      \\
                                                  & SeeSR+AFINE   & 0.6375          & 23.1074          & 0.2850          & 0.6496          & 4.1225      \\
                                                  & SeeSR+RQI     & \textbf{0.6452} & \textbf{23.3868} & \textbf{0.2839} & \textbf{0.6543} & \textbf{4.1456}      \\ \cline{2-7} 
                                                  & PiSA-SR        & \textbf{0.6664} & \textbf{23.8365} & 0.2551          & 0.6347          & 4.0892      \\
                                                  & PiSA-SR+AFINE & 0.6544          & 23.1372          & 0.2640          & 0.6506          & 4.1352      \\
                                                  & PiSA-SR+RQI   & 0.6646          & 23.5346          & \textbf{0.2539} & \textbf{0.6572} & \textbf{4.1522}      \\               \bottomrule
\end{tabular}}
\label{tab4}
\vspace{-12pt}
\end{table}

We further compare the metrics on three SR-IQA benchmarks BSD-SR \cite{ma2017learning}, QADS \cite{zhou2022quality}, SRIQA-Bench \cite{AFINE}, and one general IQA dataset Kadid-10K \cite{lin2019kadid}. 
We report both mean consistency over each source image and overall consistency. DISTS \cite{DISTS}, AFINE \cite{AFINE} and DeQA-Score \cite{deqa} are not evaluated on the Kadid-10K dataset since the models are fine-tuned based on the dataset. We show the results in Table \ref{tab3}, where RQI perform stably across all datasets and criteria. Although we train RQI only by relative quality discrepancy, it achieves good consistency across image contents, on different SR scales, and even on the general IQA task. All the results validate the robustness of the RQI metric. Also note that SSIM \cite{ssim} and PSNR only perform well when evaluating early SR models (e.g., on the BSD-SR and QADS datasets), further demonstrating the necessity of developing updated evaluation protocols to keep pace with the evolution of SR models.

\subsection{Qualitative analysis}
\label{sec5.5}

Figure \ref{fig3} shows how RQI performs robustly in different aspects. The distortion FR-IQA metric SSIM \cite{ssim} (the up-left case) favors blur regions (RealESRGAN) more than textures (SeeSR) on the lemon surface. NR-IQA metric ClipIQA \cite{wang2023exploring} (the up-right case) fail to evaluate character consistency for BSRGAN and PASD due to the lack of reference. Perceptual FR-IQA metric LPIPS \cite{LPIPS} fail (the bottom-left case) due to details in GT are vague and HAT output more resembles it. 
Asymmetric method AFINE \cite{AFINE} (the bottom-right case) struggles to distinguish images with subtle quality differences.
In contrast, the proposed RQI is perceptually accurate, aware of semantic consistency, avoids biased evaluations caused by unreliable GTs, and performs robustly for fine-grained quality differences.

\subsection{Training SR Models with RQI as a Loss}

Since RQI provides a unified regression objective towards better perceptual quality, we further optimize SOTA SR models using our RQI based metric as an auxiliary loss. Specifically, we re-train three SOTA SR models SwinIR \cite{SwinIR}, SeeSR \cite{SeeSR} and PiSA-SR \cite{Pisasr} on the LSDIR \cite{lsdir} training set \textit{from scratch}, by modifying the objective function as following:

\begin{equation}
\mathcal{L}=\mathcal{L}_{ori}+\lambda\cdot s
\label{eq:eval}
\end{equation}

\noindent
where $\mathcal{L}_{ori}$ denotes the loss function used in the baseline settings of the three SR models, and $s$ represents the RQI score. The weighting factor $\lambda$ is empirically set to $-0.1$ to ensure scaling consistency. For comparison, we retrain the SR models using only $\mathcal{L}_{ori}$ as the sole optimization objective. We also compare models by training them using the AFINE \cite{AFINE} metric as an auxiliary loss following the same setting as training with RQI.

We evaluate the models on the validation sets of LSDIR and DIV2K-wild, and present the quantitative results in Table \ref{tab4}, from which several observations can be made.
First, RQI achieves the best overall perceptual performance, demonstrating not only its effectiveness in SR evaluation but also its capacity to enhance SR model generation quality.
Second, the performance gain observed on SwinIR is relatively limited, likely due to the weaker generative capability of its non-diffusion architecture --- suggesting that such loss functions are more beneficial for models with stronger generative abilities.
Finally, we also observe that
training SR models with AFINE leads to certain improvement against the baseline, as it also follows the asymmetric design that can provide more consistent optimization directions regardless of lower quality GT.

We further provide qualitative comparisons in Figure \ref{fig4}. As illustrated, integrating RQI into the training process enables SR models to generate more realistic and visually appealing details (e.g., sharper wolf fur and finer penguin feathers on the left), while simultaneously enhancing structural fidelity (e.g., the preserved dome geometry and the regularized grille bars on the right). These observations confirm that RQI-driven optimization not only improves perceptual realism but also enforces structural consistency, leading to higher-fidelity super-resolved results.
More comparisons can be found in the Supplementary Material.

\section{Conclusion}\label{sec6}

In this work, we revisited the reliability of existing image quality metrics in evaluating modern SR models and revealed their limited consistency with human perceptual preferences. Through comprehensive analysis, we identified key challenges in SR evaluation, including imperfect references, fidelity measurement inconsistencies, and subtle perceptual variations among high-quality results. To overcome these issues, we introduced a general framework, RQI, which reformulates SR evaluation as a relative quality comparison task. RQI can be seamlessly integrated with existing IQA models, enhancing their robustness across diverse datasets and improving their perceptual alignment. Furthermore, when used as a training objective, RQI effectively guides SR models toward generating more realistic and structurally faithful results. 
We believe this framework provides a promising step toward more perceptually consistent and generalizable SR evaluation and optimization.
Since RQI is not trained on SR-specific data, it is expected to generalize to other restoration tasks such as deblurring \cite{nah2017deep}, denoising \cite{abdelhamed2018high}, and deraining \cite{yang2017deep}. Exploring these extensions remains an interesting direction for future work.

\section*{Acknowledgement}
This work was supported by grant PID2021-128178OB-I00, PID2024-162555OB-I00, funded by MCIN/AEI/10.13039/501100011033 and ERDF "A way of making Europe", the CERCA Program from Generalitat de Catalunya, the grant Càtedra ENIA UAB-Cruïlla (TSI-100929-2023-2) from the Ministry of Economic Affairs and Digital Transformation of Spain, and the 2025 Leonardo Grant for Scientific Research and Cultural Creation from the BBVA Foundation. The BBVA Foundation accepts no responsibility for the opinions, statements and contents included in the project and/or the results thereof, which are entirely the responsibility of the authors. 
Shaolin Su was supported by the HORIZON MSCA Postdoctoral Fellowship funded by the European Union (project number 101152858). 
David Serrano-Lozano was supported by the FPI grant from Spanish Ministry of Science and Innovation (PRE2022-101525).
Lei Sun was partially funded by the Ministry of Education and Science of Bulgaria’s support for INSAIT as part of the Bulgarian National Roadmap for Research Infrastructure and by the European Union's Horizon Europe -- the Framework Programme for Research and Innovation, under grant agreement 101168521.

{
    \small
    \bibliographystyle{ieeenat_fullname}
    \bibliography{main}
}

\input{supp}

\end{document}

%% file: preamble.tex
\usepackage{colortbl}
\usepackage{diagbox}
\usepackage{adjustbox}
\usepackage{multirow}
\usepackage{subcaption}
\usepackage{mwe} 
\usepackage{dsfont}

\definecolor{tableHeadGray}{gray}{.9}
\definecolor{tableSubHeadGray}{gray}{0.95}

\usepackage{arydshln} 

%% file: supp.tex
\clearpage
\setcounter{page}{1}
\setcounter{section}{0}
\renewcommand{\thesection}{\Alph{section}}
\maketitlesupplementary

The supplementary material includes the following additional information:

\begin{enumerate}[Sec. A]

\item \hyperref[secA]{Details and more analysis of the user experiment.}
\item \hyperref[secB]{Implementation details of RQI.}
\item \hyperref[secC]{Ablations on different training and testing settings of the proposed RQI scheme.}
\item \hyperref[secD]{Additional visual comparisons for RQI evaluation.}
\item \hyperref[secE]{Additional details and results when training with RQI.}
\item \hyperref[secF]{Limitations and Discussions.}
\end{enumerate}

\section{User Experiment}
\label{secA}

\subsection{User Study Details}
Since image quality comparison requires fine-grained perceptual judgment, we conduct the user study under a strictly controlled environment, unlike some prior user studies \cite{wu2023human, xu2023imagereward}. Specifically, the study was performed in a matte dark room, where the display served as the only light source. All images were shown on a 3K-resolution monitor that had been calibrated to the sRGB color space. Before the experiment, all participants received detailed instructions to the visual comparison task. 
During each trial, two images from the same LR source were presented side-by-side, and participants were asked to choose the one with better perceptual quality. The order and left–right placement of each pair were randomized to minimize positional bias. For the DRealSR dataset, because the HR images exceed 4K resolution, only center-cropped images are selected for all the evaluations to prevent scaling artifacts.
Each image pair received ratings from at least 15 participants, all of whom passed the Ishihara color-vision test.

\subsection{Analysis of User Study Results}

\begin{figure*}[thbp]
\centering
\subfloat{
\hspace{-0.4cm}
\label{fig:subfig_5a}
\begin{minipage}[t]{0.25\textwidth}
   \centering
  \includegraphics[angle=0,height=0.9\textwidth]{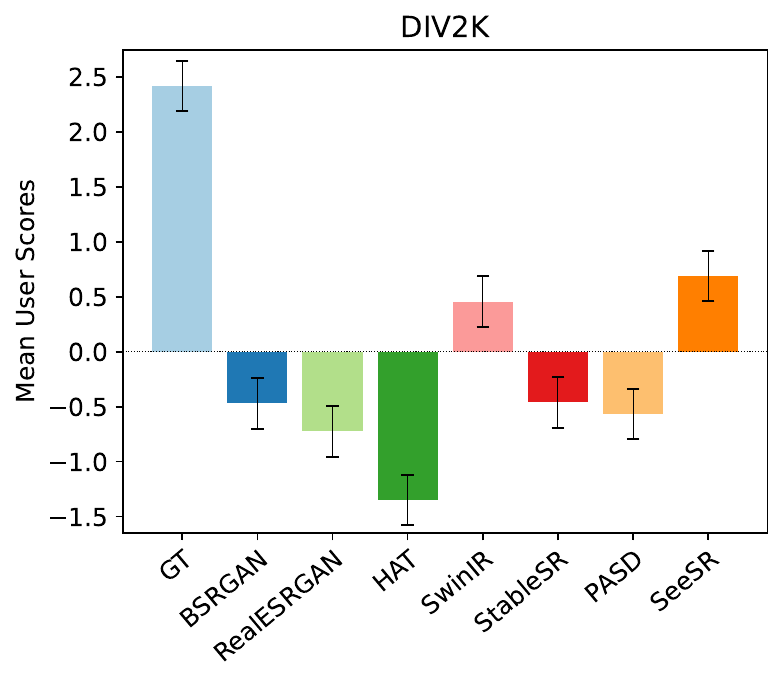}
\end{minipage}
}
\subfloat{
\hspace{-0.26cm}
\label{fig:subfig_5b}
\begin{minipage}[t]{0.25\textwidth}
   \centering
  \includegraphics[angle=0,height=0.9\textwidth]{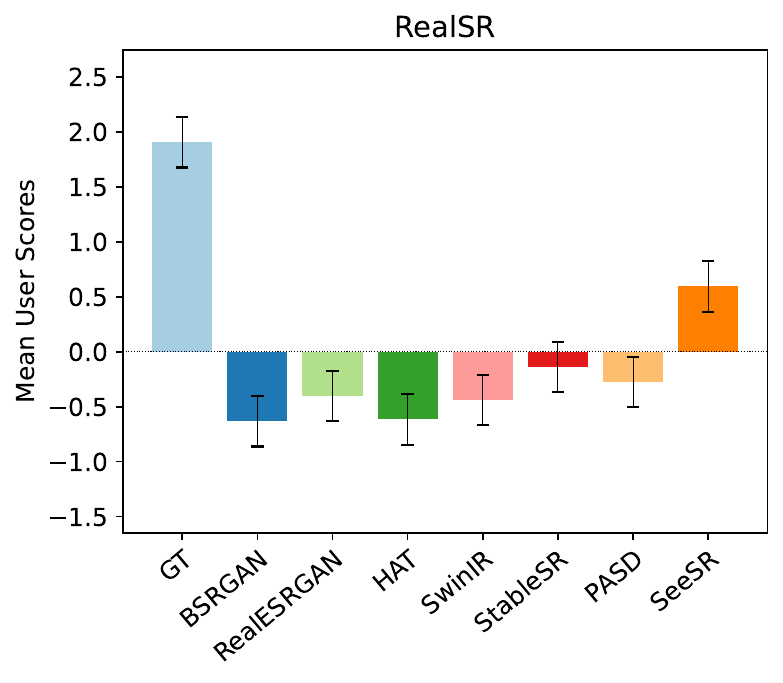}
\end{minipage}
}
\subfloat{
\hspace{-0.26cm}
\label{fig:subfig_5c}
\begin{minipage}[t]{0.25\textwidth}
   \centering
  \includegraphics[angle=0,height=0.9\textwidth]{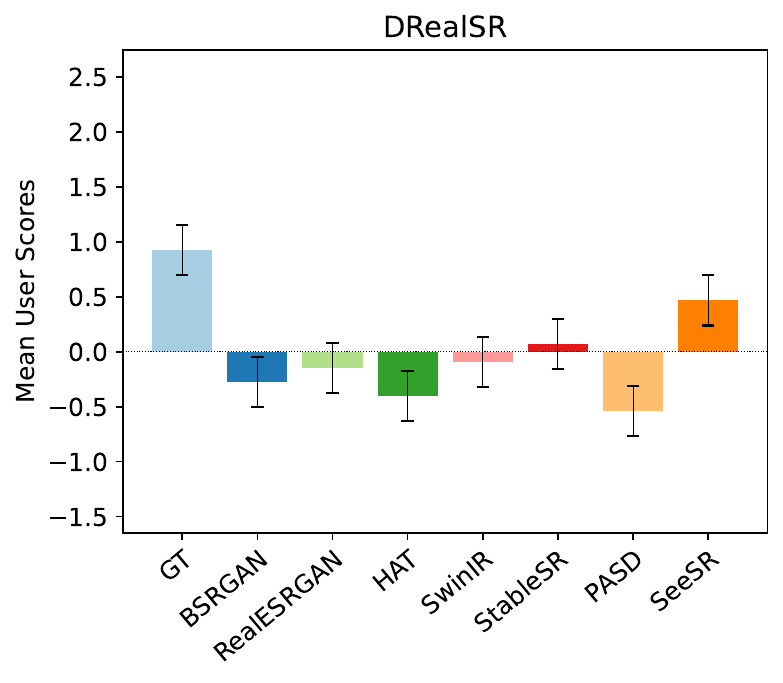}
\end{minipage}
}
\subfloat{
\hspace{-0.26cm}
\label{fig:subfig_5d}
\begin{minipage}[t]{0.25\textwidth}
   \centering
  \includegraphics[angle=0,height=0.9\textwidth]{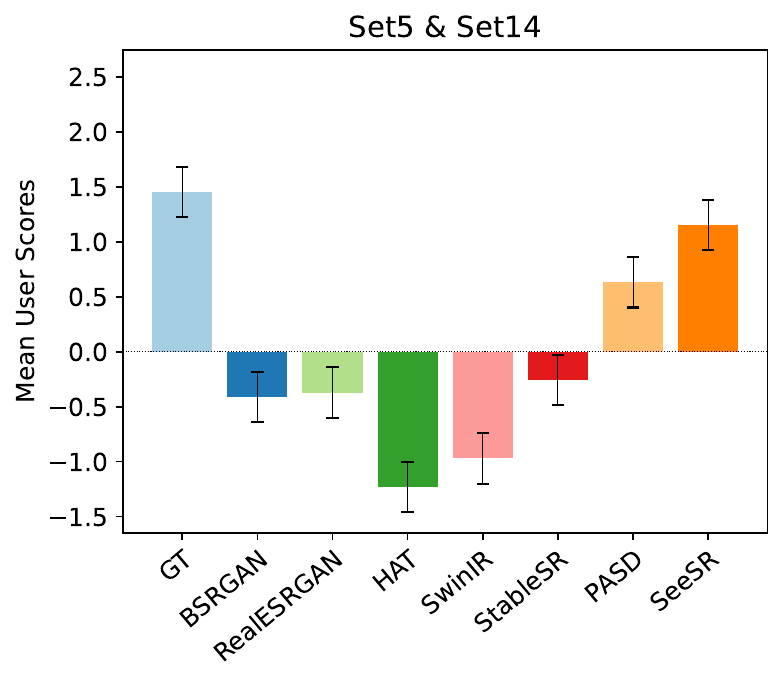}
\end{minipage}
}
\caption{Average user scores on different SR models (including GT) in four SR testing datasets. Error bars correspond to a 95\% confidence interval.}
\label{fig5}
\end{figure*}

\begin{figure}[t]
\centering
\includegraphics[width=\linewidth]{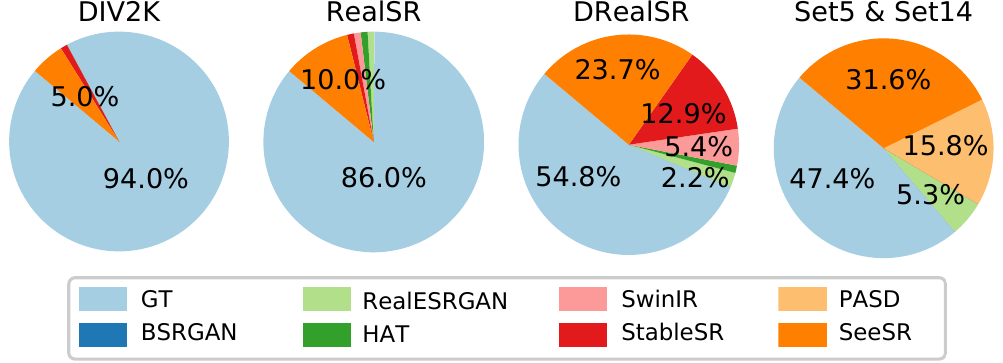}
\caption{User statistics of the best quality HR image in four SR datasets.}
\label{fig6}
\end{figure}

In this subsection, we conduct a brief analysis of the user results. Specifically, we analyze the overall perceptual preferences among the evaluated SR models and the distribution over user preferences. Figure~\ref{fig5} presents the average Thurstone scores across all datasets, and Figure~\ref{fig6} presents the proportion of best-ranked images for each evaluated model. Since Set5~\cite{bevilacqua2012low} and Set14~\cite{zeyde2012single} share similar content and contain only a limited number of images, we combine them for the analysis. Several noteworthy observations can be drawn from the statistics.

\begin{figure*}[th]
\centering
\subfloat[GT]{
\begin{minipage}[t]{0.25\textwidth}
   \centering
  \includegraphics[angle=0,width=1\textwidth]{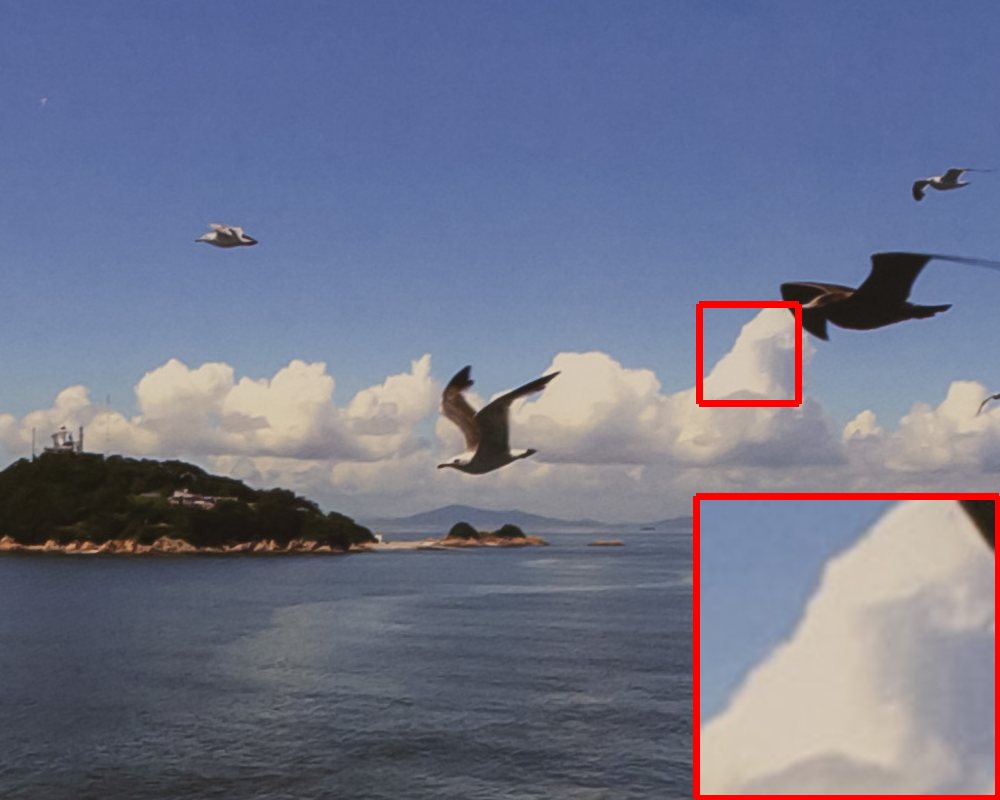}
\end{minipage}
}
\subfloat[SeeSR]{
\begin{minipage}[t]{0.25\textwidth}
   \centering
  \includegraphics[angle=0,width=1\textwidth]{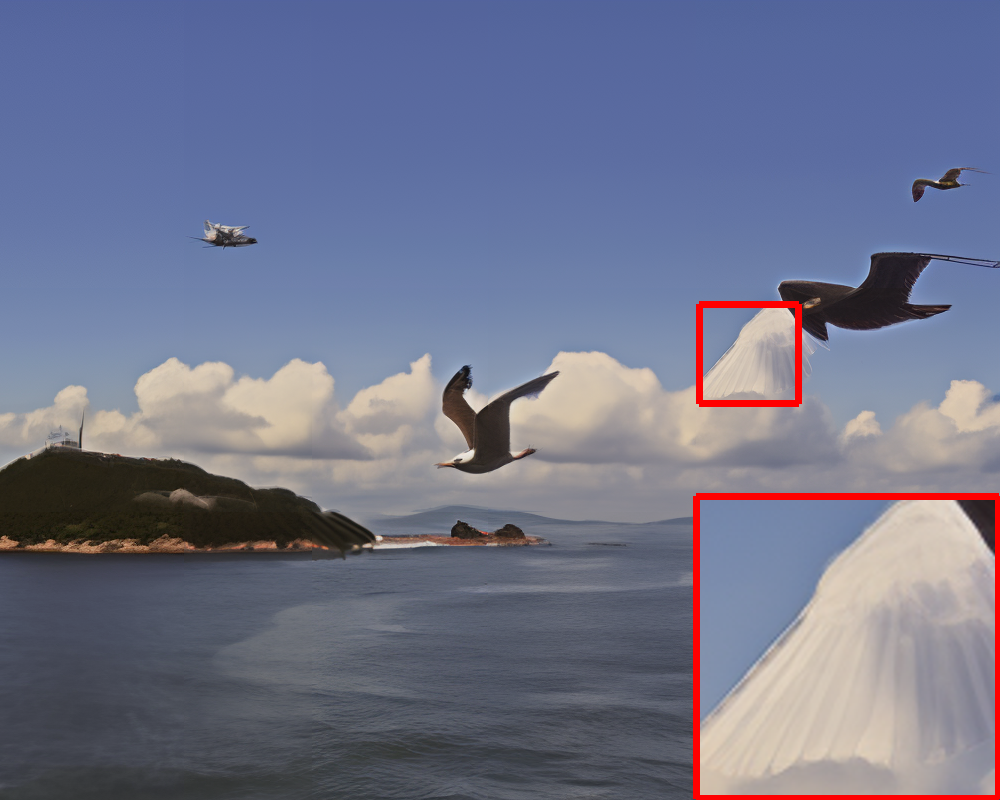}
\end{minipage}
}
\subfloat[HAT]{
\begin{minipage}[t]{0.25\textwidth}
   \centering
  \includegraphics[angle=0,width=1\textwidth]{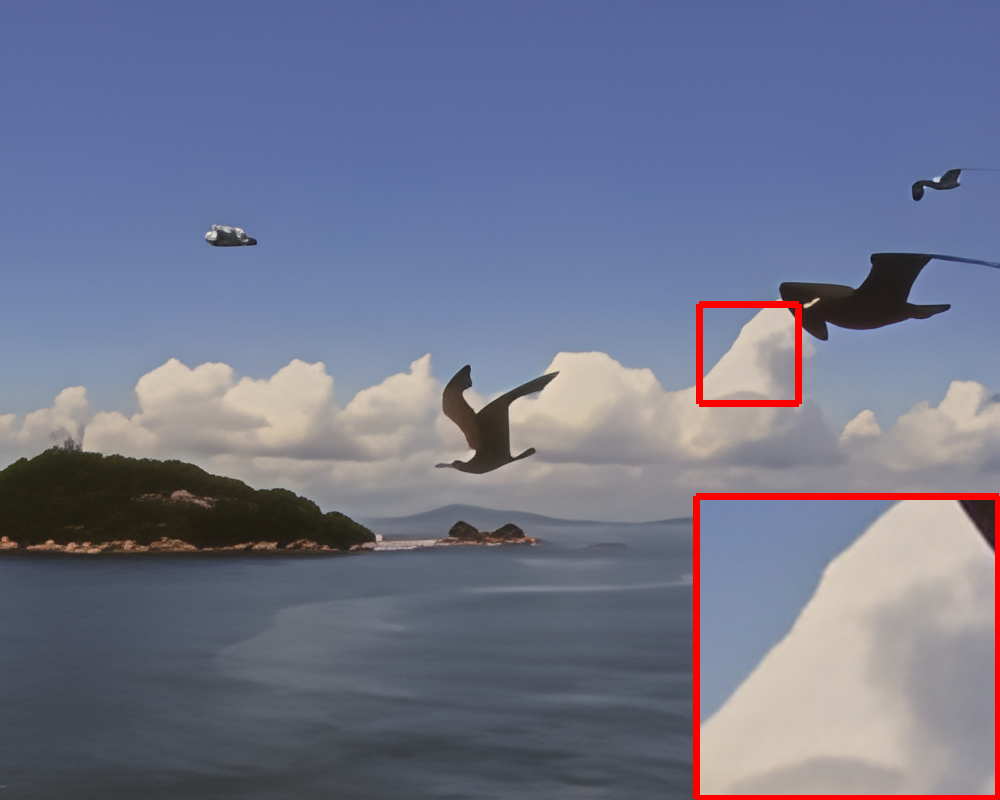}
\end{minipage}
}
\caption{We show that sometimes SeeSR \cite{SeeSR} produce hallucinations due to its strong generative capability inherited from tag-style prompts. Such degradation may exists in a small area, yet significantly affect the user perception of image quality.}
\label{fig7}
\end{figure*}

First, diffusion-based models are generally favored by human observers due to their strong ability to synthesize visually appealing and richly detailed textures. In most cases, these models deliver superior perceptual quality.
Second, although ground-truth (GT) images achieve the highest average scores overall, Figure~\ref{fig6} reveals that outputs from certain models can perceptually surpass their corresponding GTs. The proportion of such cases varies with dataset quality --- for example, only a small fraction in DIV2K~\cite{div2k}, but more than half in Set5 and Set14~\cite{bevilacqua2012low,zeyde2012single}.
Third, during the user study, we observe that localized hallucination artifacts produced by diffusion-based SR models can strongly influence human preference. Figure~\ref{fig7} illustrates an example from the RealSR~\cite{RealSR} dataset, where a white cloud is incorrectly synthesized as a seagull wing by the advanced SeeSR~\cite{SeeSR} model, partially driven by its strong generation capability inherited from tag-style prompts. Although the hallucination occurs in a small region, it substantially affects users’ subjective judgments. Such novel hallucination phenomena also introduce additional challenges for existing image quality metrics.

\subsection{More Analysis of Table \ref{tab1}}

Here we provide more analysis of the evaluation results for different metrics, shown in Table 1 from the original paper.

\begin{itemize}
    \item 
    Pixel-based metrics such as PSNR and SSIM \cite{ssim} exhibit negative correlations with human perceptual preference. This is primarily because these distortion-oriented measures tend to regress toward the average solution across multiple plausible reconstructions, a behavior inherent to the ill-posed nature of the SR problem. This mismatch reflects a fundamental contradiction in SR evaluation: metrics designed to favor pixel-wise fidelity fail to capture the perceptual realism that modern SR models aim to produce. Since this mismatch fundamentally affects the evaluation of almost all images, the two method performances appears consistently poor across all metrics.

    \item
    Although we identify the limitation of lacking reference for NR-IQA methods in making fair evaluations, such alteration of textures/structures may exist only in some of the images. This partly explains the overall good performance of NR metrics (NIQE \cite{NIQE}, PI \cite{blau20182018}, Clip-IQA \cite{wang2023exploring} and MANIQA \cite{yang2022maniqa}, \textit{etc}).

    \item
    We observe a substantial consistency improvement of RQI on the DIV2K dataset~\cite{div2k}. Since most GT images in DIV2K are of high quality, this improvement highlights RQI’s strong ability to distinguish fine-grained perceptual differences. Meanwhile, the strong consistency achieved on the DRealSR dataset~\cite{DRealSR} can be attributed to RQI’s robustness to imperfect GTs, since in this dataset SR models more frequently produce results that surpass the GTs in perceptual quality (see Figure~\ref{fig6}).

    \item 
    RQI performs slightly worse than DeQA-Score \cite{deqa} on Set5\&Set14, we attribute this to the limited resolutions of Set5 and Set14. Benefited from large-scale pretraining, LLM-based method DeQA-Score is less affected. Nevertheless, RQI still achieves better consistency than the other non-LLM metrics.

\end{itemize}

\section{Implementation Details}
\label{secB}

For the two FR-IQA models AHIQ \cite{lao2022attentions} and TOPIQ \cite{chen2024topiq}, we simply remove the last activation layer to ensure the models produce bidirectional outputs. Fot the NR-IQA model MANIQA \cite{yang2022maniqa}, we modify it to receive two images as input. Specifically, features extracted from the two inputs are concatenated before the transposed attention block to enable effective cross-image fusion.

During training, all the selected models are trained following their official implementations. For all the models, the learning rate is set to $10^{-4}$ with weight decay $10^{-5}$. The batch size is set to 4 for AHIQ \cite{lao2022attentions}, and 8 for MANIQA \cite{yang2022maniqa} and TOPIQ \cite{chen2024topiq}. AHIQ \cite{lao2022attentions} and MANIQA \cite{yang2022maniqa} randomly crop image patches with size 224, while TOPIQ \cite{chen2024topiq} randomly crop image patches with size 384. The crops are randomly flipped during training for augmentation. We split the datasets into training and validation subsets (8:2) with non-overlapping scenes, and select the best-performing models upon their performances on the validation set.

During testing, we randomly crop patches from the inputs, ensuring that each patch pair is taken from the same spatial region of the SR output and its corresponding GT. Since SR evaluation often involves high-resolution inputs, we assess perceptual quality at three different spatial scales. Specifically, we use the original resolution as well as images downsampled by factors of $\times 2$ and $\times 3$. At each scale, we randomly crop 20 patches and compute the final score by averaging across all patches. The downsampling and patch cropping are applied only to images whose resolutions exceed the required input size of the IQA models (\ie, 224 for AHIQ~\cite{lao2022attentions} and MANIQA~\cite{yang2022maniqa}, and 384 for TOPIQ~\cite{chen2024topiq}).

\section{Ablation Study}
\label{secC}

In this section, we show more ablation results of the proposed RQI scheme. Since we propose training RQI with image pairs that contain arbitrary distortions, the images contain distortions across types and levels. Therefore, we compare training RQI with image pairs containing the same type of distortions, denoted as RQI$_{\text{single distortion}}$. We also compare testing RQI on single-scale images, instead of cropping multi-scale patches, denoted as RQI$_{\text{single-scale}}$. The models are compared with our full model on user opinions collected from four datasets DIV2K \cite{div2k}, RealSR \cite{RealSR}, DRealSR \cite{DRealSR}, and Set5\&Set14 \cite{bevilacqua2012low,zeyde2012single}. RQI$_{\text{single-scale}}$ is not tested on Set5\&Set14, since the image resolutions are low and cannot be down-scaled. The results are shown in Table \ref{tab5}.

\begin{table*}[h]
\centering
\setlength{\tabcolsep}{7pt}
\renewcommand\arraystretch{1.1}
\caption{Ablation study of the RQI scheme.}
\footnotesize
\resizebox{0.98\textwidth}{!}{
\setlength{\tabcolsep}{4pt}
\begin{tabular}{l|ccc|ccc|ccc|ccc}
\hline
\rowcolor{tableHeadGray}
\multicolumn{1}{c|}{Dataset}                      & \multicolumn{3}{c|}{DIV2K \cite{div2k}}                      & \multicolumn{3}{c|}{RealSR \cite{RealSR}}                     & \multicolumn{3}{c|}{DRealSR \cite{DRealSR}}                    & \multicolumn{3}{c}{Set5 \cite{bevilacqua2012low}\&Set14 \cite{zeyde2012single}}                 \\ \hline
\multicolumn{1}{c|}{Creterion}                        & SRCC           & PLCC           & Win Rate      & SRCC           & PLCC           & Win Rate      & SRCC           & PLCC           & Win Rate      & SRCC           & PLCC           & Win Rate      \\ \hline
RQI$_{\text{single distortion}}$ & 0.653          & 0.691          & 0.58          & 0.487          & 0.474          & 0.47          & 0.416          & 0.487          & 0.52          & 0.649          & 0.656          & 0.33          \\
RQI$_{\text{single-scale}}$      & 0.721          & 0.758          & 0.63          & 0.490          & 0.479          & 0.48          & 0.493          & 0.550          & \textbf{0.53} & -              & -              & -             \\
RQI$_{\text{full}}$              & \textbf{0.744} & \textbf{0.785} & \textbf{0.65} & \textbf{0.504} & \textbf{0.484} & \textbf{0.49} & \textbf{0.529} & \textbf{0.603} & \textbf{0.53} & \textbf{0.664} & \textbf{0.673} & \textbf{0.35} \\ \hline
\end{tabular}}
\label{tab5}
\end{table*}

From Table \ref{tab5}, several observations can be drawn. First, training RQI exclusively on same-distortion image pairs leads to a clear performance drop. We attribute this to two main factors.
(1) GT and SR images typically exhibit different types of distortions. GT images often contain natural distortions such as noise or blur, whereas SR outputs introduce algorithm-induced artifacts with different characteristics. Training RQI across heterogeneous distortion types allows the model to learn a broader representation that better captures complex and diverse degradations.
(2) Image pairs constructed from the same distortion type often carry large and obvious quality differences, providing limited fine-grained supervision. As a result, the model becomes less capable of handling subtle perceptual distinctions.
Second, when applying multi-scale patch evaluation, we observe slight performance gains on RealSR \cite{RealSR}, but notably larger improvements on DIV2K \cite{div2k} and DRealSR \cite{DRealSR}. Since images in DIV2K and DRealSR are of higher resolution, evaluating RQI in multi-scale not only captures detailed texture quality but also measures structure and semantic consistency, leading to better alignment with human perception. 

In Table \ref{tab6}, we train under RQI using different losses and report the SRCC results on varying datasets. As can be seen, both L1 and L2 losses perform slightly worse than the Huber loss.

\begin{table}[t]
\centering
\caption{SRCCs results for selecting different losses under RQI.}
\label{tab6}
\resizebox{\linewidth}{!}{
\setlength{\tabcolsep}{4pt}
\begin{tabular}{l|ccccccc}
\hline
\rowcolor{tableHeadGray}
\multicolumn{1}{c|}{Metric}    & DIV2K & RealSR & DRealSR & Set5Set14 & BSD   & QADS  & SRIQA \\ \hline
RQI$_{L1}$   & 0.704 & 0.473  & 0.496   & 0.655     & 0.862 & 0.910 & 0.691 \\
RQI$_{L2}$   & 0.712 & 0.479  & 0.514   & 0.649     & 0.881 & 0.905 & 0.725 \\ \hline
RQI       & 0.744 & 0.504  & 0.529   & 0.664     & 0.901 & 0.912 & 0.733 \\ \hline
\end{tabular}
}
\end{table}

In Table \ref{tab7}, we show how RQI variants (different models trained on varying datasets under the RQI scheme) perform on the four public IQA benchmarks, where SRCC$_{\text{mean}}$ are reported. Comparing with other models in Table \ref{tab3}, all variants generally perform well across benchmarks.

\begin{table}[h]
\centering
\caption{Consistency evaluations of RQI variants on four IQA benchmarks.}
\label{tab7}
\footnotesize
\setlength{\tabcolsep}{3pt}
\begin{tabular}{l|cccc}
\hline
\rowcolor{tableHeadGray}
\multicolumn{1}{c|}{Model/Dataset}    & BSD-SR & QADS  & SRIQA-Bench & Kadid-10K \\ \hline
AHIQ/Kadid-10K   & 0.838  & 0.855 & 0.582       & -         \\
AHIQ/PieAPP      & 0.875  & 0.903 & 0.592       & 0.706     \\
AHIQ/PIPAL       & 0.896  & 0.908 & 0.752       & 0.592     \\ \hline
MANIQA/Kadid-10K & 0.842  & 0.866 & 0.571       & -         \\
MANIQA/PieAPP    & 0.880  & 0.936 & 0.615       & 0.870     \\
MANIQA/PIPAL     & 0.901  & 0.912 & 0.733       & 0.669     \\ \hline
TOPIQ/Kadid-10K  & 0.844  & 0.896 & 0.665       & -         \\
TOPIQ/PieAPP     & 0.838  & 0.885 & 0.635       & 0.763     \\
TOPIQ/PIPAL      & 0.813  & 0.867 & 0.687       & 0.520     \\ \hline
\end{tabular}
\end{table}

\section{More Qualitative Comparisons}
\label{secD}
\begin{figure*}[h]
\centering
\subfloat{
\hspace{-0.2cm}
\begin{minipage}[t]{1\textwidth}
   \centering
  \includegraphics[angle=0,width=1\textwidth]{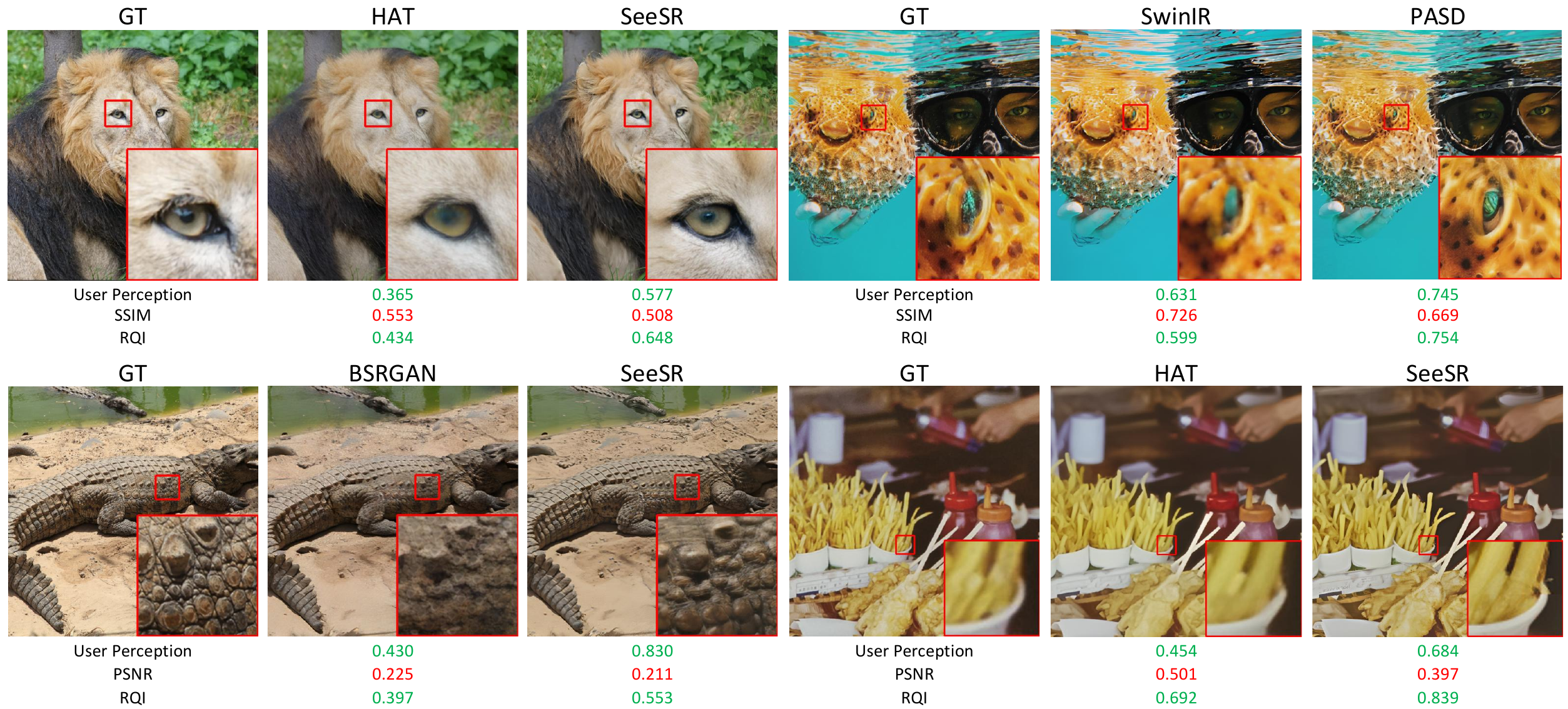}
\end{minipage}
}
\caption{Distortion-based FR-IQA metrics SSIM \cite{ssim} and PSNR tend to favor blurry regions over textures, leading to contradictory predictions with human perception.}
\label{fig8}
\end{figure*}

\begin{figure*}[h]
\centering
\subfloat{
\hspace{-0.2cm}
\begin{minipage}[t]{1\textwidth}
   \centering
  \includegraphics[angle=0,width=1\textwidth]{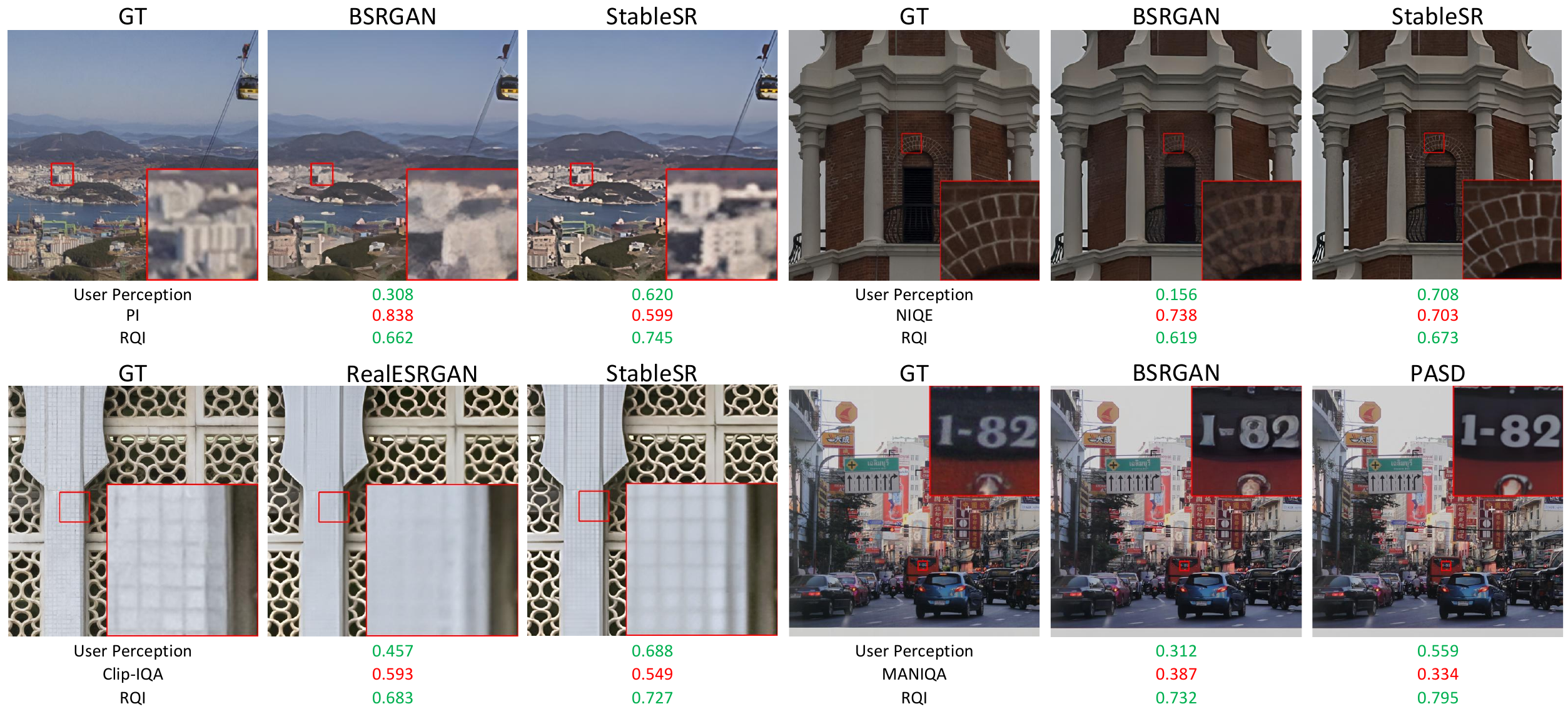}
\end{minipage}
}
\caption{NR-IQA metrics PI \cite{blau20182018}, NIQE \cite{NIQE}, Clip-IQA \cite{wang2023exploring} and MANIQA \cite{yang2022maniqa} can fail on cases where subtle structure of semantics are changed, due to the lack of proper references.}
\label{fig9}
\end{figure*}

\begin{figure*}[h]
\centering
\subfloat{
\hspace{-0.2cm}
\begin{minipage}[t]{1\textwidth}
   \centering
  \includegraphics[angle=0,width=1\textwidth]{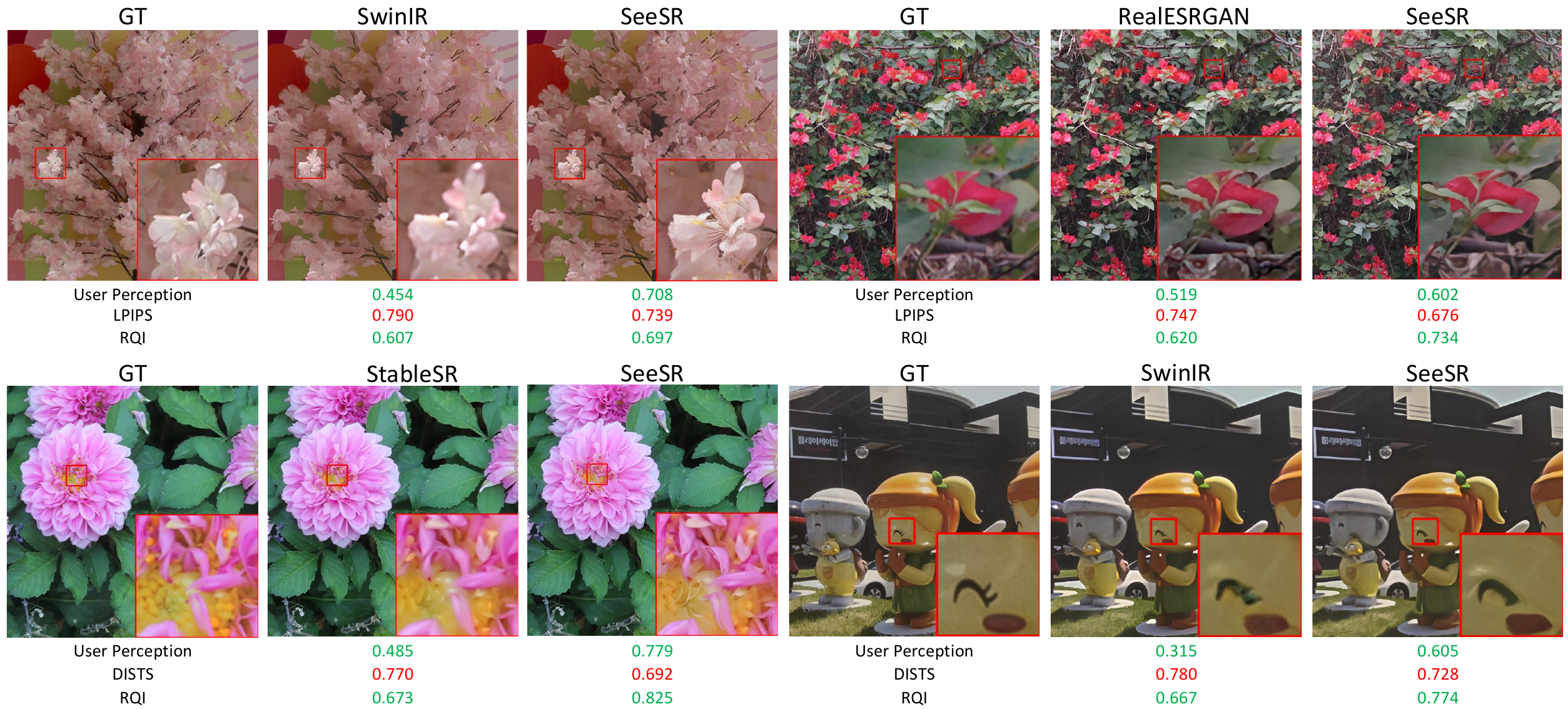}
\end{minipage}
}
\caption{Perception-based FR-IQA metrics LPIPS \cite{LPIPS} and DISTS \cite{DISTS} can fail when GT quality is relatively lower. They make contradictory evaluations for models that output perceptually higher results than GTs.}
\label{fig10}
\end{figure*}

In this section, we provide more qualitative comparisons to show how RQI outperforms different types of metrics. For easier comparison, all scores are normalized to $[0,1]$, and a higher score indicates better visual quality. Figure \ref{fig8} compares RQI with two distortion-based FR-IQA metrics (SSIM \cite{ssim} and PSNR) in assessing detailed textures, Figure \ref{fig9} compares RQI with four NR-IQA metrics (PI \cite{blau20182018}, NIQE \cite{NIQE}, Clip-IQA \cite{wang2023exploring} and MANIQA \cite{yang2022maniqa}) in evaluating subtle structure or semantic changes. Figure \ref{fig10} compares RQI with two perception-based FR-IQA metrics (LPIPS \cite{LPIPS} and DISTS \cite{DISTS}) in poor GT quality cases. As can be seen, RQI makes correct evaluations on all the cases, showing its superiority as a reliable image metric for SR evaluations.

\section{Training SR Models with RQI Loss} 
\label{secE}
\subsection{Training Details}

We train three SR models, SwinIR \cite{SwinIR}, SeeSR \cite{SeeSR} and PiSA-SR \cite{Pisasr}, following their official training configurations. For SwinIR, the model is first trained with MSE loss for 100K epochs, then trained together with the perceptual loss, GAN loss and RQI loss for another 100K epochs.
For PiSA-SR, we adopt the same two-stage strategy as in the original paper: we first train the pixel-level LoRA using the standard pixel-wise loss for 4K iterations, and then train the semantic-level LoRA by introducing the RQI loss for 8.5K iterations, as this component is responsible for the reconstruction of perceptual details.
For SeeSR \cite{SeeSR}, because its loss is defined in the latent feature space, we follow its pipeline by decoding the latent representations using the frozen decoder and computing the RQI loss on the reconstructed images, where we train 150K iterations for the whole model.

\subsection{More Qualitative Comparisons between w/ and w/o RQI Loss}

\begin{figure*}[t]
\centering

\begin{minipage}[b]{0.24\textwidth}
    \centering
    \includegraphics[width=\linewidth]{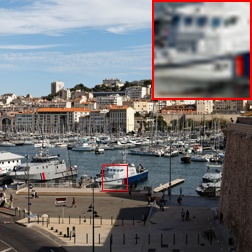}
\end{minipage}
\begin{minipage}[b]{0.24\textwidth}
    \centering
    \includegraphics[width=\linewidth]{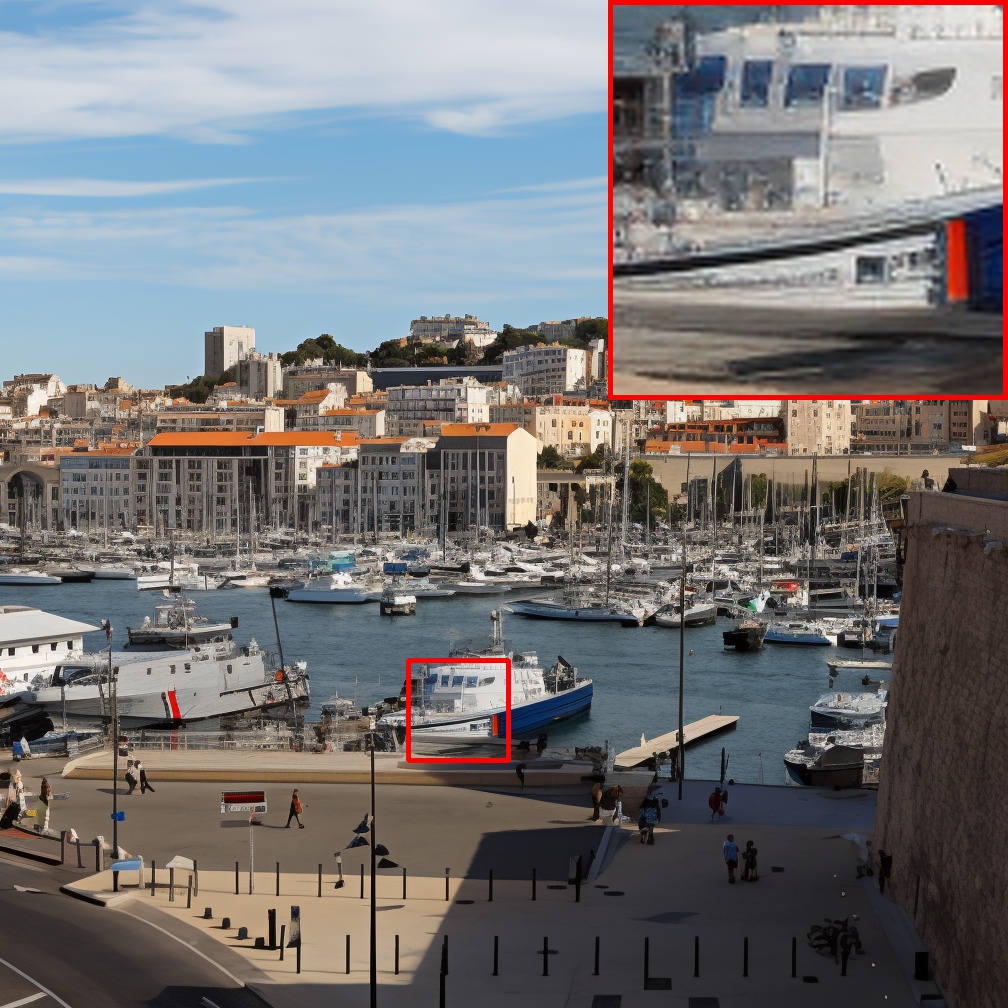}
\end{minipage}
\begin{minipage}[b]{0.24\textwidth}
    \centering
    \includegraphics[width=\linewidth]{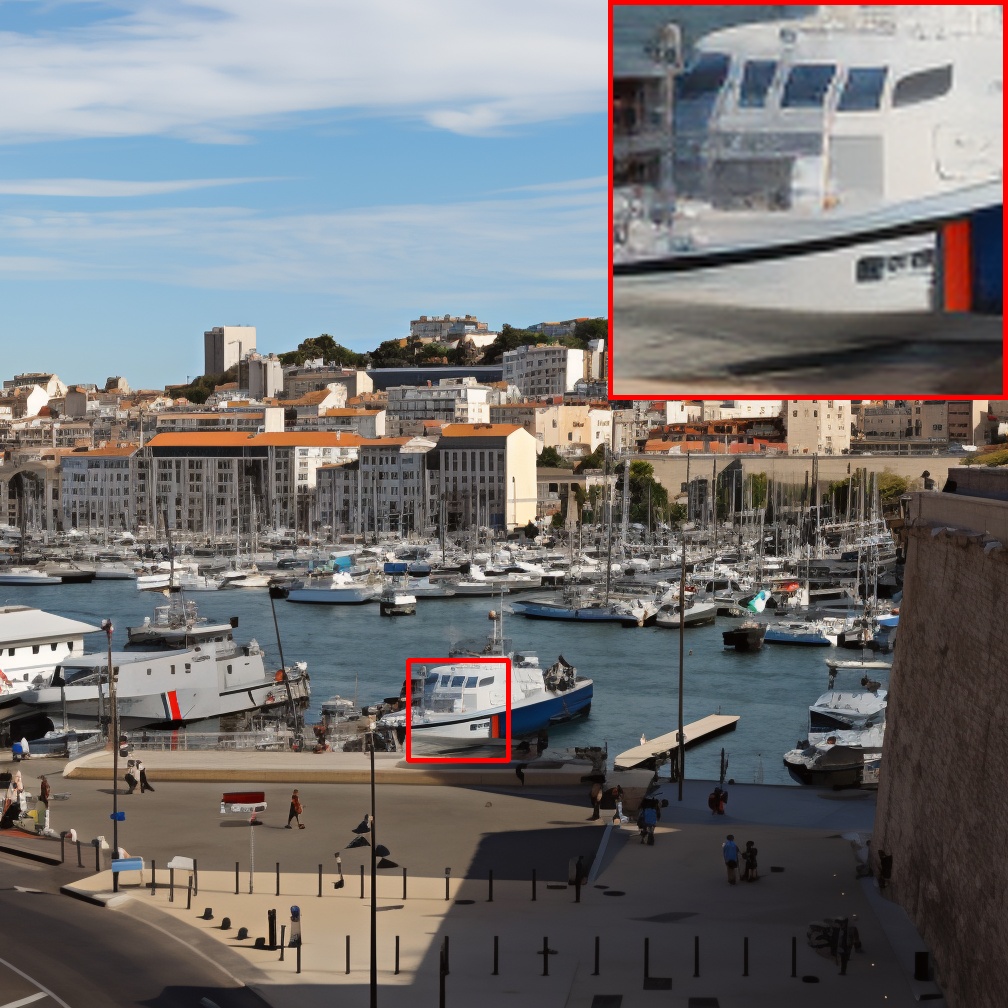}
\end{minipage}
\begin{minipage}[b]{0.24\textwidth}
    \centering
    \includegraphics[width=\linewidth]{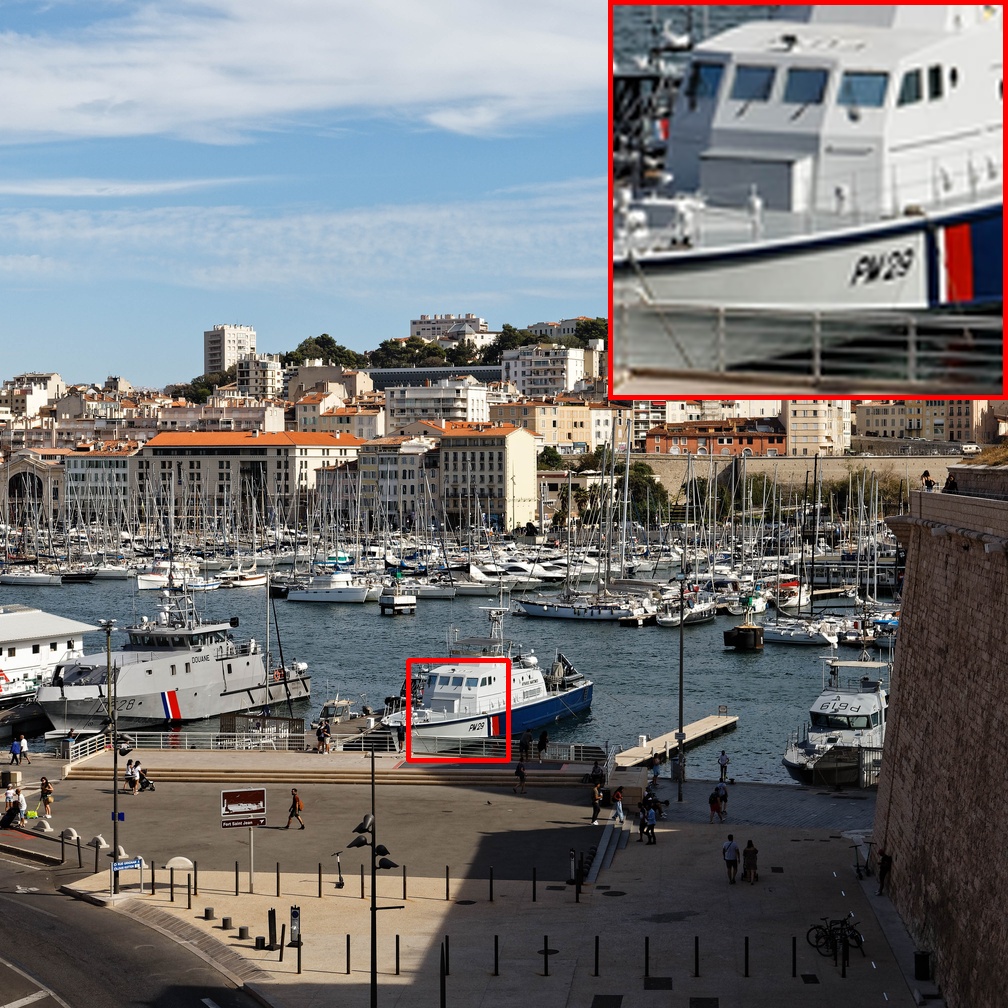}
\end{minipage}

\begin{minipage}[b]{0.24\textwidth}
    \centering
    \includegraphics[width=\linewidth]{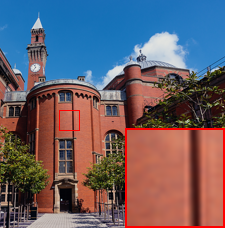}
\end{minipage}
\begin{minipage}[b]{0.24\textwidth}
    \centering
    \includegraphics[width=\linewidth]{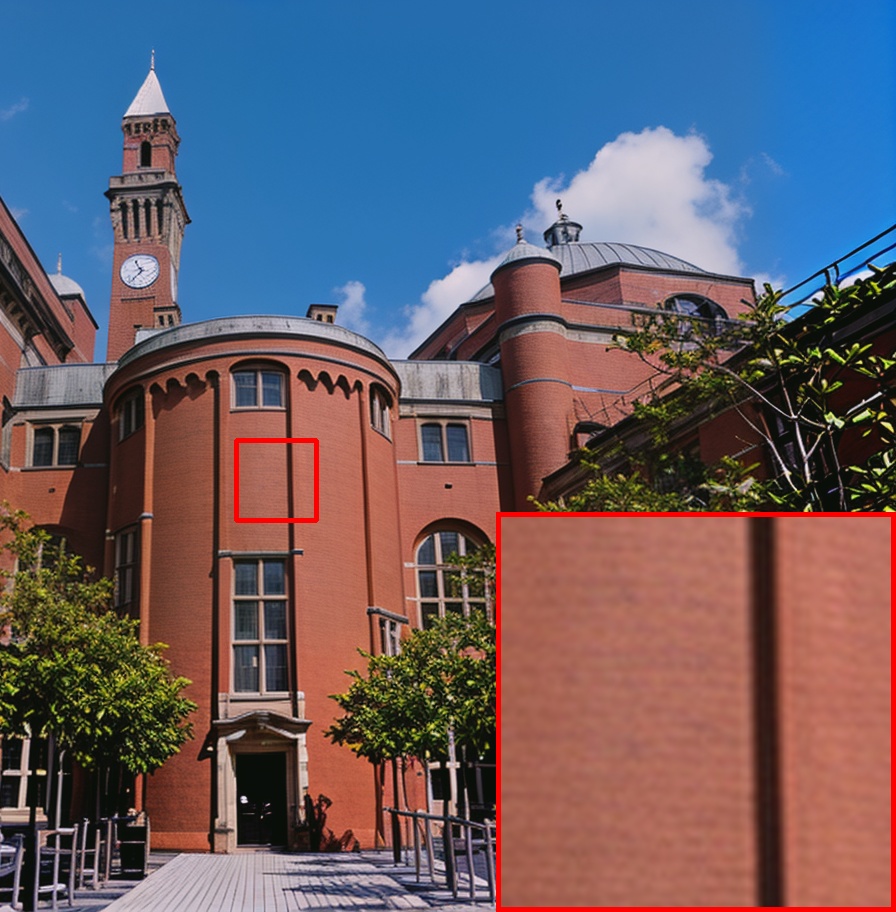}
\end{minipage}
\begin{minipage}[b]{0.24\textwidth}
    \centering
    \includegraphics[width=\linewidth]{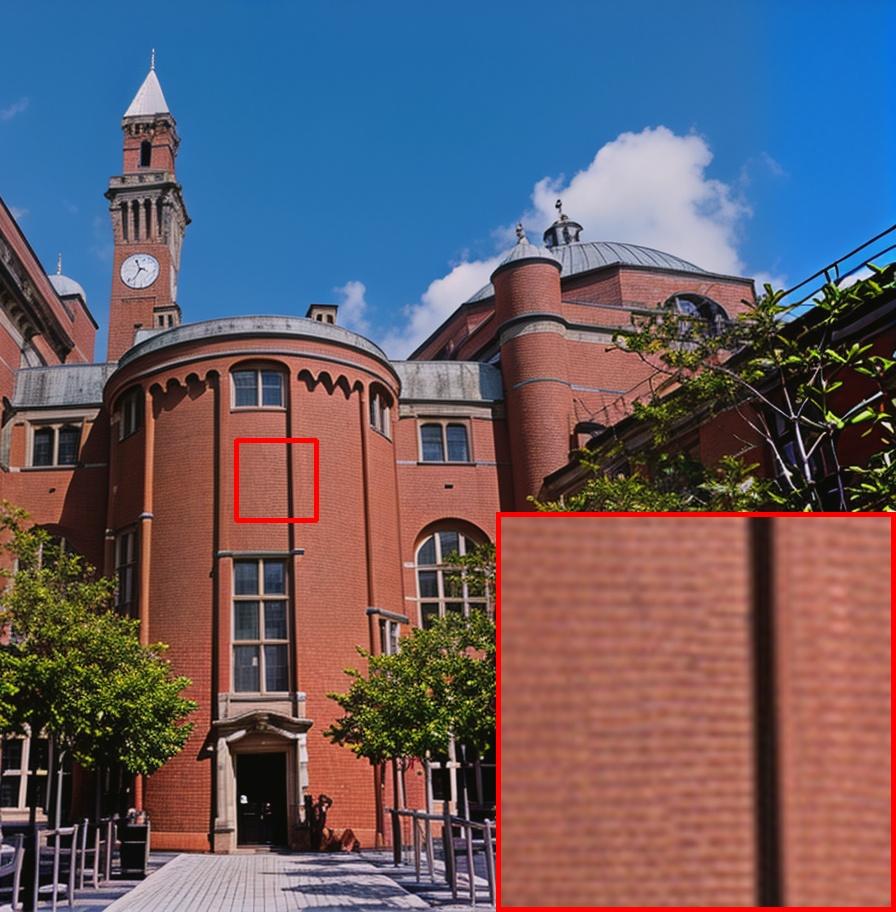}
\end{minipage}
\begin{minipage}[b]{0.24\textwidth}
    \centering
    \includegraphics[width=\linewidth]{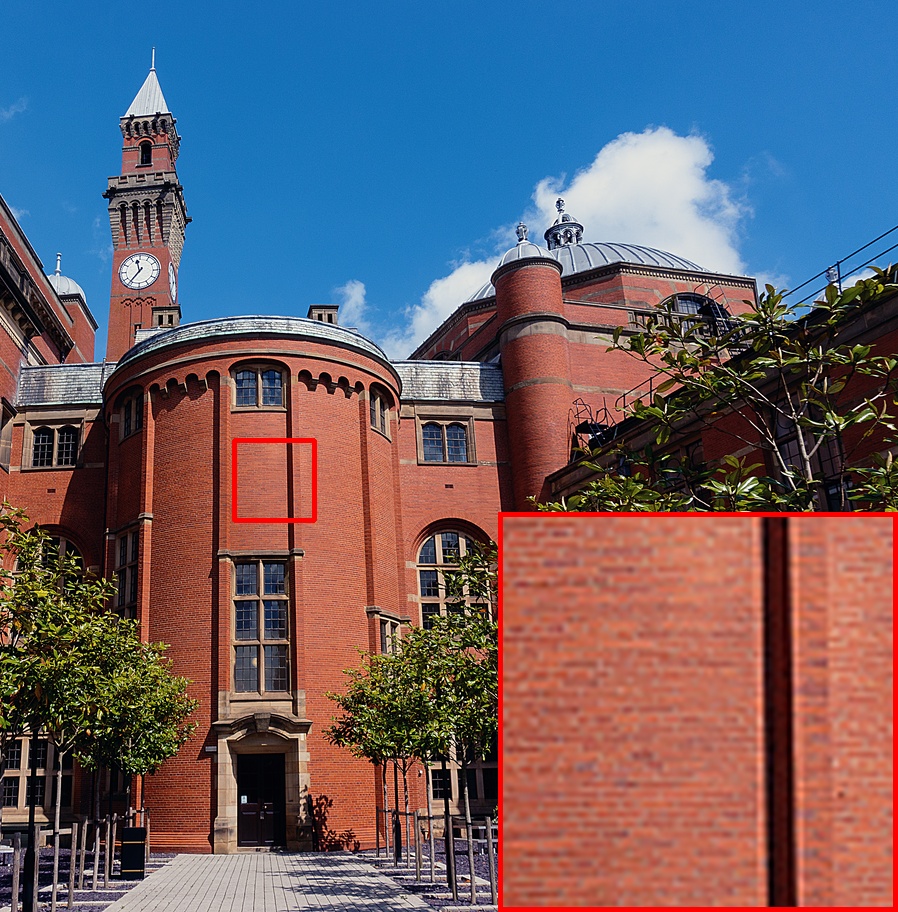}
\end{minipage}

\begin{minipage}[b]{0.24\textwidth}
    \centering
    \includegraphics[width=\linewidth]{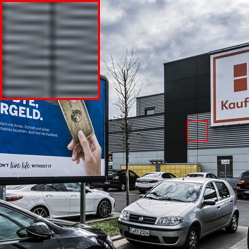}
\end{minipage}
\begin{minipage}[b]{0.24\textwidth}
    \centering
    \includegraphics[width=\linewidth]{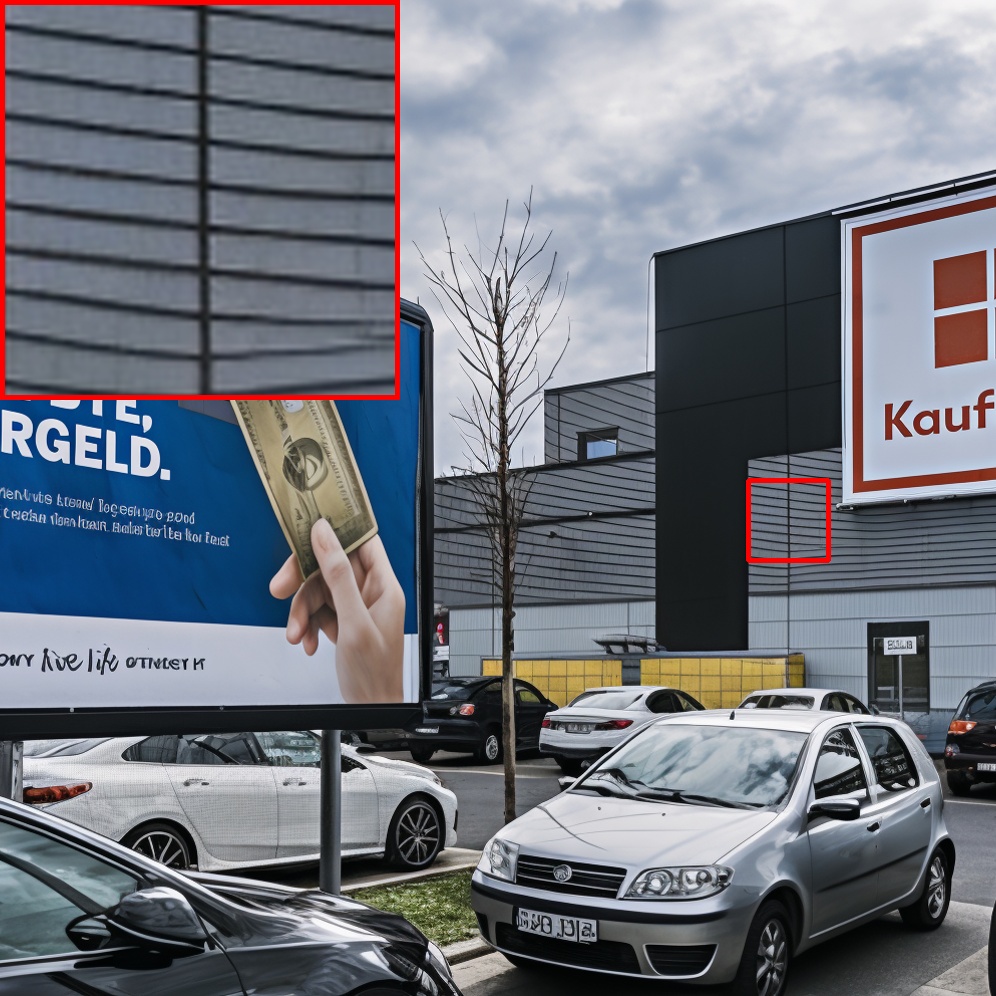}
\end{minipage}
\begin{minipage}[b]{0.24\textwidth}
    \centering
    \includegraphics[width=\linewidth]{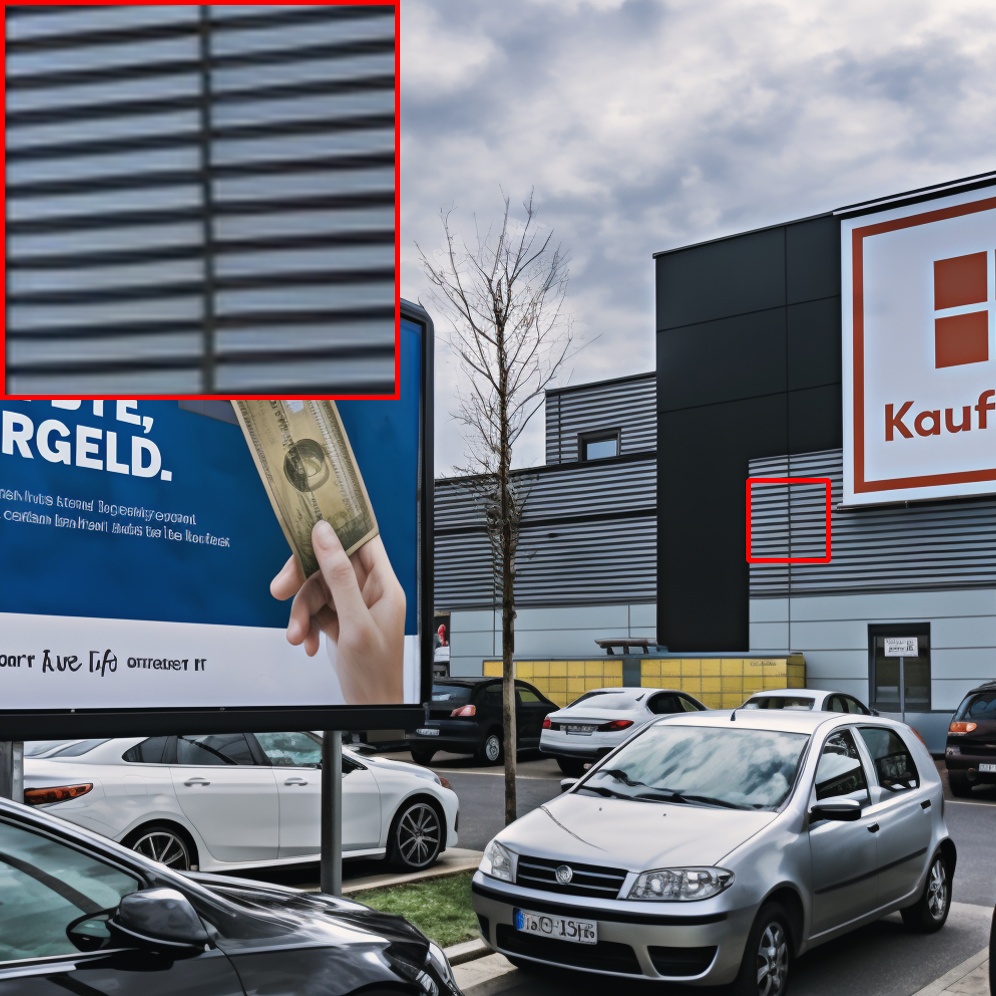}
\end{minipage}
\begin{minipage}[b]{0.24\textwidth}
    \centering
    \includegraphics[width=\linewidth]{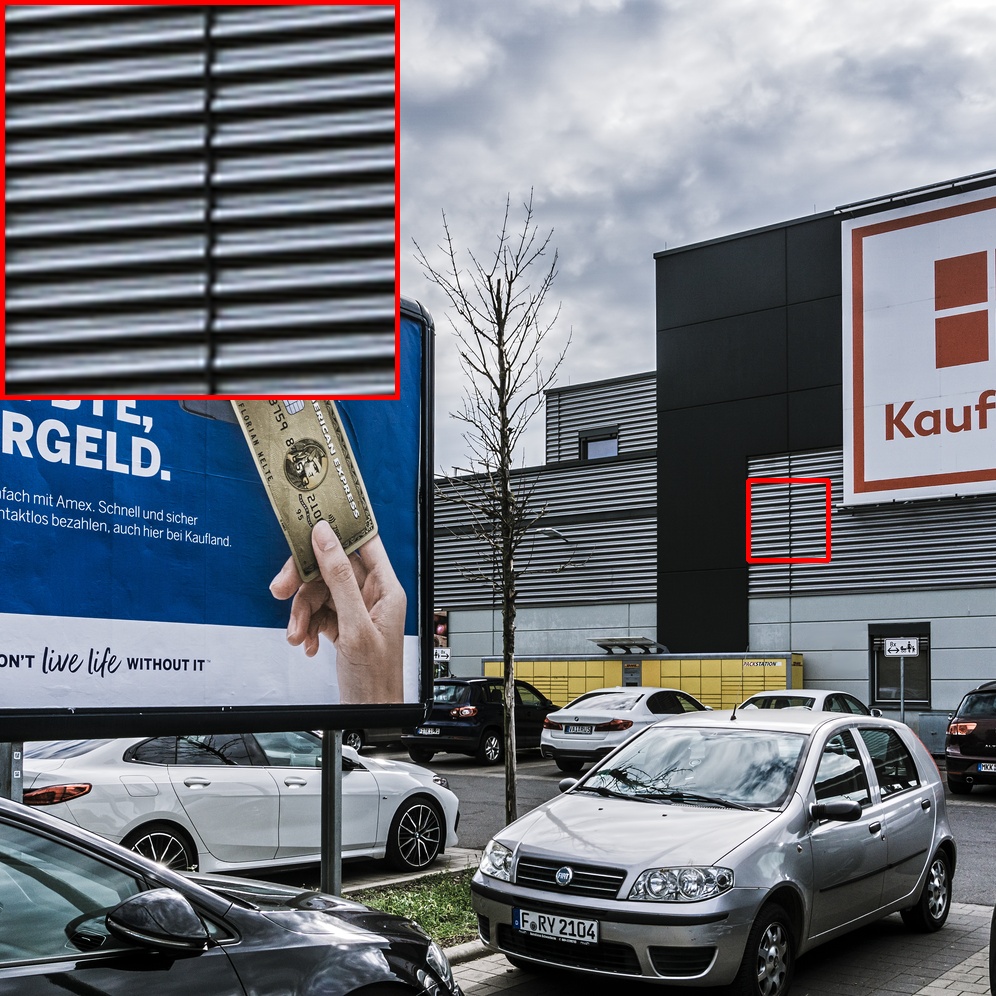}
\end{minipage}

\begin{minipage}[b]{0.24\textwidth}
    \centering
    \includegraphics[width=\linewidth]{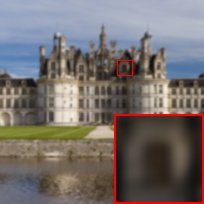}
\end{minipage}
\begin{minipage}[b]{0.24\textwidth}
    \centering
    \includegraphics[width=\linewidth]{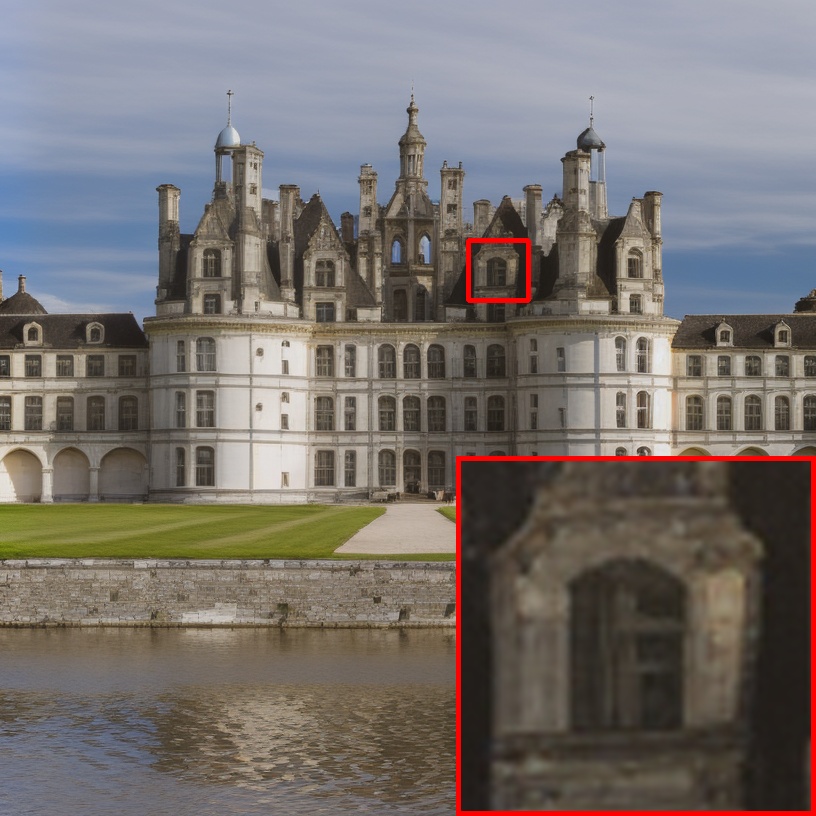}
\end{minipage}
\begin{minipage}[b]{0.24\textwidth}
    \centering
    \includegraphics[width=\linewidth]{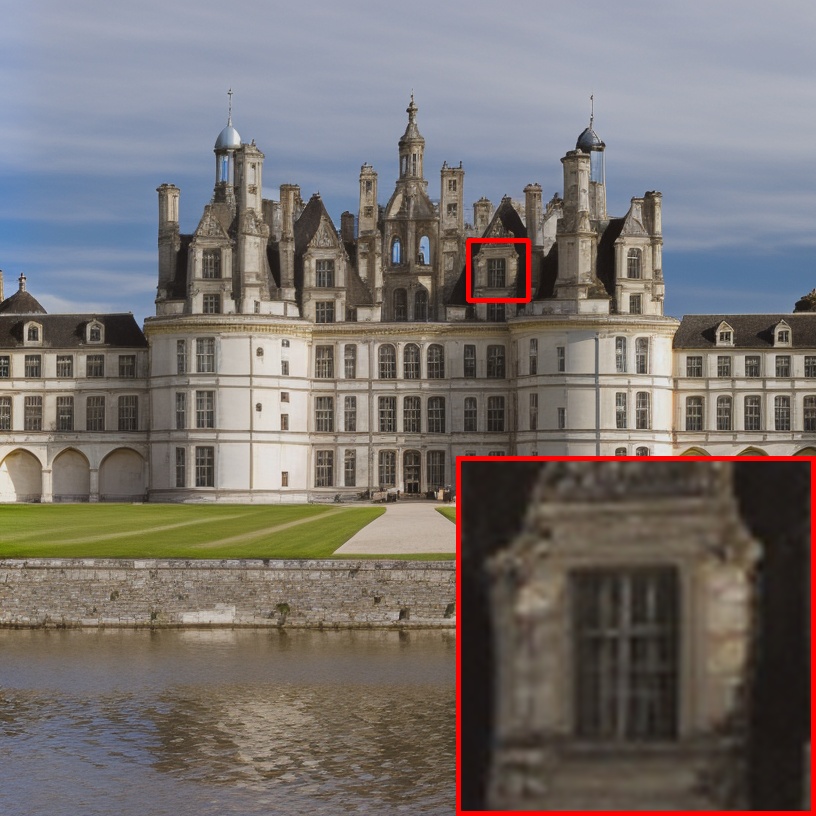}
\end{minipage}
\begin{minipage}[b]{0.24\textwidth}
    \centering
    \includegraphics[width=\linewidth]{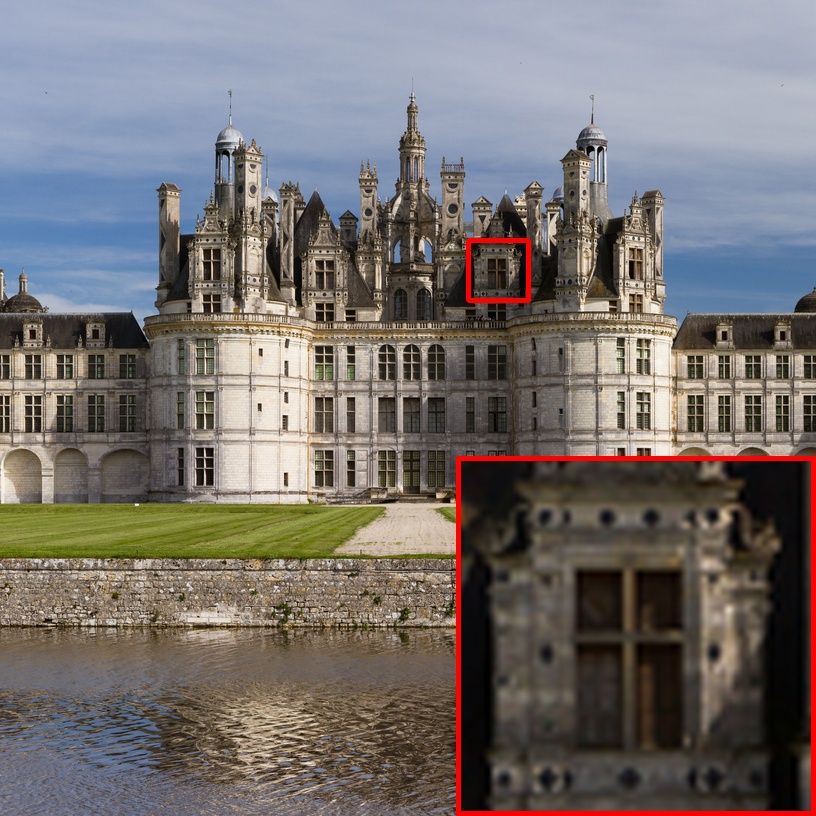}
\end{minipage}

\begin{minipage}[b]{0.24\textwidth}
    \centering
    \includegraphics[width=\linewidth]{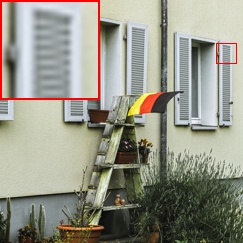}
    \\[-5pt]{\scriptsize LR}
\end{minipage}
\begin{minipage}[b]{0.24\textwidth}
    \centering
    \includegraphics[width=\linewidth]{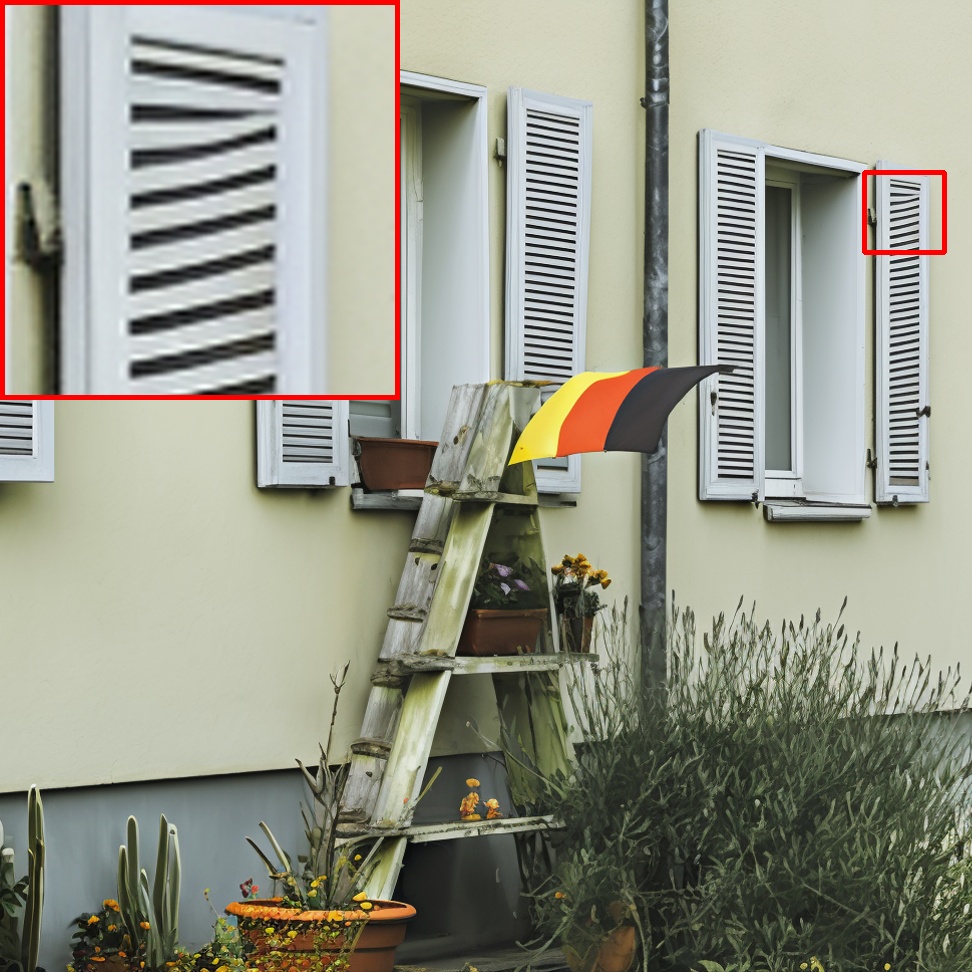}
    \\[-5pt]{\scriptsize Baseline}
\end{minipage}
\begin{minipage}[b]{0.24\textwidth}
    \centering
    \includegraphics[width=\linewidth]{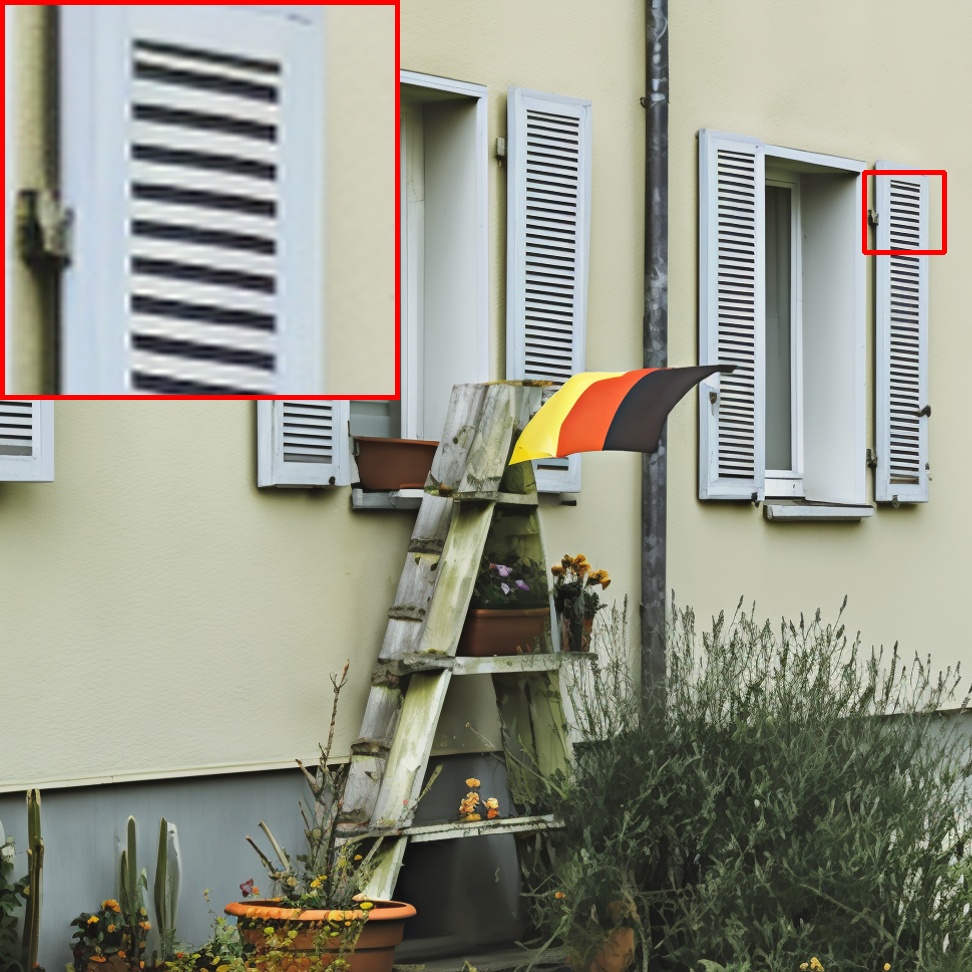}
    \\[-5pt]{\scriptsize Baseline+RQI}
\end{minipage}
\begin{minipage}[b]{0.24\textwidth}
    \centering
    \includegraphics[width=\linewidth]{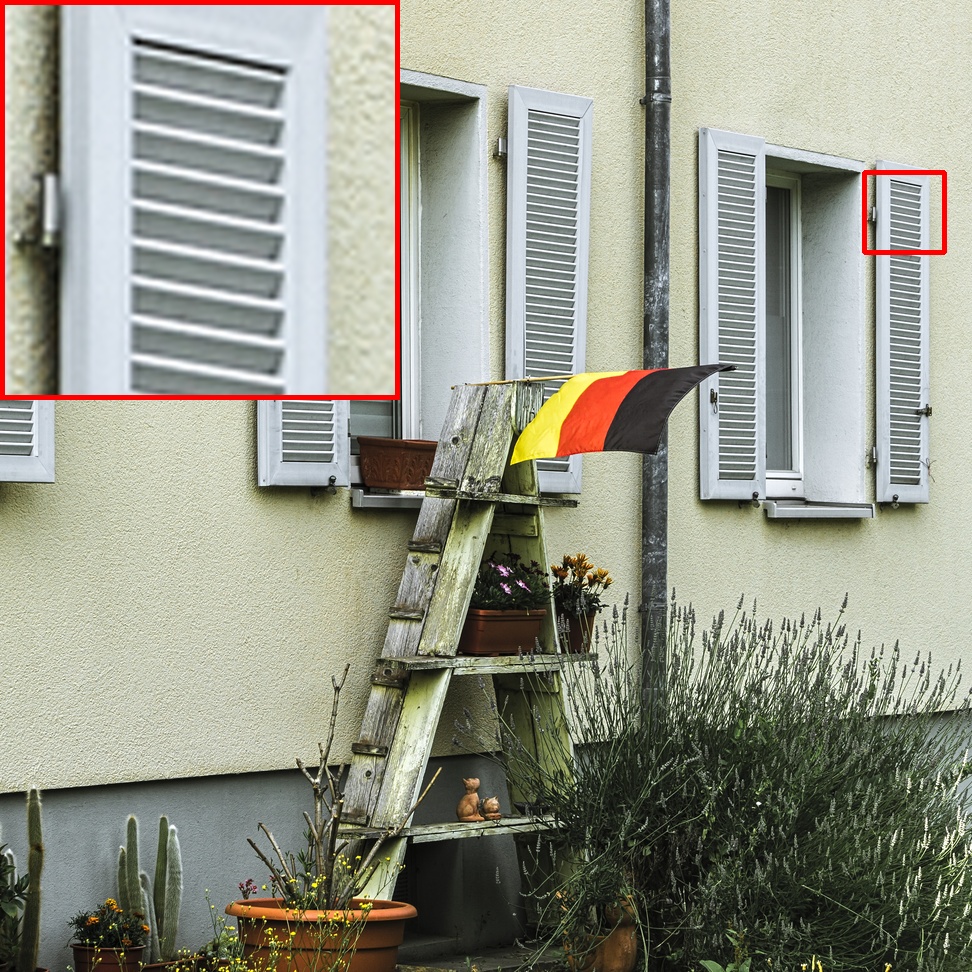}
    \\[-5pt]{\scriptsize GT}
\end{minipage}

\caption{More visual comparison of training advanced SR models with RQI as an auxiliary loss. RQI is effective in preserving structural fidelity. Please zoom in for a better view.}
\label{fig11}
\vspace{-6pt}
\end{figure*}

\begin{figure*}[t]
\centering

\begin{minipage}[b]{0.24\textwidth}
    \centering
    \includegraphics[width=\linewidth]{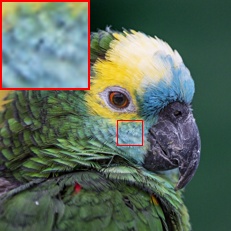}
\end{minipage}
\begin{minipage}[b]{0.24\textwidth}
    \centering
    \includegraphics[width=\linewidth]{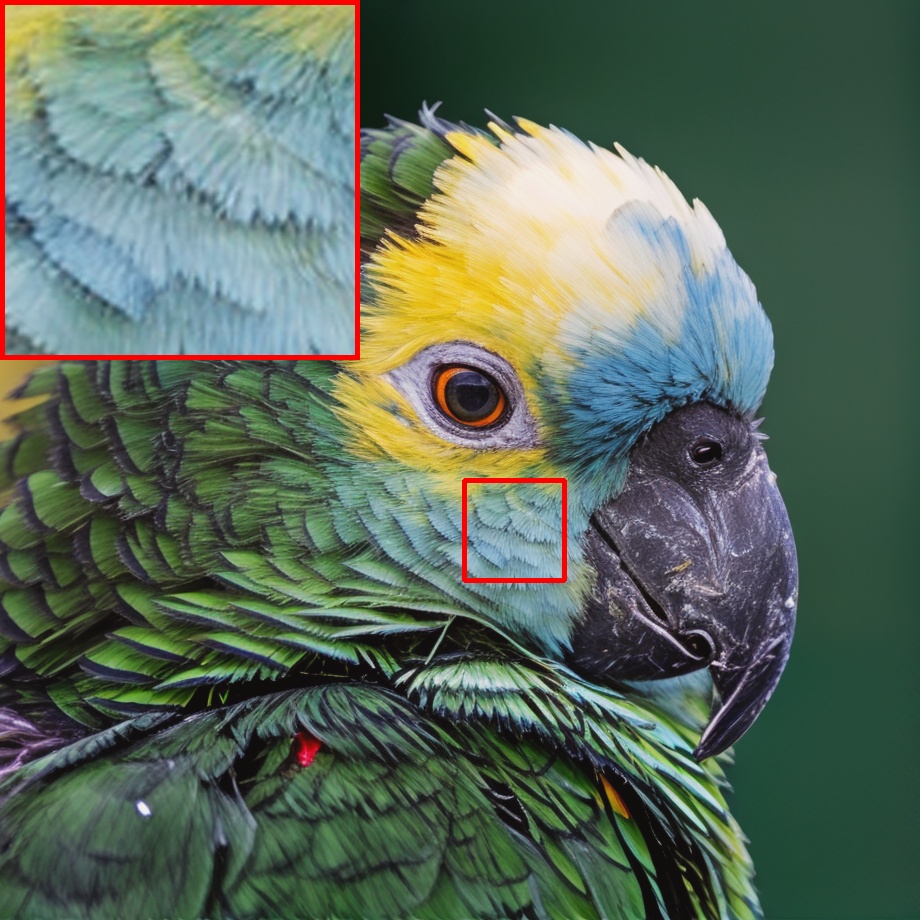}
\end{minipage}
\begin{minipage}[b]{0.24\textwidth}
    \centering
    \includegraphics[width=\linewidth]{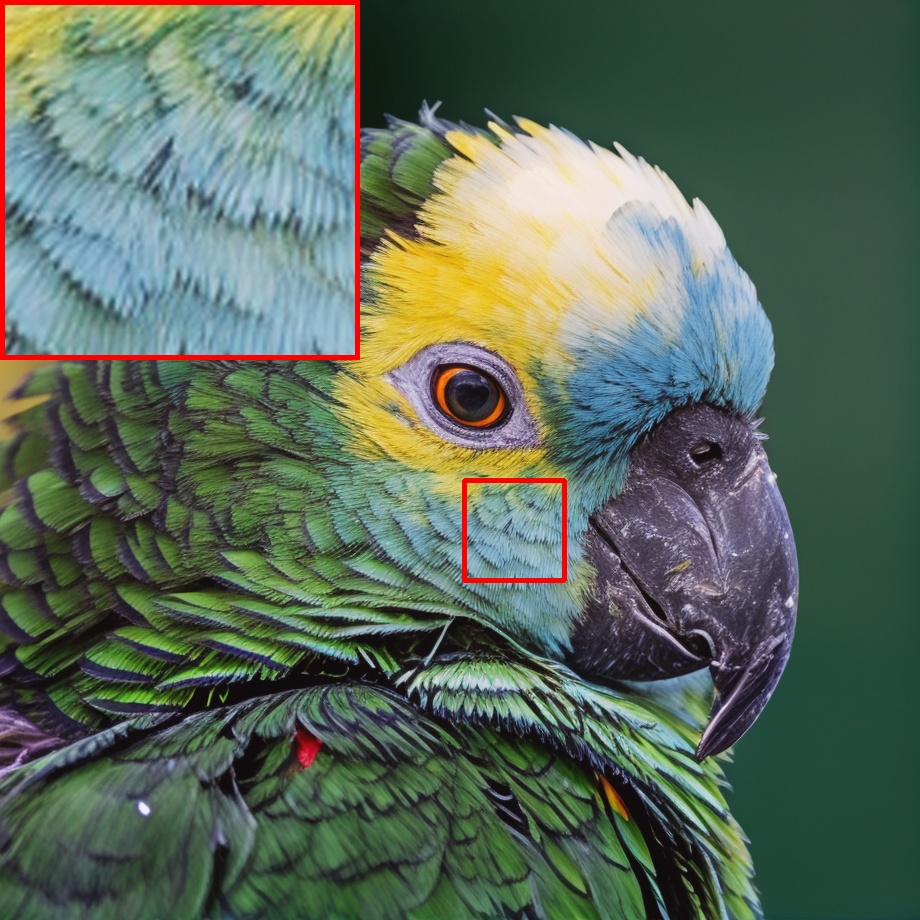}
\end{minipage}
\begin{minipage}[b]{0.24\textwidth}
    \centering
    \includegraphics[width=\linewidth]{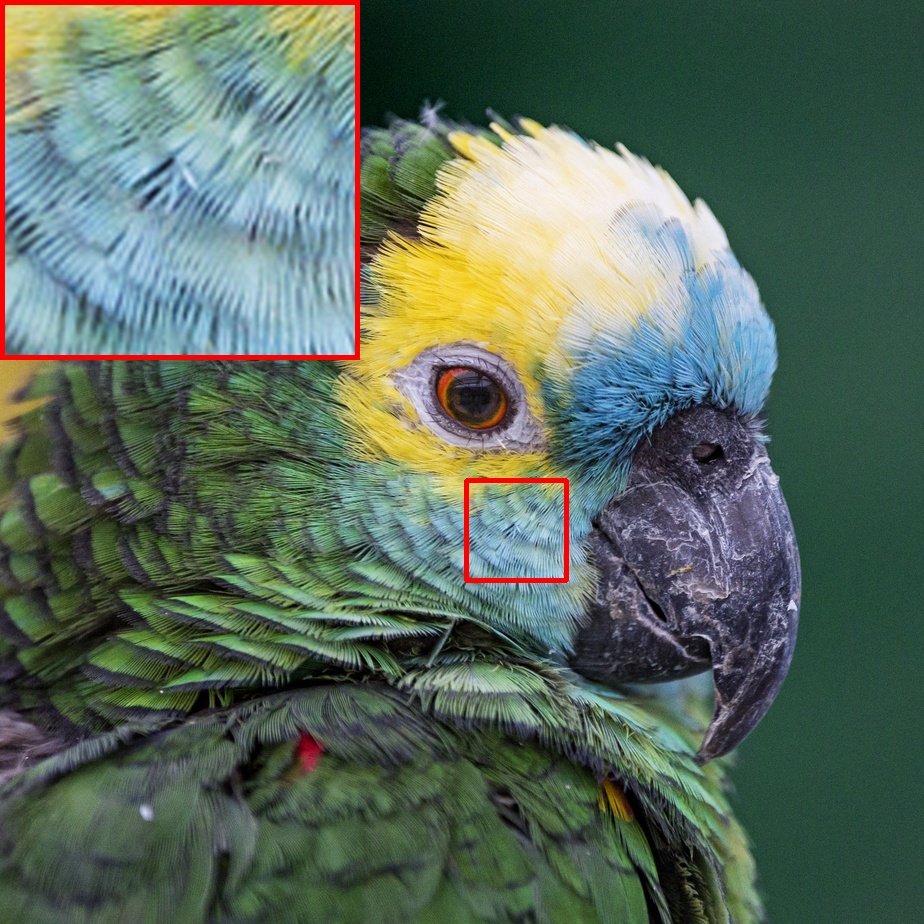}
\end{minipage}

\begin{minipage}[b]{0.24\textwidth}
    \centering
    \includegraphics[width=\linewidth]{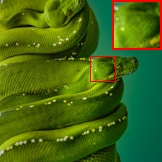}
\end{minipage}
\begin{minipage}[b]{0.24\textwidth}
    \centering
    \includegraphics[width=\linewidth]{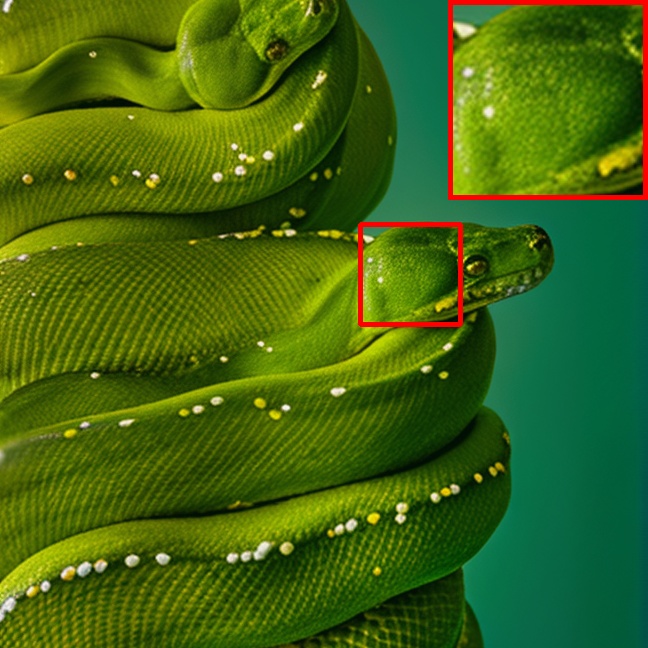}
\end{minipage}
\begin{minipage}[b]{0.24\textwidth}
    \centering
    \includegraphics[width=\linewidth]{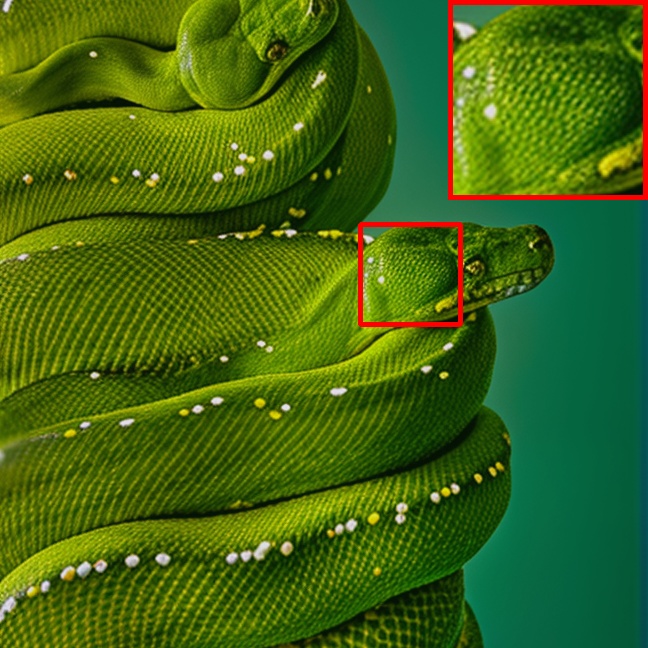}
\end{minipage}
\begin{minipage}[b]{0.24\textwidth}
    \centering
    \includegraphics[width=\linewidth]{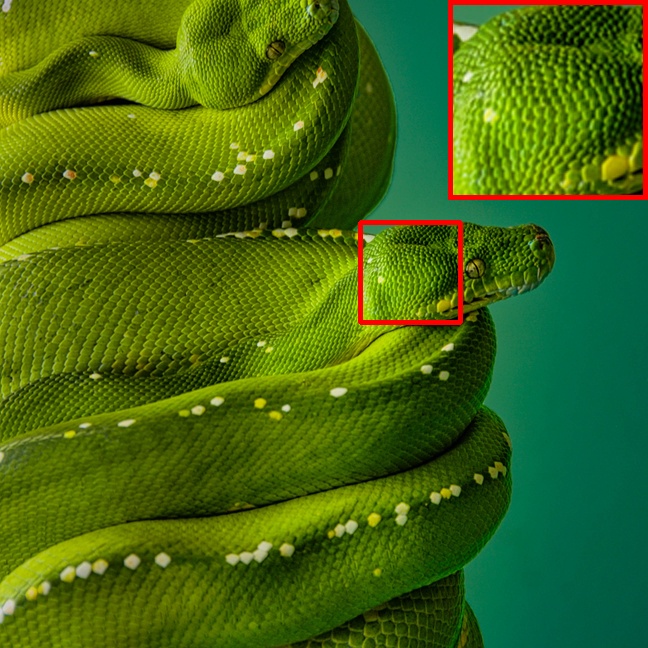}
\end{minipage}

\begin{minipage}[b]{0.24\textwidth}
    \centering
    \includegraphics[width=\linewidth]{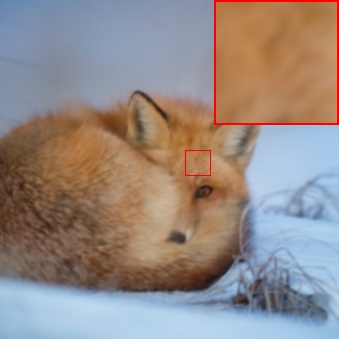}
\end{minipage}
\begin{minipage}[b]{0.24\textwidth}
    \centering
    \includegraphics[width=\linewidth]{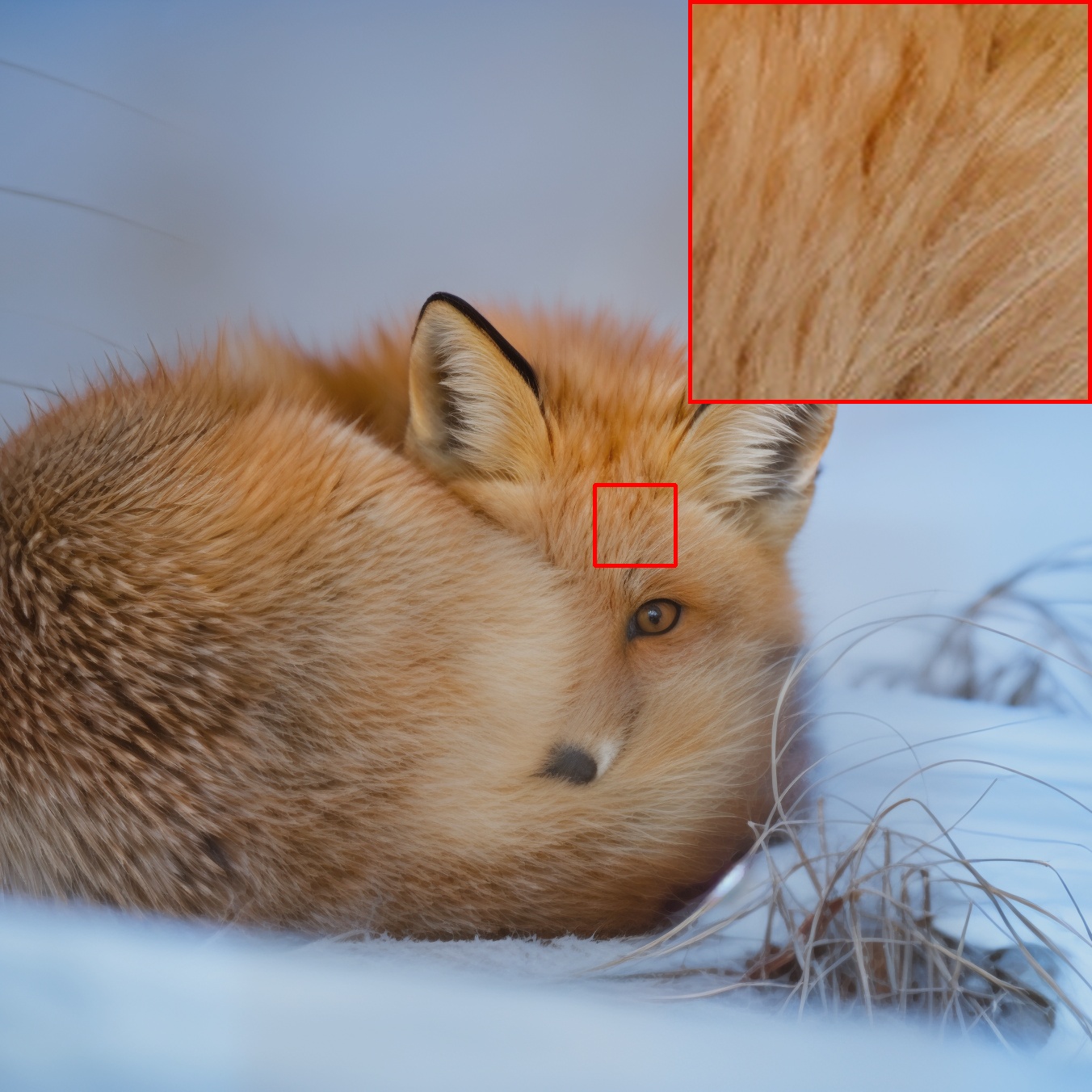}
\end{minipage}
\begin{minipage}[b]{0.24\textwidth}
    \centering
    \includegraphics[width=\linewidth]{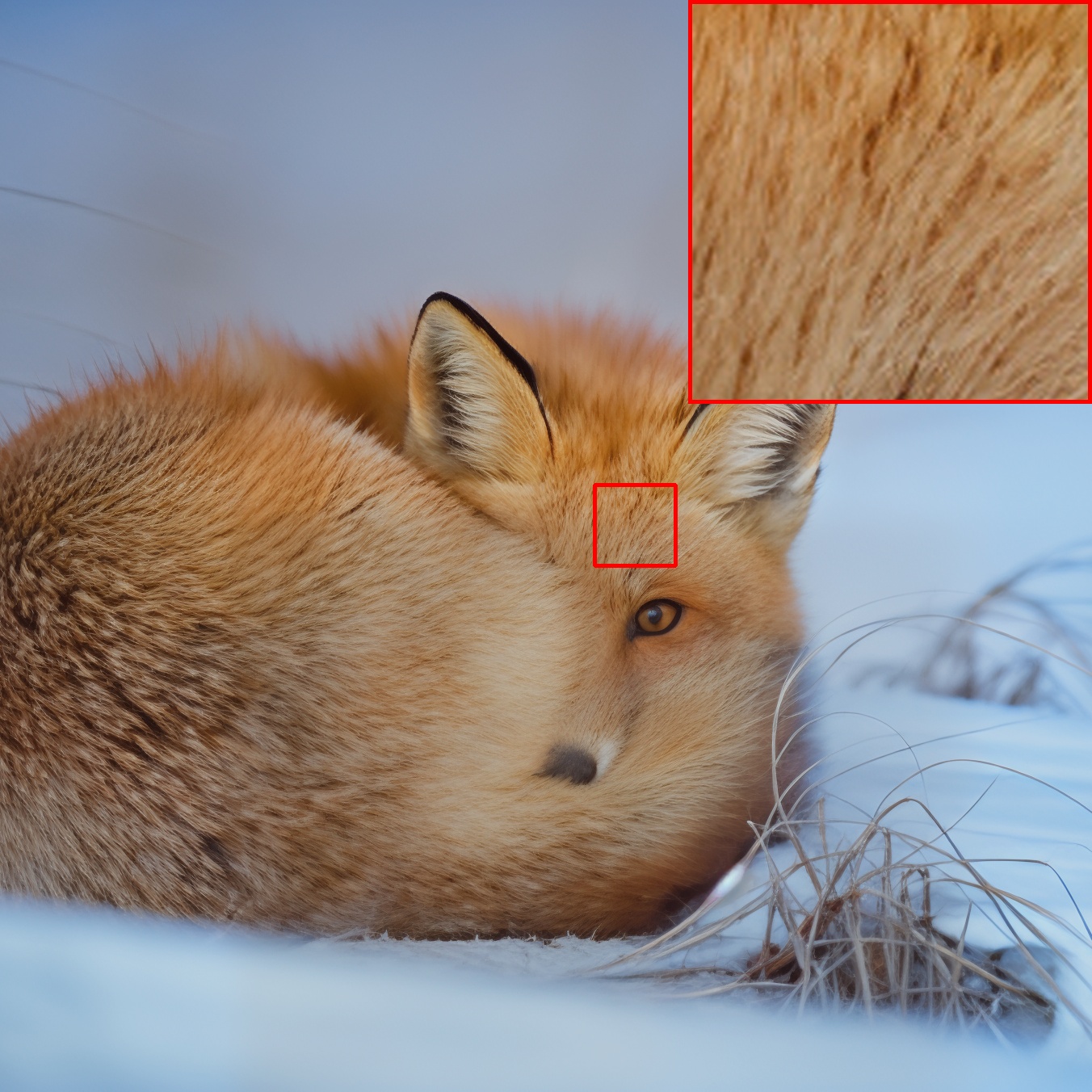}
\end{minipage}
\begin{minipage}[b]{0.24\textwidth}
    \centering
    \includegraphics[width=\linewidth]{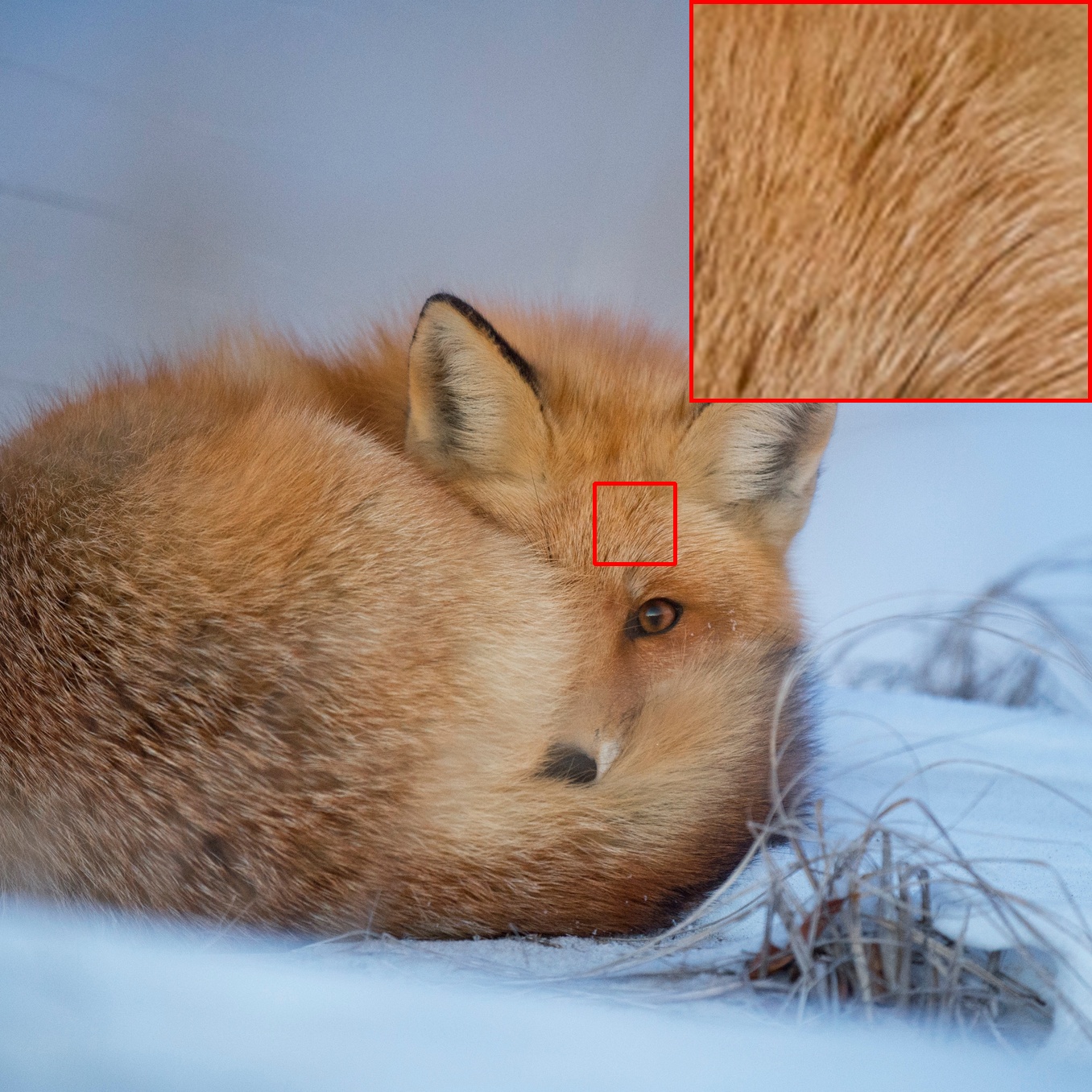}
\end{minipage}

\begin{minipage}[b]{0.24\textwidth}
    \centering
    \includegraphics[width=\linewidth]{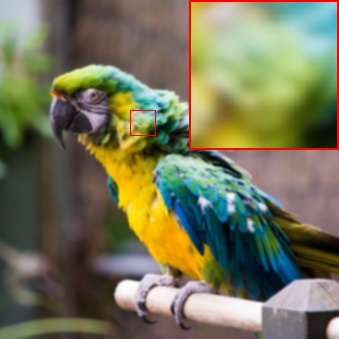}
\end{minipage}
\begin{minipage}[b]{0.24\textwidth}
    \centering
    \includegraphics[width=\linewidth]{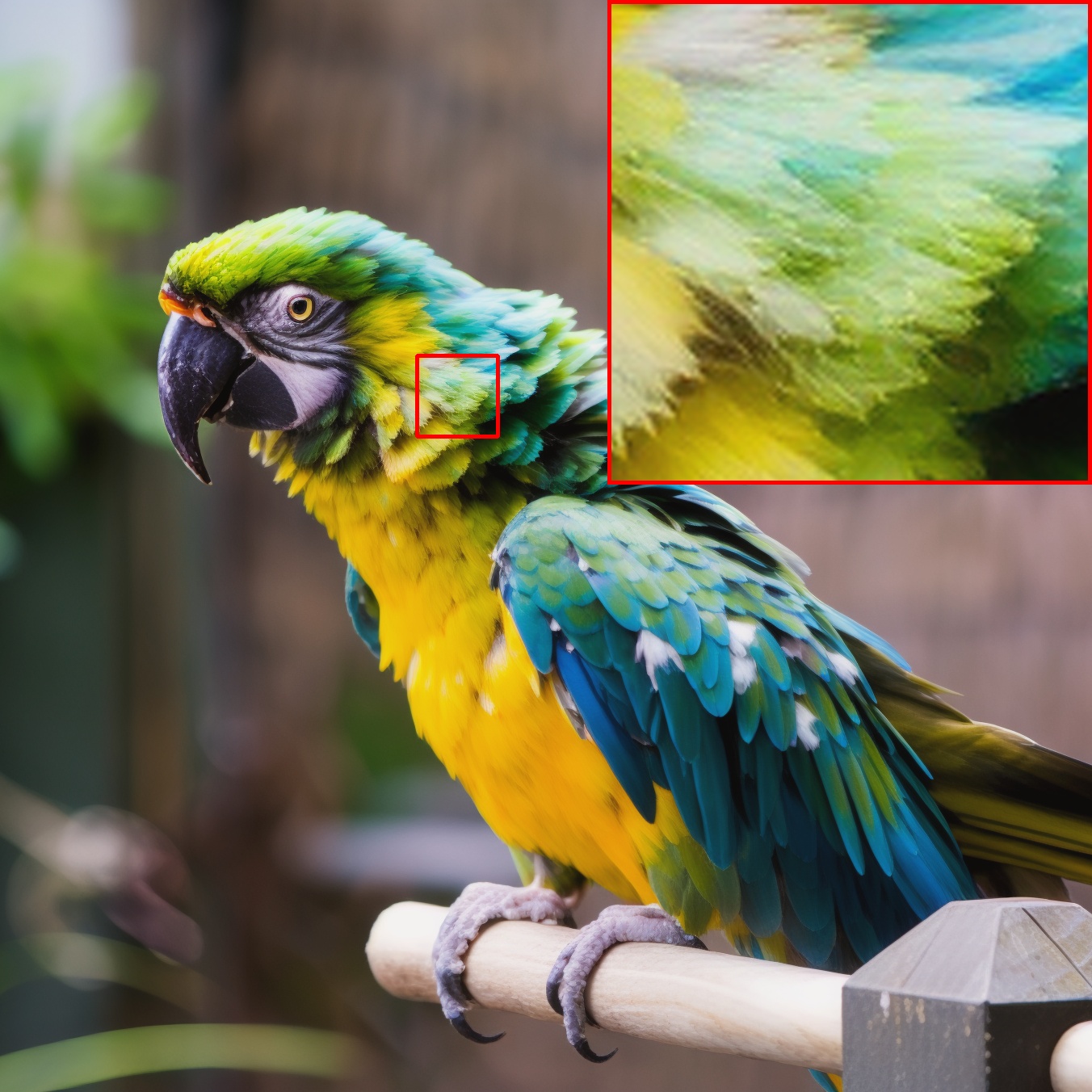}
\end{minipage}
\begin{minipage}[b]{0.24\textwidth}
    \centering
    \includegraphics[width=\linewidth]{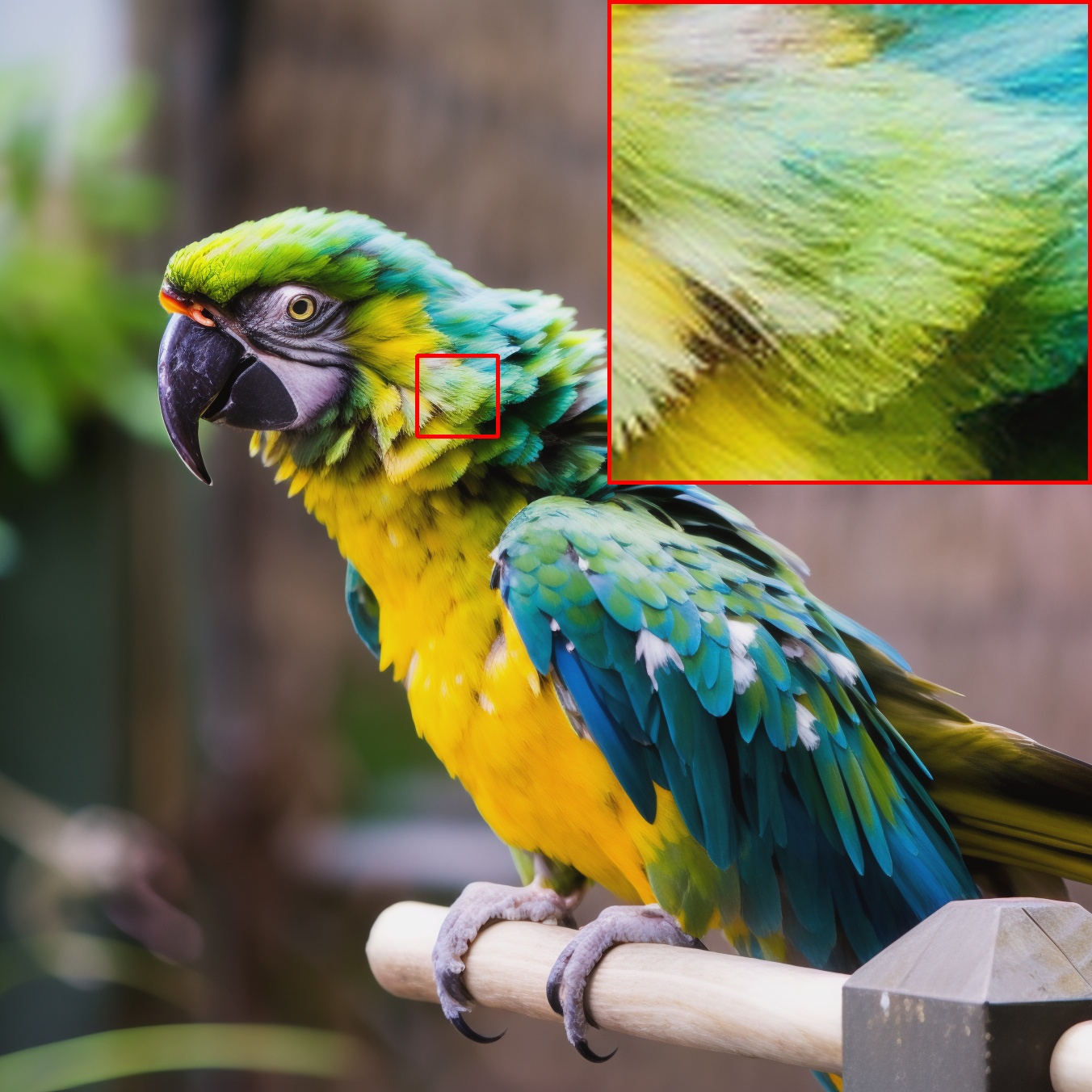}
\end{minipage}
\begin{minipage}[b]{0.24\textwidth}
    \centering
    \includegraphics[width=\linewidth]{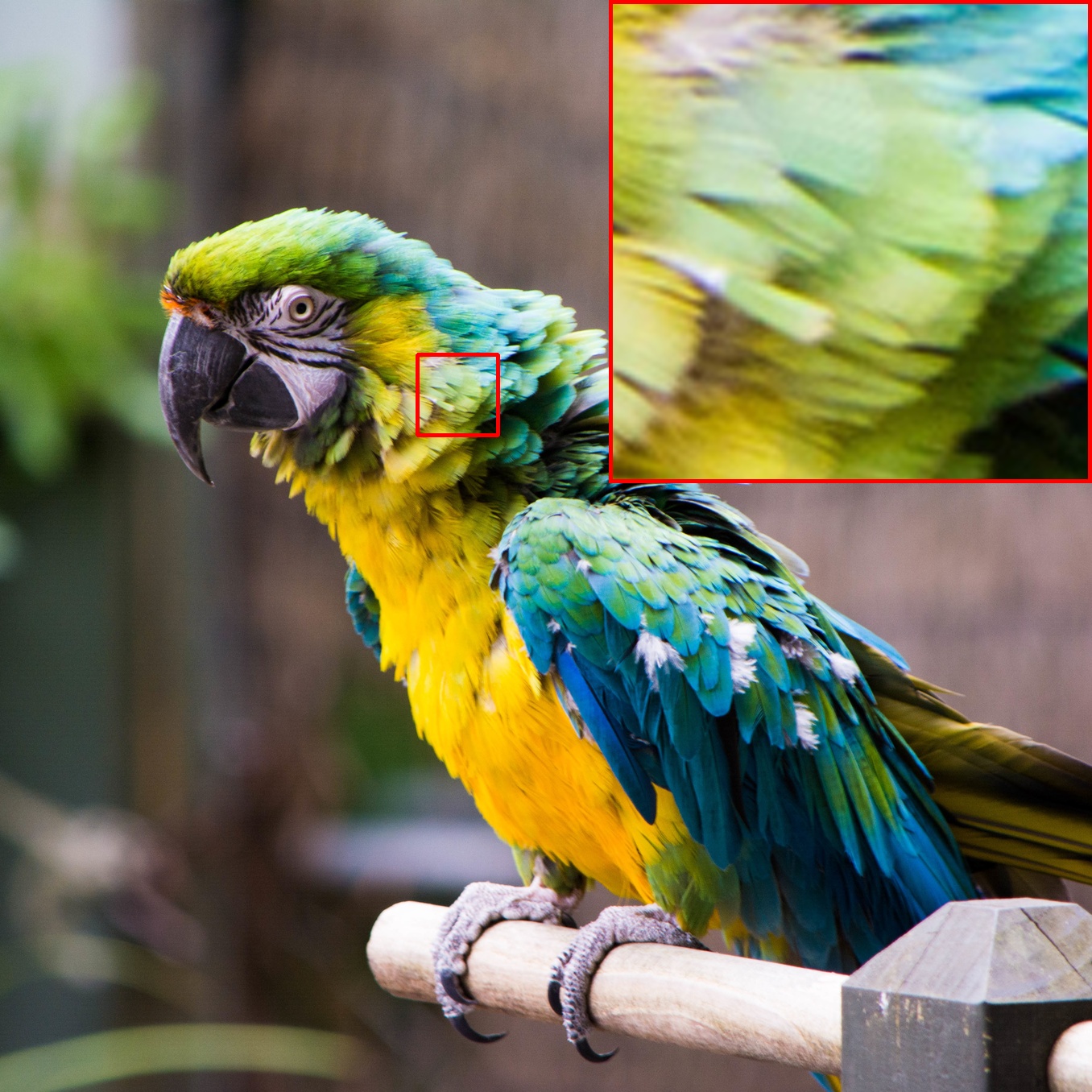}
\end{minipage}

\begin{minipage}[b]{0.24\textwidth}
    \centering
    \includegraphics[width=\linewidth]{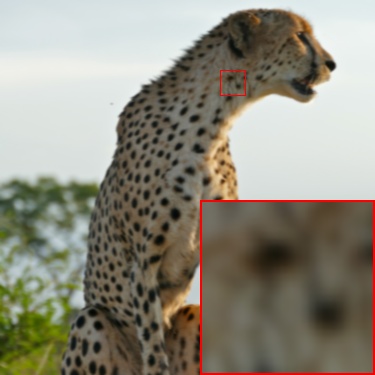}
    \\[-5pt]{\scriptsize LR}
\end{minipage}
\begin{minipage}[b]{0.24\textwidth}
    \centering
    \includegraphics[width=\linewidth]{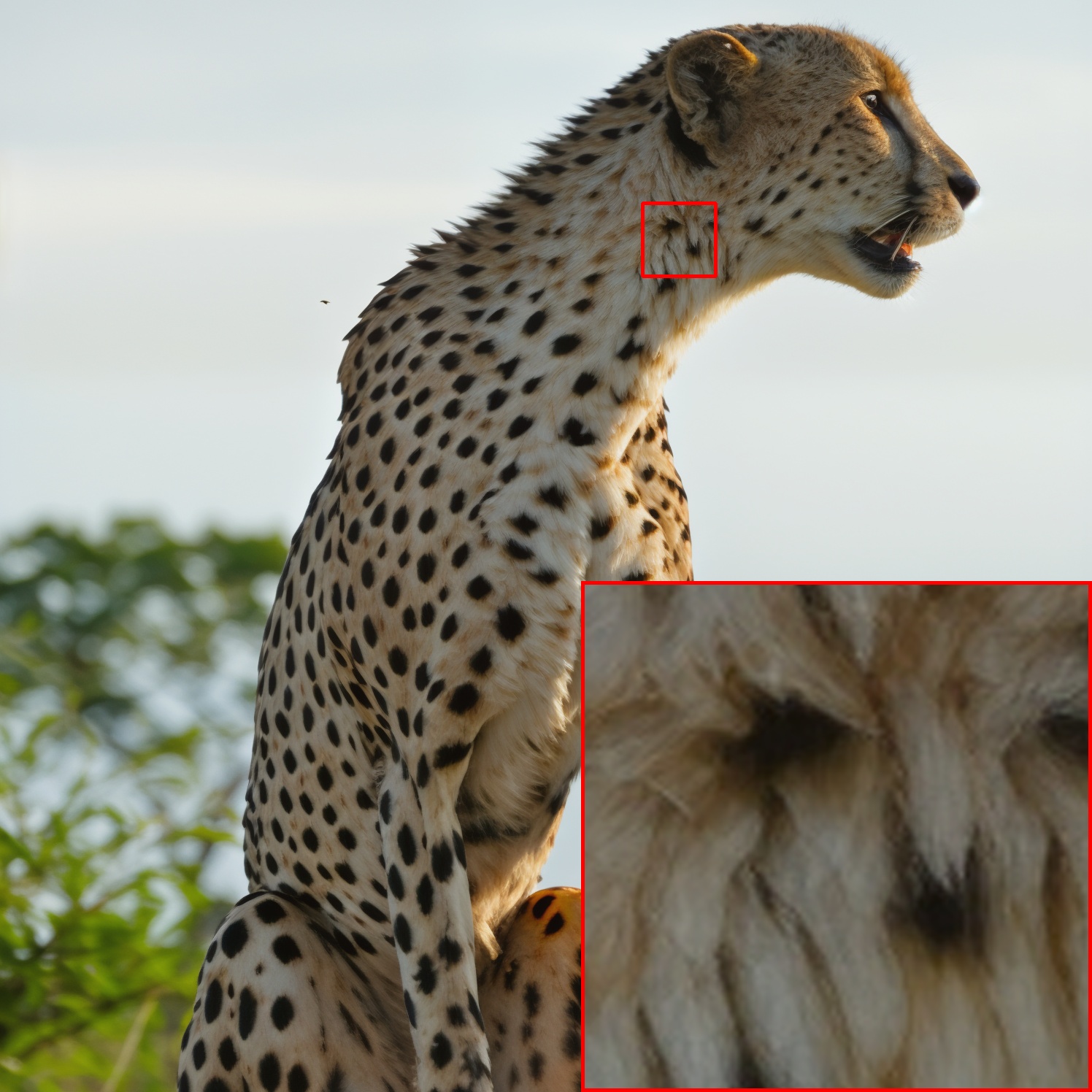}
    \\[-5pt]{\scriptsize Baseline}
\end{minipage}
\begin{minipage}[b]{0.24\textwidth}
    \centering
    \includegraphics[width=\linewidth]{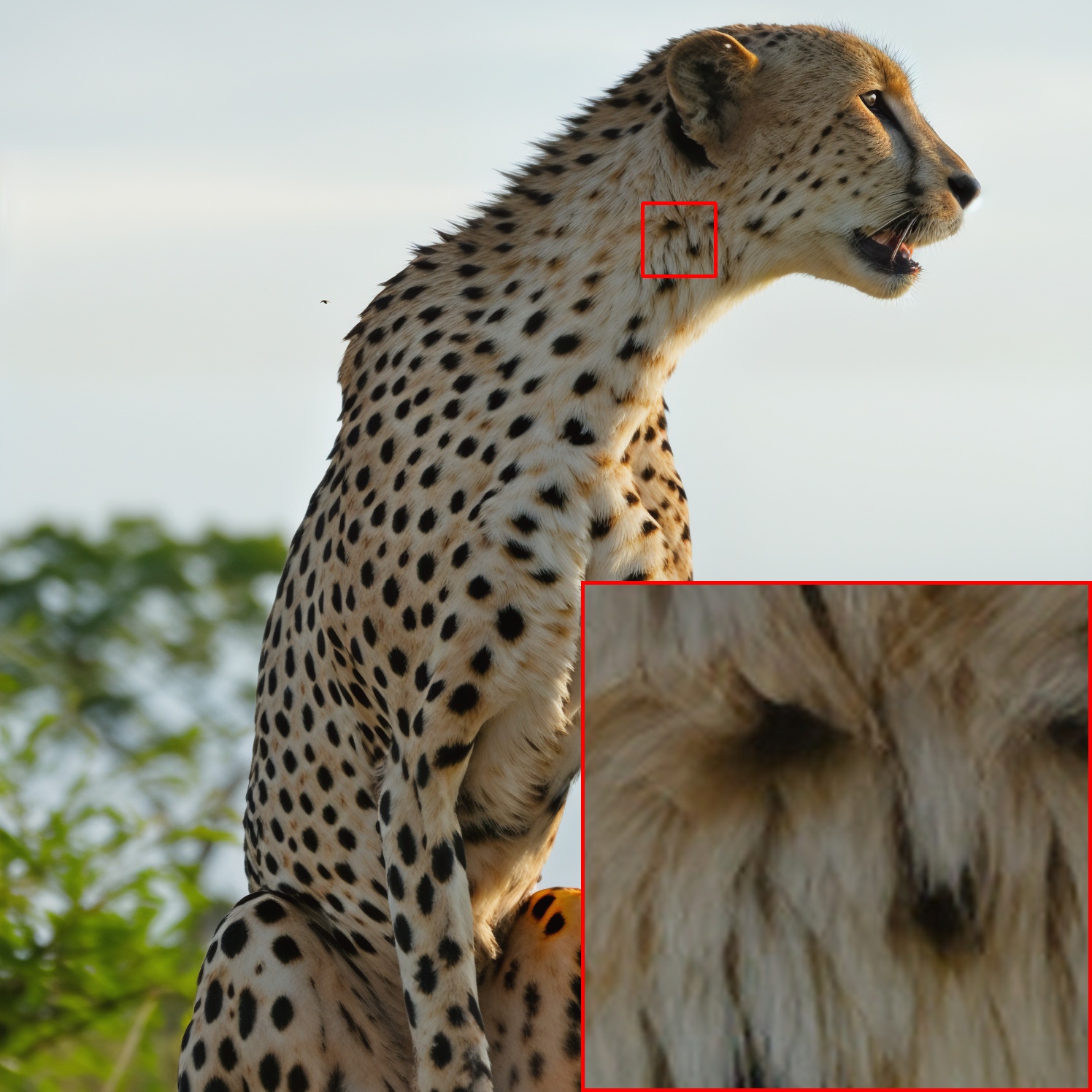}
    \\[-5pt]{\scriptsize Baseline+RQI}
\end{minipage}
\begin{minipage}[b]{0.24\textwidth}
    \centering
    \includegraphics[width=\linewidth]{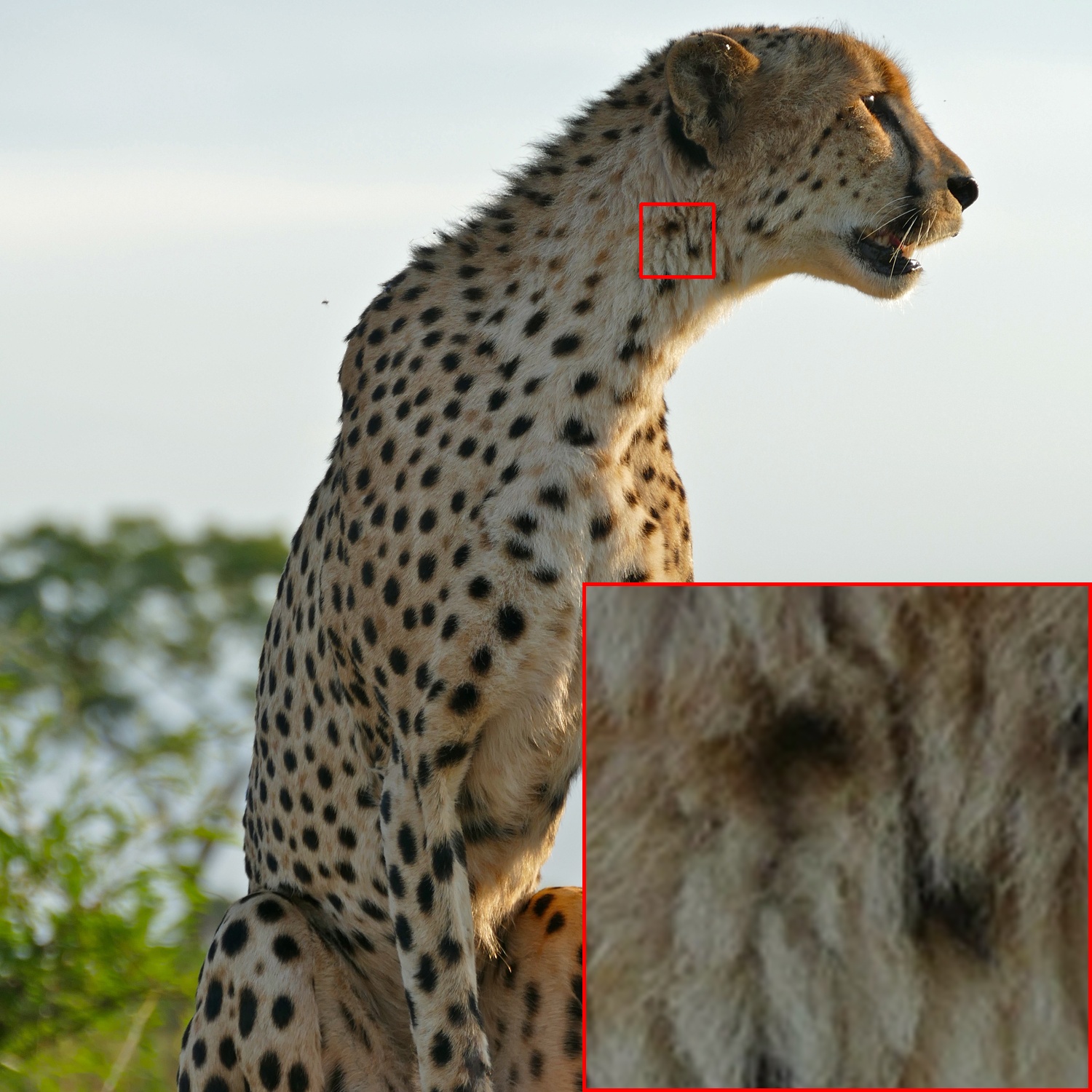}
    \\[-5pt]{\scriptsize GT}
\end{minipage}

\caption{More visual comparison of training advanced SR models with RQI as an auxiliary loss. RQI is effective in generating more visually appealing details. Please zoom in for a better view.}
\label{fig12}
\vspace{-6pt}
\end{figure*}

In Figure \ref{fig11} and Figure \ref{fig12}, we show more visual comparisons when training SR models with the RQI loss. 
Since the generation ability is limited for SwinIR \cite{SwinIR} due to its non-diffusion architecture, we compare with SeeSR \cite{SeeSR} or PiSA-SR \cite{Pisasr} as baseline models in the figures.
RQI is shown particularly effective in preserving structural fidelity (Figure \ref{fig11}) and producing more visually appealing textures (Figure \ref{fig12}). These results further demonstrate the advantages of the proposed approach.

\section{Limitations and Discussions} 
\label{secF}

Although RQI has demonstrated strong effectiveness in both evaluating and optimizing SR models, it also has a couple of limitations, as discussed below.

First, the RQI score is meaningful when comparing images that share the same reference. It cannot be used for cross-content quality comparison, because different contents may correspond to references with varying quality levels, making the scores incomparable across images.

Second, RQI primarily measures perceptual quality. In SR problem, fine textures are often irreversibly lost during the degradation process, making pixel-accurate fidelity evaluation fundamentally impossible. As a result, RQI assesses only the perceptual quality of these reconstructed textures. Designing a more principled fidelity measure for such texture-level comparisons remains an important direction for future work.

